\documentclass[10pt]{article} 
\usepackage[table]{xcolor}
\usepackage[colorlinks,citecolor=blue]{hyperref}
\usepackage[preprint]{tmlr}
\usepackage{wrapfig}
\usepackage[utf8]{inputenc} 
\usepackage[T1]{fontenc}    
\usepackage{hyperref}       
\usepackage{xurl}
\usepackage{booktabs}       
\usepackage{amsfonts}       
\usepackage{nicefrac}       
\usepackage{microtype}      
\usepackage{natbib}
\usepackage{graphicx}
\usepackage[edges]{forest}
\usepackage{tcolorbox}
\usepackage{breakurl}
\usepackage{float}
\usepackage{amsmath}
\usepackage{amsfonts}
\usepackage{subcaption}
\usepackage{amssymb}
\usepackage{makecell}
\usepackage{pbox}
\usepackage{array}
\usepackage{tabularx}
\usepackage{diagbox}
\usepackage[toc]{multitoc}
\usepackage{enumerate}
\usepackage{listings}
\usepackage{subcaption}
\usepackage{lipsum}

\usepackage{amsmath,amsfonts,bm}

\def\eqref#1{equation~\ref{#1}}

\def\1{\bm{1}}

\DeclareMathAlphabet{\mathsfit}{\encodingdefault}{\sfdefault}{m}{sl}
\SetMathAlphabet{\mathsfit}{bold}{\encodingdefault}{\sfdefault}{bx}{n}

\title{Interactive Natural Language Processing}

\def\month{MM}  
\def\year{YYYY} 
\def\openreview{\url{https://openreview.net/forum?id=XXXX}} 

\definecolor{zekundarkgreen}{RGB}{0,127,0}
\definecolor{darkblue}{rgb}{0,0.08,0.45}

\begin{document}

\newcommand*\samethanks[1][\value{footnote}]{\footnotemark[#1]}

\author{\large 
\textbf{Zekun Wang}\textsuperscript{*1,2}, 
\textbf{Ge Zhang}\textsuperscript{*1,3}, 
\textbf{Kexin Yang}\textsuperscript{4}, 
\textbf{Ning Shi}\textsuperscript{5}, 
\textbf{Wangchunshu Zhou}\textsuperscript{6}, \\
\textbf{Shaochun Hao}\textsuperscript{1,7}, 
\textbf{Guangzheng Xiong}\textsuperscript{1,8}, 
\textbf{Yizhi Li}\textsuperscript{9}, 
\textbf{Mong Yuan Sim}\textsuperscript{1,10}, \\
\textbf{Xiuying Chen}\textsuperscript{11}, 
\textbf{Qingqing Zhu}\textsuperscript{7}, 
\textbf{Zhenzhu Yang}\textsuperscript{12}, 
\textbf{Adam Nik}\textsuperscript{1,13}, 
\textbf{Qi Liu}\textsuperscript{14} \\
\\
\textbf{Chenghua Lin}\textsuperscript{\dag~9}, 
\textbf{Shi Wang}\textsuperscript{17}, 
\textbf{Ruibo Liu}\textsuperscript{15}, 
\textbf{Wenhu Chen}\textsuperscript{3}, 
\textbf{Ke Xu}\textsuperscript{2}, \\
\textbf{Dayiheng Liu}\textsuperscript{4}, 
\textbf{Yike Guo}\textsuperscript{18}, 
\textbf{Jie Fu}\textsuperscript{\dag~1,16} \\
\\
\small
\textsuperscript{1}1Cademy Community  
\textsuperscript{2}Beihang University  
\textsuperscript{3}University of Waterloo  
\textsuperscript{4}Sichuan University  \\
\textsuperscript{5}University of Alberta  
\textsuperscript{6}ETH Zurich  
\textsuperscript{7}Independent Researcher  \\
\textsuperscript{8}Beijing University of Posts and Telecommunications  
\textsuperscript{9}University of Sheffield  \\
\textsuperscript{10}The University of Adelaide  
\textsuperscript{11}KAUST  
\textsuperscript{12}China University of Geosciences, Beijing  \\
\textsuperscript{13}Carleton College  
\textsuperscript{14}City University of Hong Kong  
\textsuperscript{15}Dartmouth College  
\textsuperscript{16}Mila \\
\textsuperscript{17}ICT, Chinese Academy of Sciences 
\textsuperscript{18}Hong Kong University of Science and Technology  \\
\\
\textsuperscript{*}\text{Primary authors} \quad \textsuperscript{\dag}\text{Corresponding authors}
\thanks{Correspondence to: zenmoore@buaa.edu.cn, gezhang@umich.edu, jie.fu@polymtl.ca, and c.lin@sheffield.ac.uk. \\
Author contributions are listed at the end of the paper.
}
}

\maketitle

\vspace{2em}

\begin{abstract}
Interactive Natural Language Processing (iNLP) has emerged as a novel paradigm within the field of NLP, aimed at addressing limitations in existing frameworks while aligning with the ultimate goals of artificial intelligence. This paradigm considers language models as agents capable of observing, acting, and receiving feedback iteratively from external entities. 

Specifically, language models in this context can: 
(1) interact with humans for better understanding and addressing user needs, personalizing responses, aligning with human values, and improving the overall user experience; 
(2) interact with knowledge bases for enriching language representations with factual knowledge, enhancing the contextual relevance of responses, and dynamically leveraging external information to generate more accurate and informed responses; 
(3) interact with models and tools for effectively decomposing and addressing complex tasks, leveraging specialized expertise for specific subtasks, and fostering the simulation of social behaviors; and 
(4) interact with environments for learning grounded representations of language, and effectively tackling embodied tasks such as reasoning, planning, and decision-making in response to environmental observations. 

This paper offers a comprehensive survey of iNLP, starting by proposing a unified definition and framework of the concept. We then provide a systematic classification of iNLP, dissecting its various components, including interactive objects, interaction interfaces, and interaction methods. We proceed to delve into the evaluation methodologies used in the field, explore its diverse applications, scrutinize its ethical and safety issues, and discuss prospective research directions. This survey serves as an entry point for researchers who are interested in this rapidly evolving area and offers a broad view of the current landscape and future trajectory of iNLP. 
\end{abstract}

\clearpage
{
  \hypersetup{linkcolor=black}
  \tableofcontents
}
\clearpage

\section{Introduction}
Natural Language Processing (NLP) has witnessed a remarkable revolution in recent years, thanks to the development of generative pre-trained language models (PLMs) such as BART~\citep{lewis2019bart}, T5~\citep{2020t5}, GPT-3~\citep{brown2020language}, PaLM~\citep{chowdhery2022palm}, to name a few. 
These models can generate coherent and semantically meaningful text, making them useful for various NLP tasks such as machine translation~\citep{mbart}, summarization~\citep{bert-summary, plm-summary-1}, and question answering~\citep{radford2019language, brown2020language, 2020t5}. 
However, these models also have clear limitations such as 
misalignment with human needs~\citep{wolf2023fundamental, alignment-deepmind}, 
lack of interpretability~\citep{ai-chains, gpt4}, 
hallucinations~\citep{hallucination, hallucination-survey, gpt4}, 
imprecise mathematical operations~\citep{toolformer, augmented-lm}, 
inadequate experience grounding~\citep{experience-ground}, and 
limited ability for complex reasoning~\citep{reasoning-survey, towards-reasoning}, 
among others~\citep{borji2023categorical}. 

To address these limitations, a new paradigm of natural language processing has emerged: \textbf{interactive natural language processing (iNLP)}~\citep{experience-ground, social-neuro-ai}. 
There have been a variety of definitions for \textit{``interactive''}
in the NLP and Machine Learning literature, where the term typically refers to the involvement of humans in the process. 
For example, 
~\cite{iML} define Interactive Machine Learning (iML) as ``\textit{an active machine learning technique in which models are designed and implemented with human-in-the-loop manner.}'' 
~\cite{faltings2023interactive} view Interactive Text Generation as ``\textit{a task that allows training generation models interactively without the costs of involving real users, by using user simulators that provide edits that guide the model towards a given target text.}''  
~\cite{wang2021putting} describe Human-in-the-loop (HITL) as ``\textit{where model developers continuously integrate human feedback into different steps of the model deployment workflow.}'' 
The popularity of ChatGPT\footnote{\url{https://openai.com/blog/chatgpt}} also demonstrated the impressive capabilities of human-LM interaction via reinforcement learning from human feedback (RLHF).
Although humans are the most common type of objects for interacting with language models, recent research has revealed other important object types for interaction, which 
include Knowledge Bases (KBs)~\citep{retrieval-survey, knowledge-survey}, Models/Tools~\citep{reasoning-survey, augmented-lm, lm-cascades, yao2022react, shen2023hugginggpt, qin2023tool}, and Environments~\citep{li2022pre, yang2023foundation, ahn2022can, huang2022inner, chatgpt4robot, bubeck2023sparks-AGI}. 
Therefore, in our survey, we first define interactive natural language processing which accounts for a broader scope of objects that can interact with language models:

\textbf{Interactive Natural Language Processing (iNLP) considers language models as agents capable of \textbf{observing}, \textbf{acting}, and \textbf{receiving feedback} in a loop with external objects such as humans, knowledge bases, tools, models, and environments\footnote{
\textbf{Observation} involves all kinds of inputs to language models. 
\textbf{Action} involves all kinds of outputs of language models such as text generation~\citep{ouyang2022training}, requesting for external objects~\citep{yao2022react, toolformer}, text editing~\citep{faltings2023interactive}, etc. 
\textbf{Feedback} involves feedback messages passed from external objects to language models such as scoring from humans~\citep{ouyang2022training}.}.}

Specifically, through interaction, a language model (LM) can leverage external resources to improve its performance and address its limitations mentioned in the first paragraph. 
For example, 
interacting with humans aligns language models better with human needs and human values (e.g., helpfulness, harmlessness, honesty)~\citep{ouyang2022training, help-harm} and 
interacting with KBs can help language models 
alleviate hallucinations
~\citep{hallucination-survey}. 
Likewise, 
interacting with models or tools can improve the abilities of LMs such as reasoning, faithfulness, and exactitude of mathematical operations~\citep{augmented-lm, toolformer}. 
And finally, interacting with environments can enhance the grounded reasoning capability of LMs~\citep{liu2022mind} and promote the applications of LMs in embodied tasks~\citep{zeng2022socratic, yang2023foundation}.

Furthermore, interaction may hold the potential to unlock future milestones in language processing, which can be considered the holy grail of artificial intelligence~\citep{experience-ground}. 
In 2020, ~\cite{experience-ground} have examined the future direction of natural language processing and proposed five levels of world scope to audit progress in NLP: ``(1) Corpus; (2) Internet; (3) Perception (multimodal NLP); (4) Embodiment; (5) Social.'' Notably, the recent release of GPT-4~\citep{gpt4} and PaLM-2~\citep{google2023palm2}, which are large multimodal language models, has brought significant advancements to the third level ``Perception''. Embodied AI and Social Embodied AI fundamentally posit that a more comprehensive language representation can be learned through the establishment of an interactive loop involving language model agents, environments, and humans~\citep{experience-ground, social-neuro-ai, social-learning-theory, language-remodeling, lake2016building, driess2023palme, communicative-learning}. 
This perspective highlights the need for the NLP community to shift its attention towards the fourth and fifth levels (``Embodiment'' and ``Social Interaction'') to propel the field forward. 
In addition to models, humans, and environments, 
tools and knowledge bases that facilitate connections between language models and the external world also play a significant role in enabling (social) embodiment~\citep{qin2023tool, xie2022unifiedskg, tool-embodiment, experience-ground}. 
The future achievement of social embodiment of language models may lead to significant phenomena, including artificial self-awareness~\citep{bubeck2023sparks-AGI, kosinski2023theory} and the emergence of a language model society~\citep{park2023generative, camel}.

Therefore, interactive NLP is beneficial for both NLP researchers and practitioners, since it has the potential to address limitations such as hallucination~\citep{hallucination-survey} and alignment~\citep{wolf2023fundamental}, while also aligning with the ultimate goals of AI~\citep{bubeck2023sparks-AGI, experience-ground, qin2023tool}. 
Notably, 
with the recent release of ChatGPT and GPT-4~\citep{gpt4}, which have overwhelmed the NLP community and are considered the spark of artificial general intelligence (AGI) by some researchers due to their remarkable universal capabilities~\citep{bubeck2023sparks-AGI}, 
the NLP community is now experiencing a shift in focus towards posing new challenges in the field. 
This transition has prompted numerous surveys and position papers that aim to propose novel research directions, with many of them addressing the theme of interaction. 
For example, 
~\cite{augmented-lm} survey the strategies that PLMs employ cascading mechanisms for reasoning~\citep{lm-cascades} and utilize tools for taking action. 
But ~\citep{augmented-lm} lacks an in-depth discussion on interactivity, and focuses solely on tool use and reasoning, while overlooking other topics such as interaction with knowledge bases, and simulation of social behavior. 
\cite{yang2023foundation} investigate the cross-disciplinary research field of foundation models and decision making, with a particular emphasis on exploring the interactions of language models with humans, tools, agents, and environments. 
But they primarily focus on decision-making settings and reinforcement learning formalisms, without providing a comprehensive discussion on interacting with knowledge bases or the interaction methodology from the perspective of NLP techniques, such as chain-of-thought prompting~\citep{Wei2022ChainOT}. 
~\cite{bubeck2023sparks-AGI} discuss the interactions of language models with the world based on tool-use and embodiment, as well as their interactions with humans based on Theory of Mind (ToM) and self-explanation. 
But they primarily focus on evaluating the abilities of large language models (LLMs) and lack a comprehensive discussion of the interaction methodology employed in the studies. 
Other surveys and works~\citep{Evaluating_Human_Language_Model_Interaction, qin2023tool, chatgpt4robot, communicative-learning} have also contributed valuable insights to the theme of interaction. 
However, they are also specific to certain aspects and do not offer a unified and systematic review that covers the entire spectrum of interactive NLP. 

Clearly, the field of interactive NLP has undergone significant development in the past few years, with the emergence of new forms of interactive objects that go beyond the standard Human-in-the-loop approach. These new forms of objects encompass knowledge bases, models/tools, and environments. While the aforementioned works provide some coverage of interactions involving models/tools and environments, there is a notable absence of discussions regarding interactions with language models using knowledge bases (KB). 
Furthermore, there is a lack of a comprehensive review of methodologies in the context of interactive NLP. 
Hence, the main goals of our survey are: 
\enlargethispage{\baselineskip}
\begin{enumerate}
\item \textbf{Unified Definition and Formulation}: to provide a unified definition and formulation of interactive NLP, establishing it as a new paradigm of NLP. 
\item \textbf{Comprehensive Classification}: to provide a comprehensive breakdown of iNLP along dimensions such as interactive objects, interaction interfaces, and interaction methods, enabling a systematic understanding of its different aspects and components. 
\item \textbf{Further Discussion}: to survey the evaluation methodologies used in iNLP, examine its diverse applications, and discuss the ethical and safety issues as well as the future directions in this field. 
\end{enumerate}

\begin{figure}[ht]
    \centering
    \begin{subfigure}[b]{0.45\textwidth}
        \includegraphics[width=\textwidth]{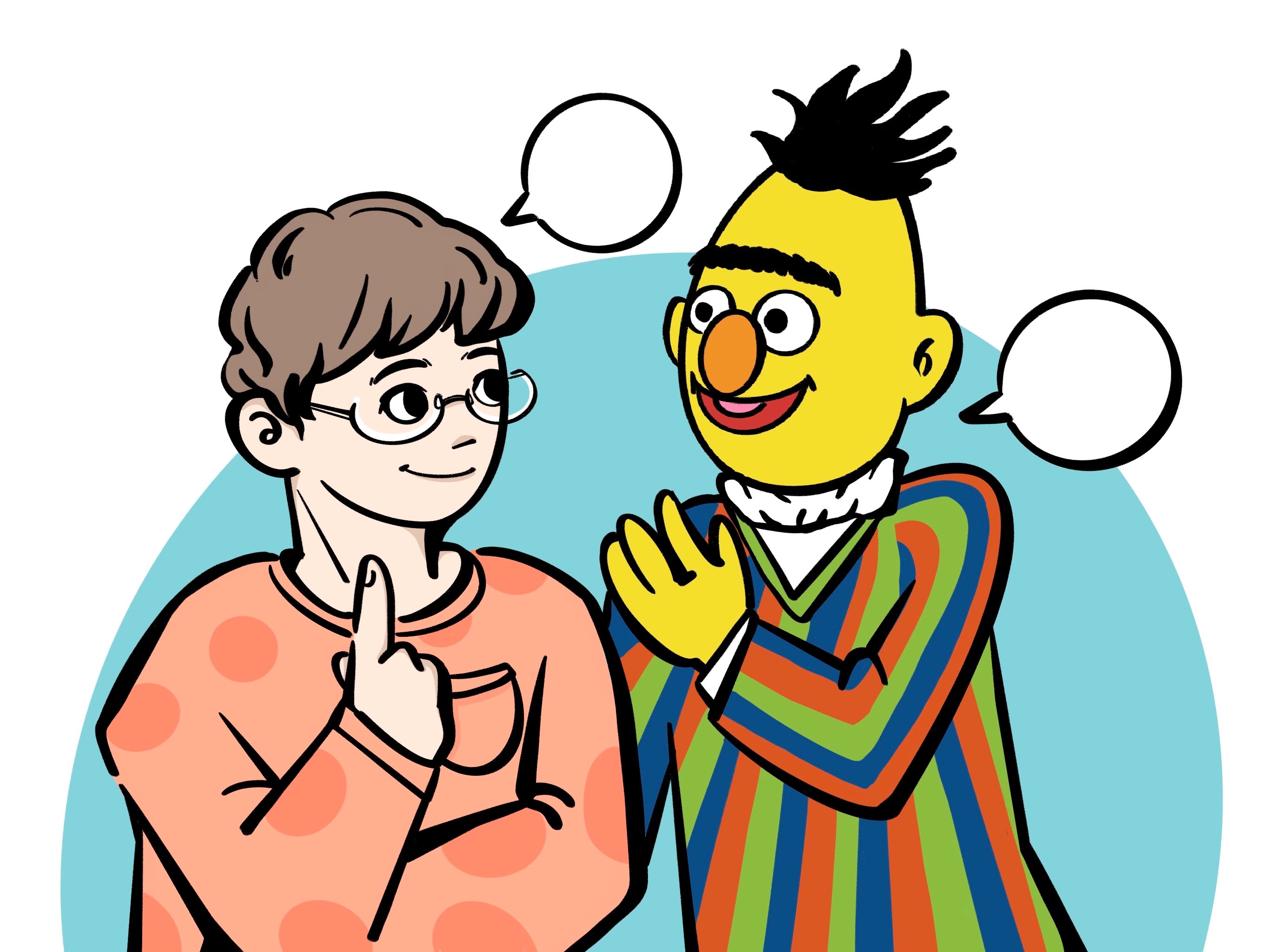}
        \caption{Interacting with Humans.}
        \label{fig:human}
    \end{subfigure}
    ~ 
    \begin{subfigure}[b]{0.45\textwidth}
        \includegraphics[width=\textwidth]{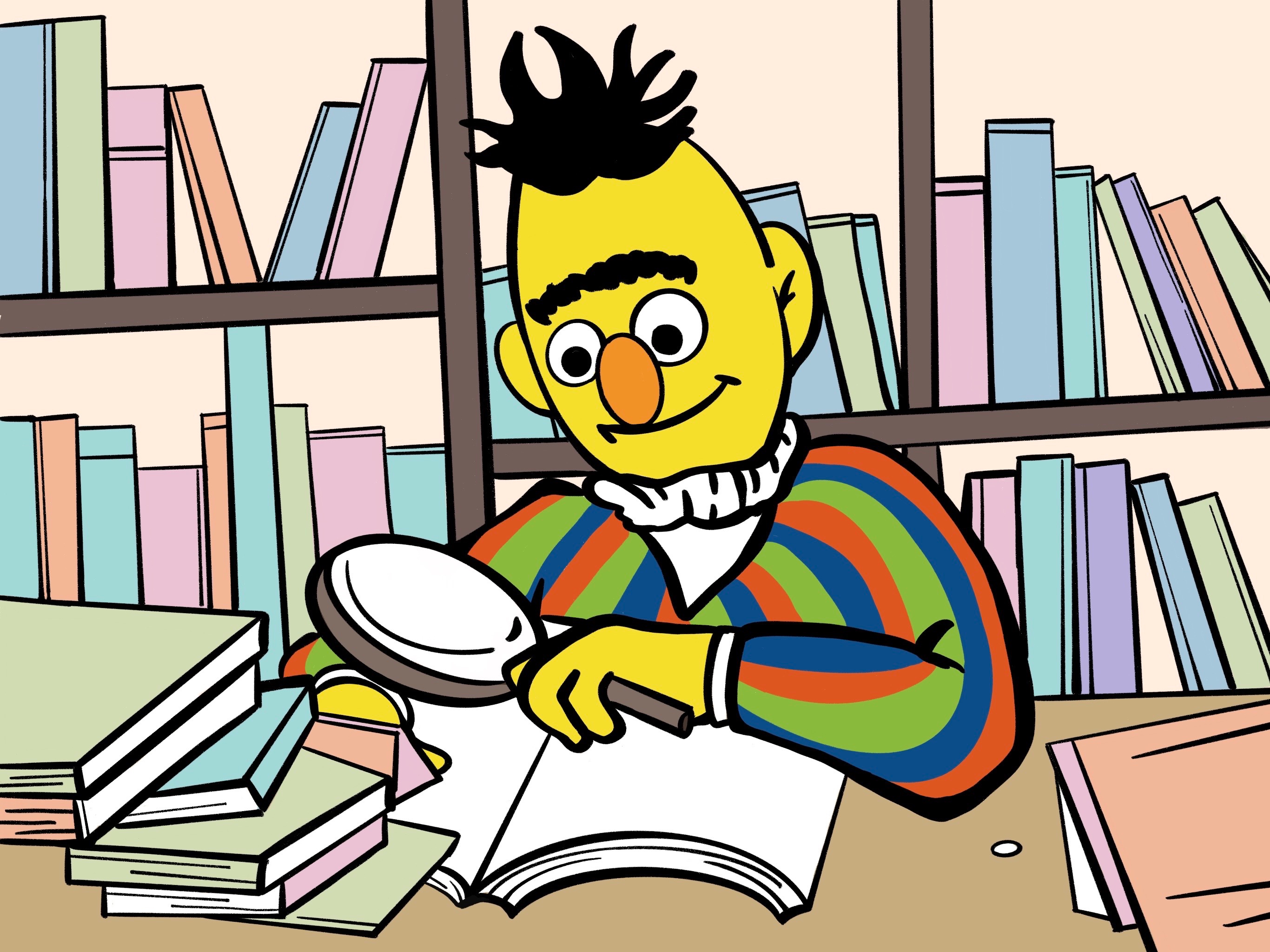}
        \caption{Interacting with Knowledge Bases.}
        \label{fig:kb}
    \end{subfigure}
    \\
    \begin{subfigure}[b]{0.45\textwidth}
        \includegraphics[width=\textwidth]{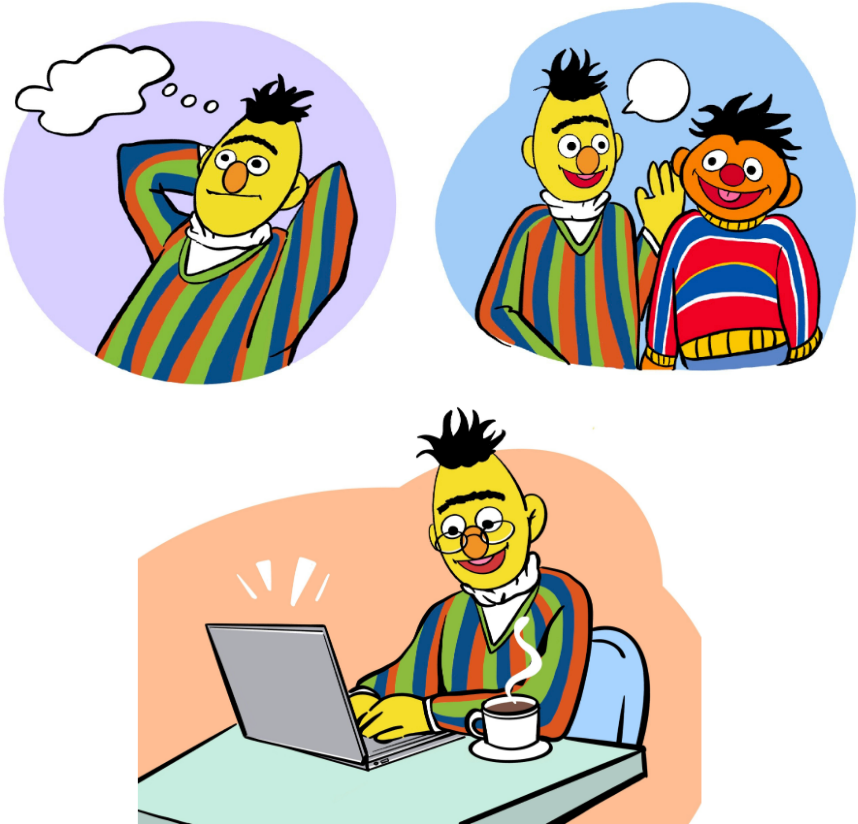}
        \caption{Interacting with Models and Tools\footnote{Self-interaction is also included.}.}
        \label{fig:model-tool}
    \end{subfigure}
    ~ 
    \begin{subfigure}[b]{0.45\textwidth}
        \includegraphics[width=\textwidth]{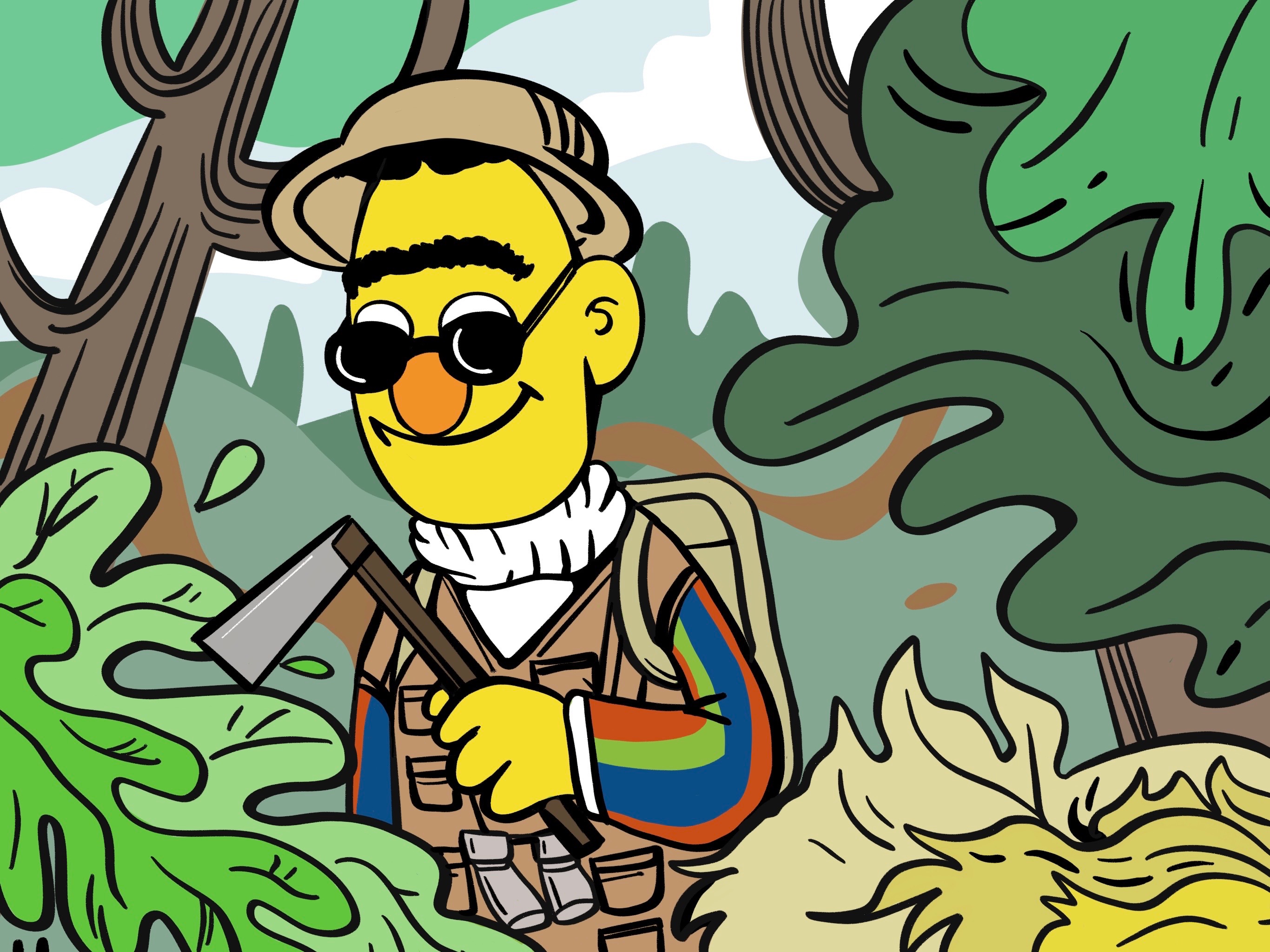}
        \caption{Interacting with Environments.}
        \label{fig:env}
    \end{subfigure}
    \caption{The paradigm of Interactive Natural Language Processing.}
    \label{fig:inlp}
\end{figure}

We believe that conducting such a survey is highly timely, and our paper aims to fill the gaps of aforementioned surveys by serving as an entry point for researchers who are interested in pursuing research in this important and fast-evolving area but may not yet be familiar with it. 
As illustrated in Figure \ref{fig:inlp}, we will start with an in-depth discussion about interactive objects (§\ref{objects}), followed by an overview of interaction interfaces by which the language models communicate with the external objects (§\ref{interface}). We then organize a variety of interaction methods by which the language models fan in and out interaction messages (§\ref{methods}). This is followed by a discussion about evaluation in the context of iNLP (§\ref{eval}). Finally, we will examine the current applications of iNLP (§\ref{apps}), discuss ethical and safety issues (\S\ref{ethics}), and suggest future directions and challenges (§\ref{future}). Taxonomy \ref{taxonomy} gives a bird's-eye view of our survey.

\vspace{5em}

\clearpage

\definecolor{lightpurple}{RGB}{228, 228, 255}
\definecolor{lightred}{RGB}{255,229,228}
\definecolor{lightblue}{RGB}{226,244,253}
\definecolor{lightorange}{RGB}{255,242,229}
\definecolor{lightviolet}{RGB}{252,231,241}

\tikzset{
    parent/.style =          {align=center,text width=1.75cm,rounded corners=3pt},
    child/.style =           {align=center,text width=2cm,rounded corners=3pt},
    grandchild/.style =      {align=center,text width=4cm,rounded corners=3pt},
    greatgrandchild/.style = {align=center,text width=1.75cm,rounded corners=3pt},
    greatgreatgrandchild/.style = {align=center,text width=1.75cm,rounded corners=3pt},
    referenceblock/.style =  {align=center,text width=7cm,rounded corners=3pt}
}

\tikzstyle{every node}=[font=\scriptsize]

\forestset{
  parent color/.style args={#1}{
    {fill=#1},
    for tree={fill/.wrap pgfmath arg={#1!##1}{1/level()*80},draw=#1!80!darkgray}
  },
  root color/.style args={#1}{fill={{#1!60!gray!25},draw=#1!80!darkgray}}
}

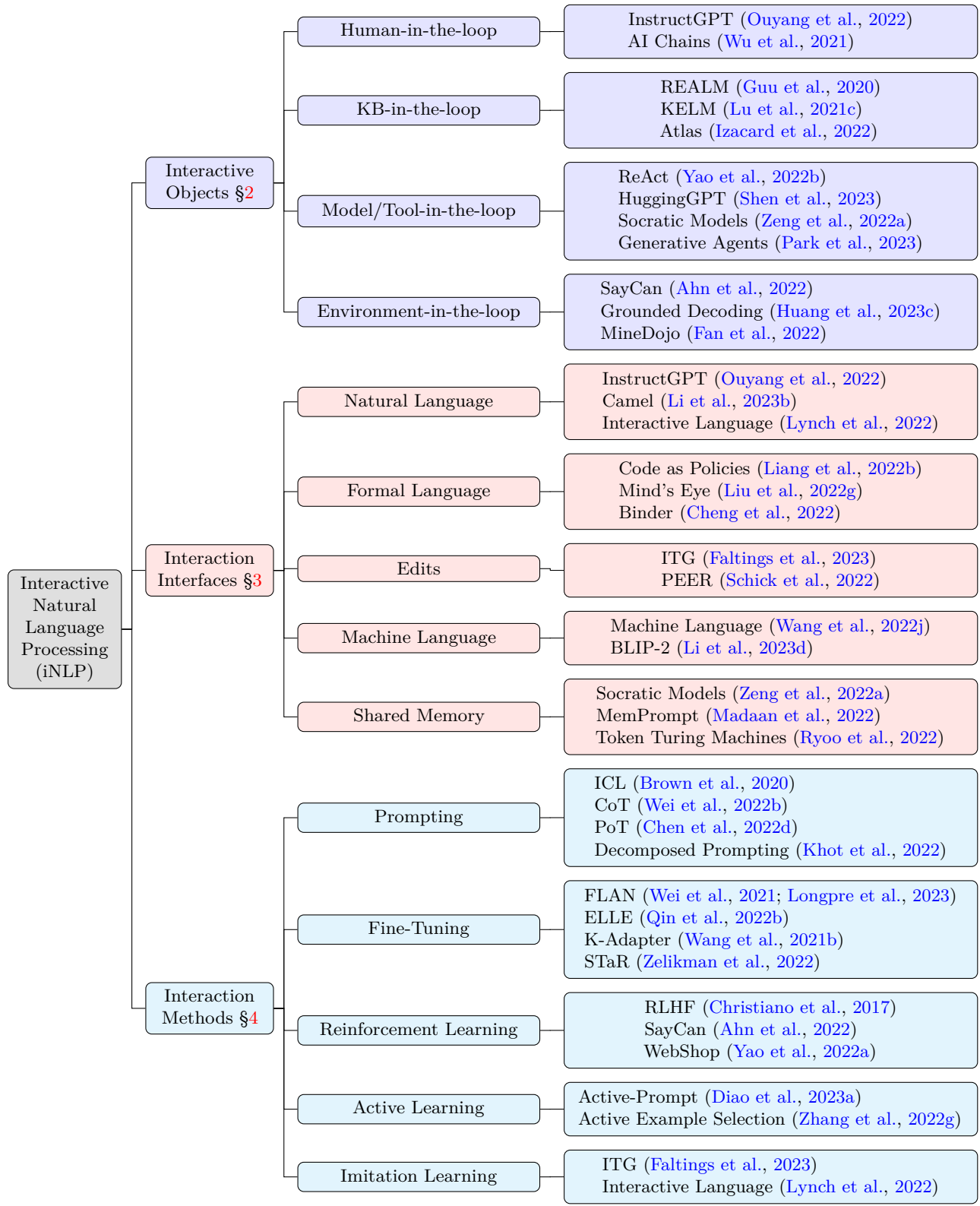
\begin{figure}[h]
\centering
        \begin{forest}
            for tree={
                forked edges,
                grow'=east,
                draw,
                rounded corners,
                node options={align=center,},
                text width=2em,
                font=\small, 
            },
            [Interactive Natural Language Processing (iNLP), fill=gray!25, parent
                [Interactive Objects §\ref{objects}, for tree={fill=lightpurple, child}
                    [Human-in-the-loop, fill=lightpurple, grandchild
                      [\pbox{7cm}{InstructGPT~\citep{ouyang2022training} \\ AI Chains~\citep{ai-chains}}, fill=lightpurple, referenceblock
                      ]
                    ]
                    [KB-in-the-loop, fill=lightpurple, grandchild
                        [\pbox{7cm}{REALM~\citep{guu2020realm}
                        \\KELM~\citep{lu2021kelm}
                        \\Atlas~\citep{izacard2022atlas}}, 
                        fill=lightpurple, referenceblock]
                    ]
                    [Model/Tool-in-the-loop, fill=lightpurple, grandchild
                        [\pbox{7cm}{
                            ReAct~\citep{yao2022react}
                            \\HuggingGPT~\citep{shen2023hugginggpt}
                            \\Socratic Models~\citep{zeng2022socratic}
                            \\Generative Agents~\citep{park2023generative}}
                        , fill=lightpurple, referenceblock]
                    ]
                    [Environment-in-the-loop, fill=lightpurple, grandchild
                        [\pbox{7cm}{
                        SayCan~\citep{ahn2022can}
                        \\Grounded Decoding~\citep{huang2023grounded}
                        \\MineDojo~\citep{fan2022minedojo}
                        }, 
                        fill=lightpurple, referenceblock]
                    ]
                ]
                [Interaction Interfaces §\ref{interface}, for tree={fill=lightred,child }
                    [Natural Language, fill=lightred, grandchild
                        [\pbox{7cm}{
                        InstructGPT~\citep{ouyang2022training}
                        \\Camel~\citep{camel}
                        \\Interactive Language~\citep{lynch2022interactive}
                        }, 
                        fill=lightred, referenceblock]
                    ] 
                    [Formal Language, fill=lightred, grandchild
                        [\pbox{7cm}{
                        Code as Policies~\citep{code-as-policies}
                        \\Mind's Eye~\citep{liu2022mind}
                        \\Binder~\citep{cheng2022binding}
                        }, 
                        fill=lightred, referenceblock]
                    ]
                    [Edits, fill=lightred, grandchild
                        [\pbox{7cm}{
                        ITG~\citep{faltings2023interactive}
                        \\PEER~\citep{schick2022peer}
                        }, 
                        fill=lightred, referenceblock]
                    ]
                    [Machine Language, fill=lightred, grandchild
                        [\pbox{7cm}{
                        Machine Language~\citep{machine-language}
                        \\BLIP-2~\citep{li2023blip2}
                        }, 
                        fill=lightred, referenceblock]
                    ]
                    [Shared Memory, fill=lightred, grandchild
                        [\pbox{7cm}{
                        Socratic Models~\citep{zeng2022socratic}
                        \\MemPrompt~\citep{madaan2022memory}
                        \\Token Turing Machines~\citep{ryoo2022token}
                        }, 
                        fill=lightred, referenceblock]
                    ]
                ]
                [Interaction Methods §\ref{methods}, for tree={fill=lightblue, child}
                    [Prompting, fill=lightblue, grandchild
                       [\pbox{7cm}{
                            ICL~\citep{brown2020language}
                            \\CoT~\citep{Wei2022ChainOT}
                            \\PoT~\citep{chen2022program}
                            \\Decomposed Prompting~\citep{DBLP:journals/corr/abs-2210-02406}
                        }, 
                        fill=lightblue, referenceblock]
                    ]
                    [Fine-Tuning, fill=lightblue, grandchild
                        [\pbox{7cm}{
                            FLAN~\citep{flan, flan-collection}
                            \\ELLE~\citep{qin-etal-2022-elle}
                            \\K-Adapter~\citep{kadapter}
                            \\STaR~\citep{zelikman2022star}
                        }, 
                        fill=lightblue, referenceblock]
                    ]
                    [Reinforcement Learning, fill=lightblue, grandchild
                        [\pbox{7cm}{
                            RLHF~\citep{christiano2017deep}
                            \\SayCan~\citep{ahn2022can}
                            \\WebShop~\citep{yao2022webshop}
                        }, 
                        fill=lightblue, referenceblock]
                    ]
                    [Active Learning, fill=lightblue, grandchild
                        [\pbox{7cm}{
                            Active-Prompt~\citep{diao2023active}
                            \\Active Example Selection~\citep{zhang2022active}
                        }, 
                        fill=lightblue, referenceblock]
                    ]
                    [Imitation Learning, fill=lightblue, grandchild
                        [\pbox{7cm}{
                            ITG~\citep{faltings2023interactive}
                            \\Interactive Language~\citep{lynch2022interactive}
                        }, 
                        fill=lightblue, referenceblock]
                    ]
                ]
            ]
        \end{forest}
        \caption{Taxonomy of interactive NLP.}
        \label{taxonomy}
\end{figure}

\clearpage

\section{Interactive Objects\label{objects}}

In this section, we will discuss the objects that interact with language models as illustrated in Figure \ref{fig:inlp}. When an entity is interacting with a language model, it is considered to be ``in the loop'', meaning that it is an active participant in the process of model training or model inference. 
As previously mentioned, the interactive objects include humans, knowledge bases, models/tools, and environments; each of which will be introduced in the following subsections.

\subsection{Human-in-the-loop\label{human-in-the-loop}}

\begin{wrapfigure}[16]{r}{7cm}
\centering
\includegraphics[width=1 \linewidth]{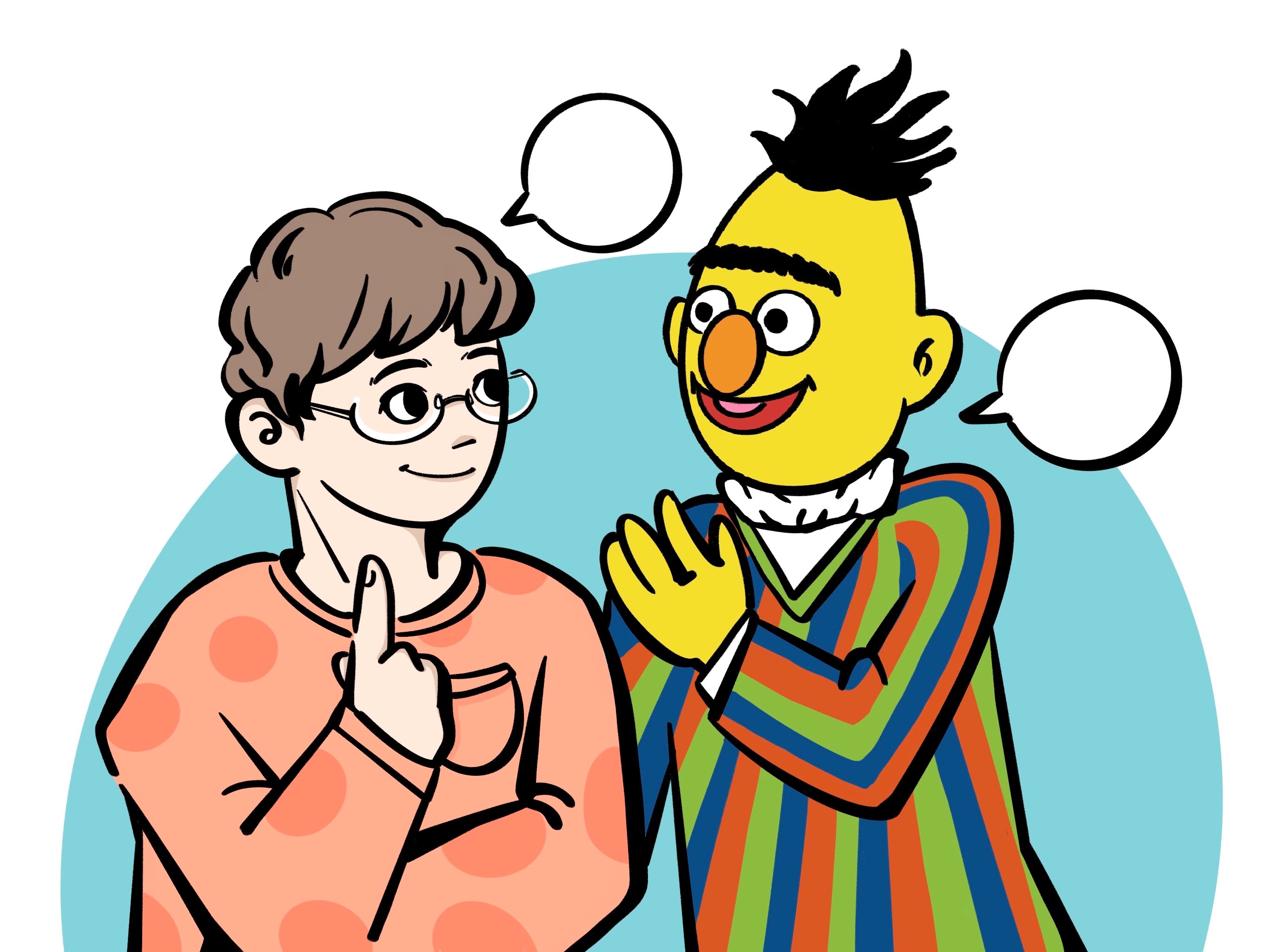}
\caption{Human-in-the-loop. 
}
\label{fig:human-wrap}
\end{wrapfigure}

Human-in-the-loop NLP represents a paradigm that emphasizes information exchange between humans and language models~\citep{wang2021putting}. This approach seeks to more effectively address users' needs and uphold human values, a concept known as Human-LM Alignment~\citep{help-harm, alignment-deepmind, ouyang2022training, scalable-align}. In contrast, earlier research on text generation primarily concentrated on the input and output of samples, overlooking aspects such as human preferences, experiences, personalization, diverse requirements, and the actual text generation process~\citep{Evaluating_Human_Language_Model_Interaction}.
In recent years, as pre-trained language models (PLMs) and large language models (LLMs) have matured, optimizing human-model interactions has emerged as a prevalent concern within the community. Incorporating human prompts, feedback, or configurations during the model training or inference stages, using either real or simulated users, proves to be an effective strategy for enhancing the Human-LM alignment~\citep{faltings2023interactive, ouyang2022training, ai-chains}.

Subsequently, we divide human-in-the-loop NLP into three types according to the schemes of user interaction, along with an additional section that delves into the simulation of human behaviors and preferences for these types, in order to enable scalable deployment of human-in-the-loop systems. These categories are: 

\begin{enumerate}
    \item Communicating with Human Prompts: users can interact with the model consecutively in a conversation.
    \item Learning from Human Feedback: users can provide feedback to update the parameters of LMs.
    \item Regulating via Human Configuration: users can configure the settings of LMs.
    \item Learning from Human Simulation: simulations of users are employed for the three aforementioned types, ensuring practical implementation and scalability.
\end{enumerate}

\paragraph{Communicating with Human Prompts.} 
This is the most general form of Human-LM interaction, which allows a language model to interact with a human in a conversational manner. 
The main purpose of this interaction scheme is to maintain real-time and continuous interaction, 
so typical application scenarios include dialogue systems, real-time translation, and multiple rounds of question answering. This interactive process of alternating iterations allows the output of the model to realign gradually to meet user requirements. 

Generally, this interaction scheme does not update the model's parameters during the interaction, instead requiring users to continuously input or update prompts to elicit more meaningful responses from the language model. As a result, conversation can be inflexible and labor-intensive due to the need for prompt engineering or dialogue engineering. To address these limitations, editing-based methods have been proposed by \cite{malmi-etal-2022-text, schick2022peer, faltings2023interactive, shi-etal-2022-text, du-etal-2022-read} to encourage the language model to modify existing output (c.f., \S\ref{chap-edits}). Additionally, context-based methods have been developed that enhance model output by adding examples or instructions to the input context, such as few-shot prompting or in-context learning~\citep{brown2020language}.

However, since these approaches do not involve adapting language models to accommodate human users, numerous trial edits or prompts may be required to achieve the desired outcome, resulting in lengthier dialogue rounds. As such, this interaction scheme can be inefficient and may lead to a suboptimal user experience. 

\paragraph{Learning from Human Feedback.} 
In contrast to ``Communicating with Human Prompts'', this interaction scheme provides feedback on the model's outputs, such as scoring, ranking, and offering suggestions, for model optimization. This feedback is therefore used to adjust the model's parameters, rather than simply acting as prompts for language models to respond. The primary objective of this interaction is to better adapt LMs for user needs and human values~\citep{help-harm}. 

For instance, ~\cite{Godbole2004interactive} and ~\cite{closing-the-loop} employ active learning to provide human feedback. By labeling a few examples based on model predictions, they update the model parameters to improve its understanding of human needs. More recently, \cite{blenderbot3} enhance a language model through continuous learning from user feedback and dialogue history. InstructGPT~\citep{ouyang2022training} initially trains GPT-3 using supervised instruction tuning and subsequently fine-tunes it via reinforcement learning from human feedback (RLHF), where the reward model is trained on annotated human preference data. This reward model, in turn, serves as a user simulator which can provide feedback for model's predictions. 
~\cite{ramamurthy2023is} demonstrate that RLHF is more data- and parameter-efficient than supervised methods when a learned reward model provides signals for an RL method, not to mention that preference data is easier to collect than ground-truth data. 
~\cite{fernandes2023bridging} and ~\cite{wang2021putting} provide a comprehensive survey on the topic of ``learning from feedback''. We refer the readers to these two surveys for more information.

\paragraph{Regulating via Human Configuration.} 
The two interaction schemes previously discussed involve engagement with simulated or real humans through prompts or feedback. Regulation through human configuration, on the other hand, relies on users to customize and configure the language model system according to their needs. 
This customization can include adjustments to the system's structure, hyperparameters, decoding strategy, and more. 
Although it may not be the most flexible method, it is one of the simplest ways to facilitate interaction between the user and the system. 

For example, ~\cite{ai-chains} predefine a set of LLM primitive operations, such as ``ideation'', ``split points'', ``compose points'', etc.; each operation being controlled by a specific prompt template. Users can customize the usage and chaining schemes of different operations to meet a set of given requirements. Similarly, PromptChainer~\citep{wu2022promptchainer} is an interactive interface designed to facilitate data transformation between different steps of a chain. It also offers debugging capabilities at various levels of granularity, enabling users to create their own LM chains. 
Users can also configure some hyperparameters to control the performance of LLMs. This includes, but is not limited to, temperature (which controls the stochasticity of the output), the maximum number of tokens to generate, and ``top-p'' controlling diversity via nucleus sampling~\citep{nucleus-sampling}\footnote{\url{https://platform.openai.com/playground}}. 
~\cite{chatgpt4robot} have proposed the concept of ''\textit{user-on-the-loop}'', implying that users can configure the LM-robot interaction with human instructions, ensuring that the process and results of the interaction are centered around the user's needs. 

\paragraph{Learning from Human Simulation.} 
In many cases, training or deploying language models with real users is impractical, prompting the development of various user simulators to emulate user behavior and preferences. For instance, \cite{ouyang2022training} initially rank generated responses with real annotators based on their preferences and then train a reward model—initialized from GPT-3~\citep{brown2020language}—on this preference data to serve as a user preference simulator. 
~\cite{preference-transformer} propose a method to simulate human preference by utilizing a transformer model that captures important events and temporal dependencies within segments of human decision trajectories. Additionally, this approach relies on a weighted sum of non-Markovian rewards. 
~\cite{faltings2023interactive} simulate user editing suggestions through BertScore-based~\citep{DBLP:conf/iclr/ZhangKWWA20} token-wise similarity scores and dynamic programming to compute an alignment between a draft and a target. 
~\cite{lynch2022interactive} collect numerous language-annotated trajectories, with the policy trained using behavioral cloning on the dataset. These collected trajectories can also be viewed as a user simulator.

The design of a user simulator is critical for the successful training and evaluation of language models. For example, to accurately replicate the behavior and preferences of real users when developing a generic dialogue system, it is vital to collect a diverse and extensive range of user data for training the simulator. This allows it to encompass the full spectrum of user preferences and behaviors. Moreover, when developing language models for rapidly changing application scenarios, it is essential to continually update and refine the simulator to adapt to shifts in user demographics and their evolving preferences.

\subsection{KB-in-the-loop}
\label{Sec:KB-in-the-loop}
KB-in-the-loop NLP has two main approaches: one focuses on utilizing external knowledge sources to augment language models during inference time~\citep{knn-lm, guu2020realm, lewis2020retrievalaugmented,cheng-etal-2021-guiding,izacard2022atlas, menick2022teaching, retro, nakano2021webgpt, knowledge_f1, wang2023shall, lewis2021paq, chen2022augmenting}, while the other aims to employ external knowledge to enhance language model training, resulting in better language representations~\citep{lu2021kelm, kbert, zhang2019ernie, sun2019ernie, fevry2020entities, sun2021ernie, xiongpretrained, relational-memory, knowledge-survey}. Interacting with KB during training can help improve the model's representation to incorporate more factual knowledge. In contrast, interacting with KB during inference can assist the language model in generating more accurate, contextually relevant, and informed responses by dynamically leveraging external knowledge sources based on the specific input or query at hand. 

\begin{wrapfigure}[13]{r}{6cm}
\centering
\includegraphics[width=1 \linewidth]{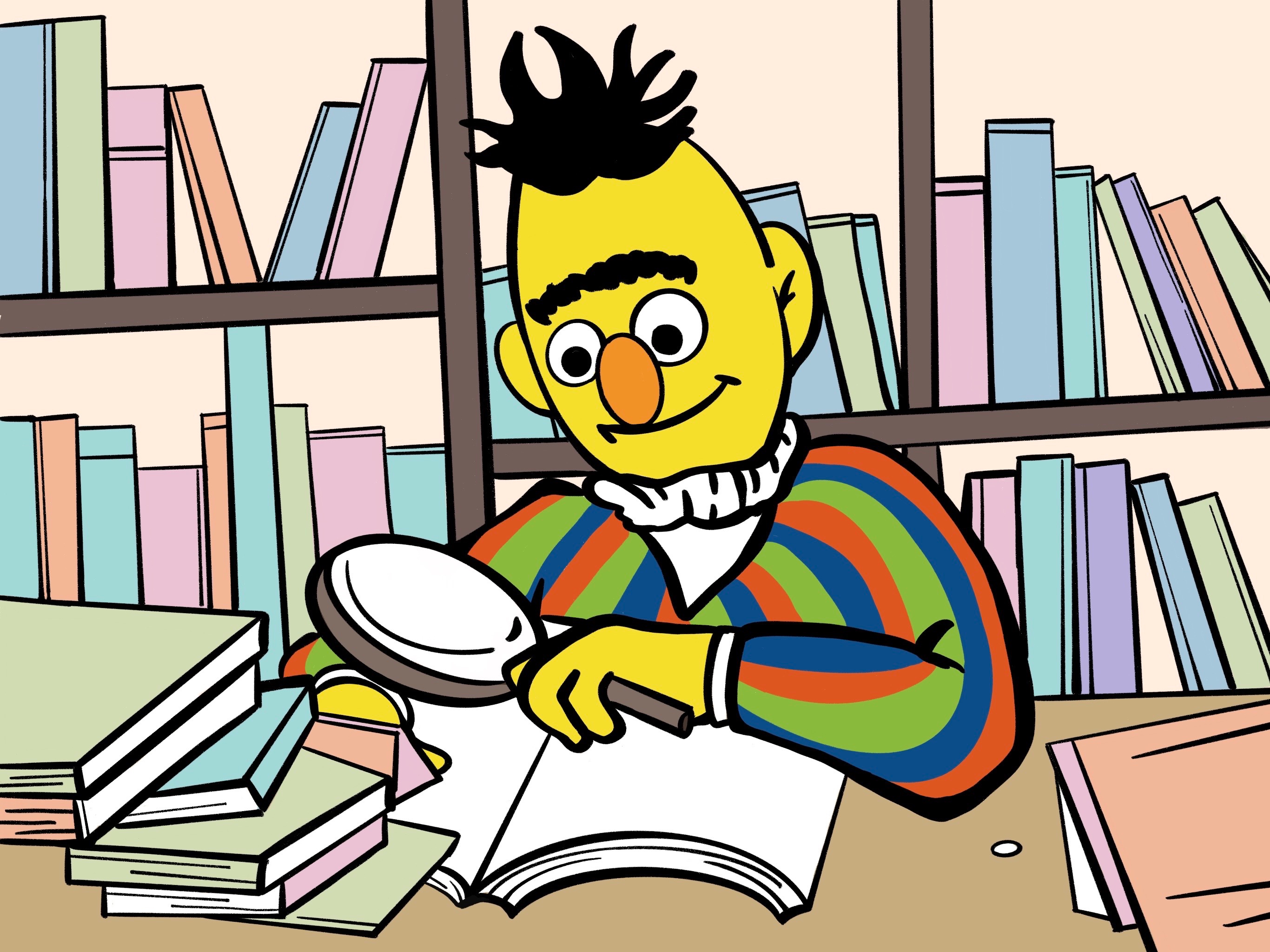}
\caption{KB-in-the-loop.}
\label{fig:kb-wrap}
\end{wrapfigure}

In this following sections, we will discuss knowledge sources and knowledge retrieval. As for knowledge integration, we refer the readers to §\ref{fusion} for more details.

\paragraph{Knowledge Sources.}
Knowledge sources are normally categorized into the following types:

(1) Corpus Knowledge: Typically, corpus knowledge is stored in an offline collection from a specific corpus, which the language model accesses to enhance its generation capabilities. Common examples of corpus knowledge include the Wikipedia Corpus~\citep{wikipedia}, WikiData Corpus~\citep{wikidata}, Freebase Corpus~\citep{freebase}, PubMed Corpus\footnote{\url{https://pubmed.ncbi.nlm.nih.gov/}}, and CommonCrawl Corpus\footnote{\url{https://commoncrawl.org/}}, among others. Most previous research has focused on corpus knowledge due to its controllability and efficiency. Retrieval-Augmented Language Models~\citep{guu2020realm,lewis2020retrievalaugmented,retro,knowledge_f1,izacard2022atlas} have been proposed to develop language models capable of utilizing external knowledge bases for more grounded generation~\citep{knowledge-survey, retrieval-survey}. To further improve interpretability, subsequent studies~\citep{lewis2021paq,chen2022augmenting,wu2022efficient} have suggested using extracted Question-Answer pairs as the corpus for more fine-grained knowledge triple grounding. Recently, there has been growing interest in incorporating citations to enhance grounding in language models, as demonstrated by GopherCite~\citep{menick2022teaching}. Another line of work, including KELM~\citep{lu2021kelm}, ERNIE~\citep{sun2019ernie,sun2020ernie,sun2021ernie}, and others~\citep{xiongpretrained,fevry2020entities}, primarily employs recognized entities as the foundation for integrating knowledge graph information into neural representations.

(2) Internet Knowledge: One challenge associated with corpus knowledge is its limited coverage and the need for specialized retrieval training. A potential solution involves offloading the retrieval process to search engines and adapting them to find the desired content. The Internet-augmented language model~\citep{lazaridou2022internet} was first introduced to answer open-domain questions by grounding responses in search results from the Internet. This approach has since been demonstrated to effectively answer time-sensitive questions~\citep{kasai2022realtime}. The Internet has also been employed for post-hoc attribution~\citep{gao2022attributed}. WebGPT~\citep{nakano2021webgpt} proposes powering language models with a web browser, which searches the web before generating knowledgeable or factual text. MineDojo~\citep{fan2022minedojo} equips a video-language model with Internet-scale knowledge to tackle diverse tasks within a \textit{Minecraft} environment. ToolFormer~\citep{toolformer} similarly integrates a search engine into the tool-use adaptation of language models. ReAct~\citep{yao2022react} suggests leveraging the Internet to augment reasoning capabilities in black-box large language models.

While corpus knowledge and internet knowledge are both valuable resources that language models can utilize to enhance their capabilities, they inherently differ in terms of controllability and coverage. 
Corpus knowledge is pre-collected and stored offline in a controlled setting, making it easy to access and integrate into a language model. However, it is limited by the information within the corpus and may not be up-to-date or comprehensive. 
In contrast, internet knowledge offers a vast and diverse pool of constantly updated information, providing more comprehensive coverage. However, controlling and curating internet knowledge is challenging, as the information obtained from the internet may be more noisy or even more misleading. 
Additionally, it is worth noting that there are other miscellaneous types of knowledge sources, such as visual knowledge~\citep{valm}, rule-based knowledge~\citep{rulebert, ptr-prompt, kadapter, liu2022mind}, implicit knowledge~\citep{lmkb}, database knowledge~\citep{li2023llm}, and documentation knowledge~\citep{zhou2022docprompting}. These can be categorized into either corpus knowledge or internet knowledge, depending on their nature.

\paragraph{Knowledge Retrieval\label{para:retrieval}.} 
Enhancing language models with knowledge requires careful consideration of knowledge quality. Knowledge quality is primarily affected by issues such as knowledge missing and knowledge noise~\citep{ye2022ontologyenhanced}. 
Knowledge missing can be mitigated by changing or extending the knowledge source to provide more comprehensive information. 
To tackle knowledge noise, an intuitive approach is to filter out the noisy information. ~\cite{kbert} and ~\cite{ye2022ontologyenhanced} propose addressing this issue by using a visibility matrix that functions on the attention scores between the knowledge and input. This helps in better integration of high-quality knowledge into the language model. 
Despite the success of these methods, improving knowledge retrieval remains the most critical aspect of addressing these challenges. This is because improving knowledge retrieval directly impacts the precision and recall of knowledge that is selected and integrated into the language model, leading to better overall performance. 
There are overall three methods for knowledge retrieval: 

(1) \textbf{Sparse Retrieval}: 
In this approach, knowledge is retrieved based on lexical matches between words or phrases in the input text and a knowledge source or the similarity between sparse representations.
For example,
ToolFormer~\citep{toolformer} employs BM25~\citep{bm25} as a metric to retrieve knowledge from Wikipedia.
DrQA~\citep{drqa} retrieves documents using TF-IDF vectors.
RepoCoder~\citep{zhang2023repocoder} incorporates the Jaccard index~\citep{jaccord} as one of its retrieval metrics.
Moreover, researchers explore on utilizing the sparse representations from pre-trained language model compound with the lexical matching methods~\citep{dai2020deepct, zhao2020sparta, formal2021splade}.

(2) \textbf{Dense Retrieval}: 
Dense retrieval approach retrieves knowledge based on the meaning of the input text rather than merely matching exact words or phrases. The meaning is typically encoded by a learned retriever. A dual encoder or cross encoder can be used as the retriever.
For example, 
REALM~\citep{guu2020realm} employs a latent knowledge retriever that is trained in an unsupervised manner to extract relevant information and context from a vast corpus during both the training and inference stages. 
Retro~\citep{retro} retrieves chunks from an external knowledge base using a dual encoder and integrates the retrieved chunks into language models through cross attention. 
\cite{cai-etal-2021-neural} jointly train a translation memory retriever and neural machine translation model. 
RepoCoder\citep{zhang2023repocoder} also employs an embedding model to compute the cosine similarity between input and knowledge. 
Atlas~\citep{izacard2022atlas} retrieves knowledge with Contriever~\citep{Contriever}, a dense dual encoder-based retriever trained via contrastive learning. 
\cite{izacarddistilling} and RePlug~\citep{shi2023replug} propose distilling knowledge from a reader to a retriever model, which requires very few annotated training data.

(3) \textbf{Generative Retrieval} \label{generative-retriever}: 
Instead of retrieving knowledge through matching, a generative retriever directly produces the document id or content as knowledge. As such, the generative retriever, typically in the form of a language model, can be considered a type of knowledge base, which is also known as implicit knowledge~\citep{lmkb, 10.1162/tacl_a_00324, liu-etal-2022-generated}. 
For example, 
DSI~\citep{dsi} encodes numerous documents with their ids into the language model's parameters. During inference, the model generates the id of the most relevant document. 
~\cite{sun2022recitationaugmented} propose augmenting language models with recitations, which are relevant knowledgeable content generated by language models. 
~\cite{generate-rather-than-retrieve} prompt a large language model to generate diverse contextual documents based on a given question and then read the generated documents to produce a final answer, where the in-context demonstrations for the LLM prompting are sampled from a clustered document pool. 
It is worth noting that knowledge distillation may also fall within this category. 
For example, 
~\cite{llmteacher} allow large language models to serve as teachers, distilling their reasoning skills into smaller language models. The knowledgeable large language model can be viewed as a generative retriever-like knowledge base for the smaller language models.

(4) \textbf{Reinforcement Learning}: 
Knowledge retrieval can also be formulated as a reinforcement learning problem. 
For example, 
WebGPT~\citep{nakano2021webgpt} learns to retrieve and select documents via behavior cloning (BC) and reinforcement learning from human feedback (RLHF). 
~\cite{zhang2022active} formulate the example retrieval problem as a Markov Decision Process (MDP) and propose a reinforcement learning (RL) method to select examples.

\subsection{Model/Tool-in-the-loop\label{model-tool-in-loop}}

\begin{wrapfigure}[20]{r}{7cm}
\centering
\includegraphics[width=1 \linewidth]{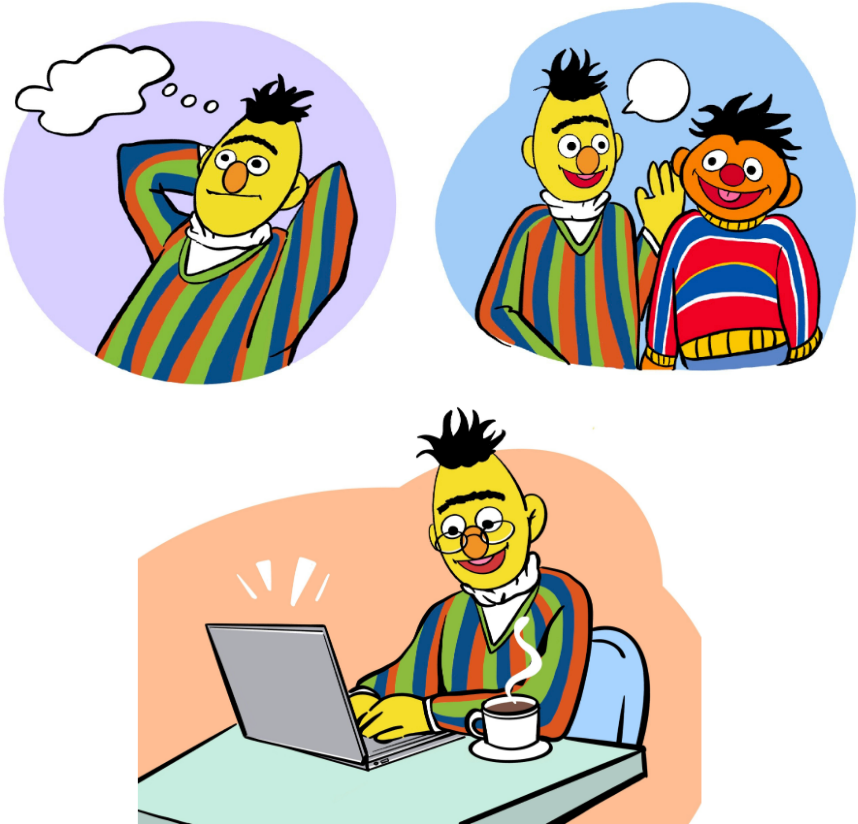}
\caption{Model/Tool-in-the-loop. }
\label{fig:model-tool-wrap}
\end{wrapfigure}

Addressing complex tasks often necessitates the implementation of strategic methodologies that can simplify the process. 
One such effective strategy is the explicit decomposition of the task into modularized subtasks and then solve these subtasks step by step~\citep{Wei2022ChainOT, zhou2022least, lm-cascades, reasoning-survey}. 
Alternatively, another strategy involves the implicit decomposition of the task through the division of labor among multiple language model agents. This approach enables a natural and adaptive breakdown of the work, as each agent assumes a specific role in the larger task~\citep{zeng2022socratic, bara-etal-2021-mindcraft, workspace}. 
The procedure of task decomposition not only allows subtask modularization, but also enables subtask composition. 
Furthermore, by breaking the task into multiple steps, specific steps can be allocated to certain expert models or external tools, such as those specializing in arithmetic computation, web search, counting, and more~\citep{toolformer, yao2022react, qin2023tool}. 
Inspired by ~\citep{augmented-lm,yao2022react}, 
there are primarily three fundamental operations involved in decomposing and solving these subtasks: 

\begin{enumerate}
    \item \textbf{Thinking}: The model engages in self-interaction to reason and decompose complex problems into modularized subtasks~\citep{yao2022react, augmented-lm, bubeck2023sparks-AGI, lm-cascades}; 
    \item \textbf{Acting}: The model calls tools or models to solve these intermediate subtasks, which may result in effects on the external world~\citep{yao2022react, augmented-lm, qin2023tool}; 
    \item \textbf{Collaborating}: Multiple models with distinct roles or division of labor communicate and cooperate with each other to achieve a common goal or simulate human social behaviors~\citep{using-language, premack_woodruff_1978, bara-etal-2021-mindcraft, kosinski2023theory, park2023generative, camel}. 
\end{enumerate}

\paragraph{Thinking.} 
For example, consider the question, ``\textit{What is the biggest animal in Africa?}'', which can be decomposed into a chain of three subtasks: ``\textit{What animals are in Africa?}'' → ``\textit{Which of these animals are large?}'' → ``\textit{Which of these is the largest?}'' These three subtasks form a prompt chain (c.f., §\ref{prompt-chaining}), allowing for the individual solving of each subtask by a single LM, multiple LMs, or even tools. That is, through the process of thinking, the overall task can be decomposed into multiple subtasks that can be efficiently tackled through interactions among language models or tools in a chained manner. 

The preliminary instantiation of such a cognitive process is \textbf{Chain-of-Thought (CoT)}~\citep{Wei2022ChainOT}, which seeks to elicit multi-hop complex reasoning capabilities from large language models using a cascading mechanism~\citep{lm-cascades}. 
Instead of directly producing the answer, multiple thoughts (i.e., reasoning steps) are generated beforehand~\citep{Wei2022ChainOT, wang2022self, zhou2022least, self-ask}. 
Thus, CoT decomposes the task into two sub-tasks: \textit{thought generation → answer generation}. 
However, typical CoT involves solving these subtasks in a single model run~\citep{Wei2022ChainOT} without an interaction mechanism. 

Derivative works of CoT have shown an increasing tendency to utilize a self-interaction loop that involves iteratively calling the same language model to solve different subtasks~\citep{zhou2022least, ite-prompting, self-ask, yao2022react}, also known as \textbf{multi-stage CoT}~\citep{icl_survey, reasoning-survey}. 
Furthermore, some other derivative works share similar principles with CoT or multi-stage CoT but employ \textbf{different training strategies}, such as bootstrapping~\citep{zelikman2022star} (as discussed in ~\S\ref{semi-supervised}). 
Some works go beyond the subtask of \textit{thought generation} and \textbf{introduce new subtasks}, including \textit{thought verification}~\citep{self-verification}, \textit{fact selection and inference}~\citep{creswell2022selectioninference}, and \textit{self-refinement and self-feedback}~\citep{madaan2023selfrefine}, among others. 
Indeed, all of these works can be seen as instantiations of the thinking cognitive process. They employ a self-interaction mechanism, wherein a single language model is utilized iteratively to decompose tasks into subtasks, and effectively solve these subtasks.

\paragraph{Acting.\label{sec-acting}} 
Different from the process of thinking, acting involves the interaction of the LM with external entities, such as other LMs and tools. Since different models or tools can possess specific expertise, the LM can invoke these external entities to perform specific subtasks when the task is decomposed into subtasks. 
For example, 
\textit{thought verification} can be accomplished using a discriminative model~\citep{see-think-confirm}, and \textit{fact selection} may utilize a retriever model~\citep{guu2020realm}. 
External tools such as calculators~\citep{math-verifier, toolformer}, simulators~\citep{pnas-simulation, liu2022mind}, search engines~\citep{yao2022react, nakano2021webgpt}, code interpreters and executors~\citep{ni2023lever, gao2022pal, chen2022program}, and other APIs~\citep{parisi2022talm, yao2022react, toolformer, thoppilan2022lamda, blenderbot3, augmented-lm, liang2023taskmatrixai, visual-chatgpt, qin2023tool} can also be incorporated into the loop to tackle subtasks that language models typically encounter difficulties with. 
Generally, tasks emphasizing faithfulness and exactitude (e.g., real facts, complex mathematical operations) and tasks beyond the LM training corpus (e.g., up-to-date information, low-resource languages, awareness of time, image generation) are better solved using external tools than LMs~\citep{hallucination, maynez2020faithfulness, patel-etal-2021-nlp, komeili-etal-2022-internet, lin2022fewshot, time-aware, toolformer, augmented-lm, liang2023taskmatrixai, visual-chatgpt, qin2023tool}.

For example, 
ToolFormer~\citep{toolformer} enhances language models with tool-use capabilities by retraining on a tool-use prompted corpus and involving tools such as calculators, calendars, search engines, question-answering systems, and translation systems. 
ART~\citep{art-tool} begins by selecting demonstrations from a task library that involve multi-step reasoning and tool usage. These demonstrations serve as prompts for the frozen LLM to generate intermediate reasoning steps in the form of executable programs. 
ReAct~\citep{yao2022react} combines both chain-of-thought reasoning and task-specific tool-use actions to improve the interactive decision-making capabilities of language models. 
TaskMatrix.AI~\citep{liang2023taskmatrixai} presents a vision for a new AI ecosystem built on tool-use APIs, proposing an architecture composed of an API platform, API selector, multimodal conversational foundation model, API-based action executor, and integrating RLHF and feedback to API developers to optimize the system. This architecture benefits from its ability to perform digital and physical tasks, its API repository for diverse task experts, its lifelong learning ability, and improved interpretability. 
HuggingGPT~\citep{shen2023hugginggpt} and OpenAGI~\cite{ge2023openagi} use ChatGPT as a task controller, planning tasks into multiple subtasks that can be solved by models (tools) selected from the HuggingFace platform\footnote{\url{https://huggingface.co/}}.

Moreover, acting can have a tangible impact on the external world through tool-use~\citep{augmented-lm}, also referred to as Tool-Oriented Learning~\citep{qin2023tool}. 
For instance, 
ChatGPT Plugins\footnote{\url{https://openai.com/blog/chatgpt-plugins}} empower LLMs to directly utilize tools for tasks such as travel bookings, grocery shopping, and restaurant reservations, among others. 
LM-Nav~\citep{shah2022lmnav} leverages a visual navigation model (VNM) to execute the actions planned by the LLM, enabling real-world robotic navigation. 
In these cases, the overall task is still decomposed into subtasks, but some of which are connected with the external world. 
By employing specific models or tools to address these subtasks, tangible effects can be realized in the environment. 
Readers can refer to ~\S\ref{Sec:Env-in-the-loop} for additional information related to the interaction between the language model and the environment.

\paragraph{Collaborating\label{collaborating}.} 
Most of the aforementioned research relies on manual task decomposition. 
Although some existing works propose automatic task decomposition through distant supervision~\citep{min2019multihop, DBLP:conf/naacl/TalmorB18, perez-etal-2020-unsupervised} or in-context learning~\citep{zhou2022least, self-ask, DBLP:journals/corr/abs-2210-02406, DBLP:conf/emnlp/DuaG0G22, augmented-lm}, explicit task decomposition is not always straightforward. 
On the one hand, it requires human expertise or extensive manual effort. 
On the other hand, in certain cases, different language model agents may share a common goal that is difficult to explicitly decompose~\citep{the-dynamics-1998, DBLP:conf/iclr/LazaridouPB17, camel, bara-etal-2021-mindcraft}. 
In such scenarios, task decomposition or division of labor may emerge implicitly as different agents with specialized skills assume different roles within the task and interact with one another~\citep{using-language, premack_woodruff_1978, workspace, li-zhou-2020-connecting, liu2022stateful, liu2022coordinating, bara-etal-2021-mindcraft, kosinski2023theory, camel}. 
For example, in \textit{MineCraft}, agents with distinct yet complementary recipe skills can communicate and collaborate to synthesize a material, where the specialized agents may automatically discover a potential division of labor~\citep{bara-etal-2021-mindcraft}. 
To the best of our knowledge, we can categorize collaboration-based approaches into three clusters: 

(1) \textbf{Closed-Loop Interaction} refers to a collaborative process where multiple agents interact with each other in a feedback loop~\citep{freedman2019responsive, ahn2022can, zeng2022socratic, huang2022inner, dasgupta2023collaborating, see-think-confirm}. 
In the context of control theory, a closed-loop controller uses feedback to control states or outputs from a dynamical system\footnote{\url{https://en.wikipedia.org/wiki/Control_theory}}. 
Generally, closed-loop controllers are preferred over open-loop controllers as they offer greater adaptability and robustness in changing or uncertain environments. 
Likewise, closed-loop interaction between language model agents is more effective and robust compared to open-loop interaction~\citep{huang2022language, lynch2022interactive}, making it a primary paradigm for collaboration-based methods. 
For example, 
Socratic Models~\citep{zeng2022socratic} and Inner Monologue~\citep{huang2022inner} enable language models to collaborate with vision-language models, audio-language models, or humans to conduct egocentric perception and robotic manipulation tasks, respectively. The language-based closed-loop feedback is incorporated into LLM planning, significantly improving instruction completion abilities~\citep{huang2022inner}. 
Planner-Actor-Reporter~\citep{dasgupta2023collaborating} uses an LLM (Planner) to generate instructions for a separate RL agent (Actor) to execute in an embodied environment. The state of the environment is reported back to the Planner (via the Reporter) to refine instructions and complete the feedback loop. 
Note that closed-loop interaction is highly applicable in Environment-in-the-loop scenarios, where closed-loop feedback from the environments can be transferred via a model connected to the environment~\citep{huang2022inner, zeng2022socratic}.

(2) \textbf{Theory of Mind} in language models has garnered growing attention in the research community~\citep{premack_woodruff_1978, rabinowitz2018machine, zhu2021fewshot, bara-etal-2021-mindcraft, kosinski2023theory, liu2023computational}. 
According to ~\cite{kosinski2023theory}, ``\textit{Theory of Mind (ToM), or the ability to attribute unobservable mental states to others, is central to human social interactions, communication, empathy, self-consciousness, and morality.}''. ~\cite{kosinski2023theory} demonstrates that large language models, like ChatGPT, can successfully tackle 93\% of ToM tasks. 
This finding suggests that ToM-like capabilities may have naturally emerged in large language models. 
In line with this, 
MindCraft\citep{bara-etal-2021-mindcraft} assigns different material composition tables (sub-skills) to two dialogue agents, enabling them to cooperate and complete the material composition task through mutual communication. 
\cite{zhu2021fewshot} provide a speaker and listener formulation of ToM, where the speaker should model the listener's beliefs (i.e., action possibilities over some instruction candidates). 
These ToM mechanisms are beneficial for collaborative tasks~\citep{liu2023computational}.

(3) \textbf{Communicative Agent} perceives language models as agents~\citep{jacob-lm-agent} and delves into the study of multi-agent communication~\citep{DBLP:conf/iclr/LazaridouPB17}. 
In addition to Theory of Mind, multi-agent communication also investigates the scenarios of 
referential game~\citep{DBLP:conf/iclr/LazaridouPB17}, 
language acquisition~\citep{liu2023computational}, 
language emergence~\citep{machine-language}, and 
role playing~\citep{camel}, implying an effort towards LLM society~\citep{camel}. 
For example, 
~\cite{machine-language} enable two communicative agents, a speaker and a listener, to learn to play a \textit{Speak, Guess and Draw} game and automatically derive an interaction interface between them, which is so-called machine language. 
Camel~\citep{camel} proposes a role-playing framework that involves two cooperative agents, an AI user and an AI assistant. The two language models are prompted with a shared task specifier prompt and different role assignment prompts, which is referred to as \textit{Inception Prompting}. With the condition of Inception Prompting, they communicate with each other without any additional human instruction to solve the specified task. 
Generative Agents~\citep{park2023generative} introduces a novel architecture that extends a LLM to enable believable simulations of human behavior in an interactive sandbox environment, demonstrating the agents' ability to autonomously plan and exhibit individual and social behaviors. 
~\cite{communicative-learning}'s formalism even views existing machine learning paradigms such as passive learning and active learning, as communicative learning, which is in line with ~\cite{hoeve2021towards}'s interactive language modeling. 
In these paradigms, the language model agents are grouped into teachers and students, where the students learn from the teachers through interaction. They frame learning as a communicative and collaborative process.

\subsection{Environment-in-the-loop}
\label{Sec:Env-in-the-loop}

\begin{wrapfigure}[16]{r}{7cm}
\centering
\vspace{-5pt}
\includegraphics[width=1 \linewidth]{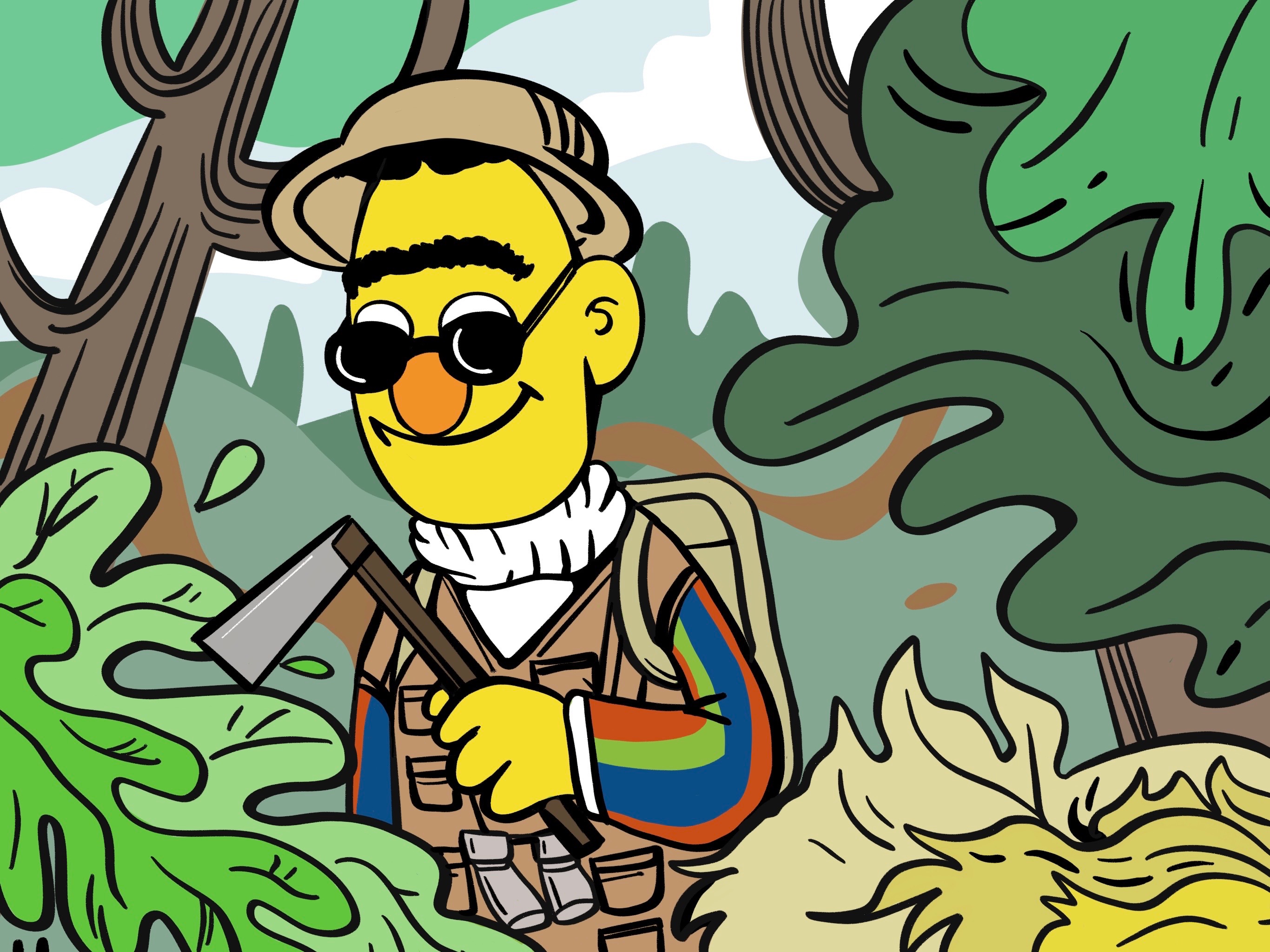}
\caption{Environment-in-the-loop. 
}
\label{fig:env-wrap}
\end{wrapfigure}

A new trend within the NLP community is to harness the power of LMs to address embodied tasks such as robot manipulation, autonomous driving, and egocentric perception, among others~\citep{ahn2022can, huang2022inner, code-as-policies, chen2022openvocabulary, shah2022lmnav, zeng2022socratic, dasgupta2023collaborating, carta2023grounding, huang2023grounded}. 
In these scenarios, the environment is integrated into an interactive loop with language models. 
The aim of environment-in-the-loop NLP is language grounding, which is to represent language with meaning reference to environments and experiences~\citep{experience-ground}. 
It has been argued that only if LMs are put into interaction with real-world or virtual environments can they learn a truly grounded representation of language~\citep{experience-ground}. 
During this interaction, the environment assumes the responsibility of furnishing the LM with low-level observations, rewards, and state transitions. Simultaneously, the LM is tasked with generating solutions for environmental tasks, including reasoning, planning, and decision-making~\citep{experience-ground, li2022pre, yang2023foundation}.

\begin{figure}[h]
    \centering
        \includegraphics[width=.6\textwidth]{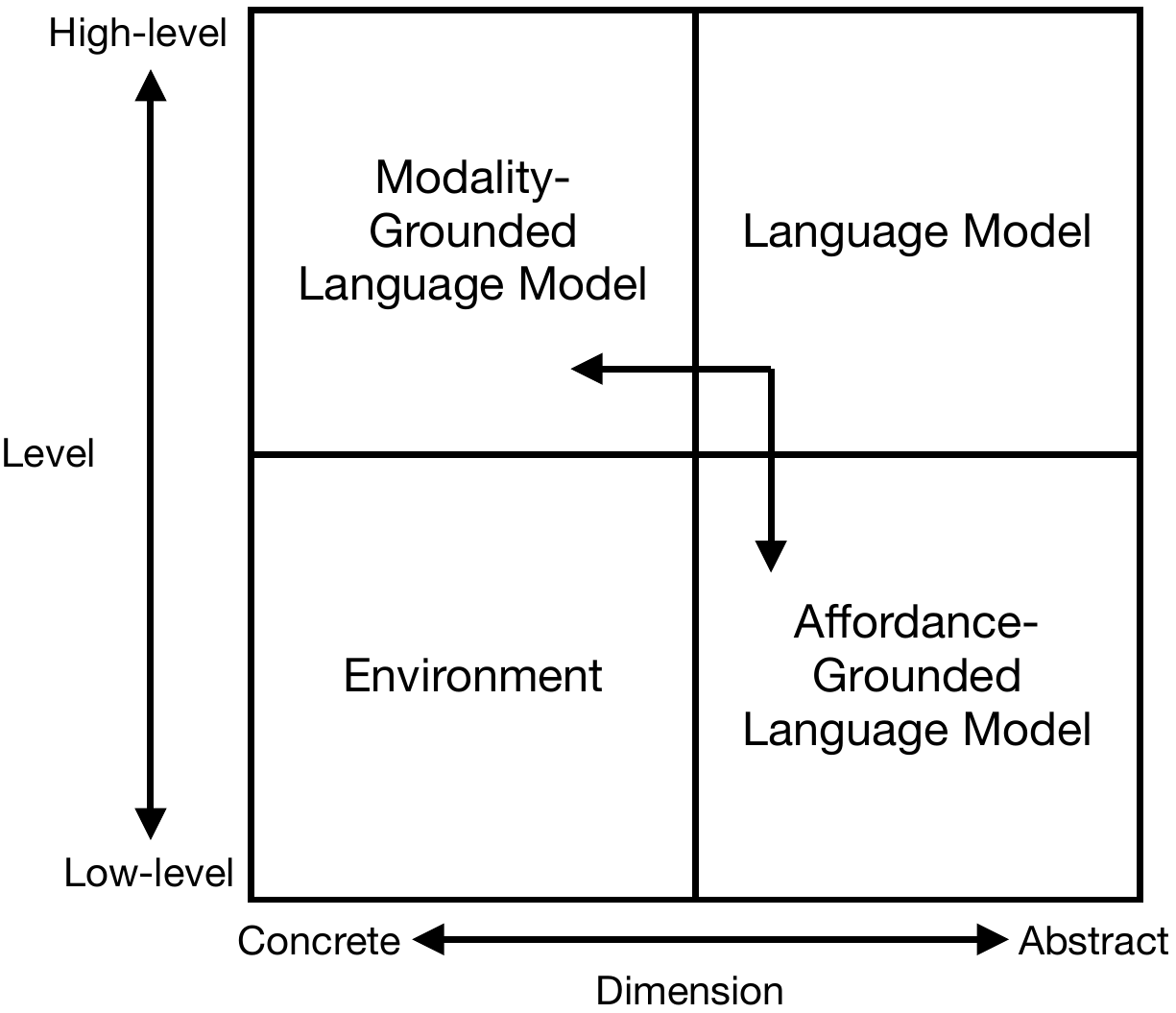}
        \caption{
            Two directions for language grounding. 
            A third direction for language grounding may be social interaction~\citep{experience-ground, social-neuro-ai, DBLP:conf/iclr/LazaridouPB17, liu2023computational} 
            which is not illustrated in this figure but 
            we have discussed it partly in §\ref{collaborating}.
        }
        \label{fig:grounding}
\end{figure}

We define two dimensions for language grounding, as shown in Figure \ref{fig:grounding}. 
The horizontal axis spans from the \textit{concrete} end to the \textit{abstract} end. 
The term \textit{concrete} refers to models that capture high-dimensional data of the world, such as images, audio, and other similar sensory inputs. 
On the other hand, the term \textit{abstract} pertains to models that capture low-dimensional data, such as language, code, or other symbolic representations. 
Compared to a more concrete representation, abstract or bottle-necked representation brings stronger generalization and reasoning ability~\citep{kawaguchi2017generalization, trauble2023discrete, liu2021discretevalued}.

The vertical axis ranges from the \textit{low-level} end to the \textit{high-level} end, where 
\textit{low-level} means a more direct and embodied interaction with the environment, such as perception or manipulation, while 
\textit{high-level} means a more indirect and conceptual interaction with the environment, such as reasoning, planning, and decision-making. 
This axis can reflect the degree of the model's contextual and situational understanding of the environment.

Generally, the environment can be the real world or virtual world simulated by programs such as MuJoCo~\citep{todorov2012mujoco} and MineCraft\footnote{\url{https://www.minecraft.net}}. 
Hence, the environment is in the bottom-left quadrant in Figure \ref{fig:grounding} with a concrete representation of data and low-level interaction processes. While the language model is in the top-right quadrant in Figure \ref{fig:grounding} with an abstract representation of data and high-level interaction processes. 
This discrepancy makes it necessary to ground language models for LM-env interaction. 
There are mainly two directions: \textbf{modality grounding} and \textbf{affordance grounding}.

(1) \textbf{Modality Grounding}~\citep{beinborn-etal-2018-multimodal} aims to move the language model from the abstract quadrant to the concrete quadrant. It is intuitive to incorporate information in image, audio or other modalities into it. In this way, language models can capture more complete observations from the environment.

(2) \textbf{Affordance Grounding}~\citep{ahn2022can} strives to transition language models from the high-level quadrant to the low-level quadrant. The goal is to align the outputs of language models with the contextual scene, ensuring that the generated text correspond to the surrounding environment rather than being detached from it.

It is worth noting that these two goals are not independent processes, and often form a synergy towards the environment. Moreover, other additional requirements such as preference and safety are also possible directions~\citep{huang2023grounded}, which may further involve human in the loop. 

\paragraph{Modality Grounding.} 
Modality-Grounded Language Model (MGLM) is designed to allow language models to process data of more modalities such as vision and audio. 
In the context of visual grounding (i.e., vision-language pre-trained model), for example, there are three ways: 
(1) Dual-Tower modeling which trains different encoders for different modalities~\citep{tan2019lxmert, vilbert, clip, xu2022bridge, albef, yu2022coca, zeng2022x2vlm}; 
(2) Single-Tower modeling using the concatenation of multimodal data to train a single model~\citep{su2019vlbert, li2019visualbert, uniter, oscar, one-for-all, gato, rt1, fromage, driess2023palme, chen2022pali, beit3, wang2022foundation, diao2023write, mllm}; 
(3) Interaction between frozen pre-trained vision and language models~\citep{zeng2022socratic, huang2022inner, alayrac2022flamingo, li2023blip2, visual-chatgpt, chatgpt-blip-2, see-think-confirm}. 
These methods involve the utilization of visual information during both the training and inference stages of a language model. By incorporating visual signals, these approaches enable a visually grounded representation of language. This enhancement in representation facilitates improved interaction efficiency between the language model and the environment, as it allows for increased information throughput. 

For example, 
WebShop~\citep{yao2022webshop} and Interactive Language~\citep{lynch2022interactive} use ResNet~\citep{resnet} and a Transformer model~\citep{transformer} to process visual and linguistic data respectively, and input the fused representations into another Transformer to generate action outputs; 
VIMA~\citep{jiang2022vima} and Gato~\citep{gato} use one single model to simultaneously process the concatenated multimodal data and predict actions; 
Socratic Models~\citep{zeng2022socratic}, Inner Monologue~\citep{huang2022inner}, and LM-Nav~\citep{shah2022lmnav} use multimodal language models to convert visual inputs into language captions or phrases and use LLMs for planning, reasoning and question-answering in order to perform embodied tasks. 
ViperGPT~\citep{suris2023vipergpt} equips the LLM with an API for various perceptual and knowledge modules, along with a Python interpreter, enabling the LLM to generate executable code for visual reasoning tasks. 

Another goal of Modality Grounding is to preserve as much high-level knowledge as possible in the language model to ensure that the model is still able to effectively perform tasks such as commonsense reasoning, planning, question answering, code generation, etc. 
These capabilities become more pronounced and complex as the size of the model increases, known as emergent abilities~\citep{kaplan2020scaling, wei2022emergent}. 
These capabilities serve as one of the primary purposes of leveraging language models for embodied tasks. 
An illustrative example of these capabilities is demonstrated in the context of completing long-horizon navigation tasks. In such tasks, the effective planning of instructions by the LLM is crucial~\citep{shah2022lmnav}. 

\paragraph{Affordance Grounding.} 
However, in general, in order to make MGLM knowledge-rich, the model needs to be pre-trained with a large amount of data from open domains, which may result in outputs that are too diverse and therefore do not match the conditions in the real environment~\citep{ahn2022can, chen2022openvocabulary, huang2023grounded}. 
Therefore, some low-level information from the environment is needed to be incorporated into language models, which is referred to as Affordance Grounding~\citep{ahn2022can}.

According to ~\cite{theoryaffordance} and ~\cite{khetarpal2020theory}: ``\textit{Affordances describe the fact that certain states enable an agent to do certain actions, in the context of embodied agents.}''. 
Likewise, according to ~\cite{ahn2022can}: ``\textit{The learned affordance functions (Can) provide a world-grounding to determine what is possible to execute upon the plan}''. 
However, ~\cite{chen2022openvocabulary} argues that ~\cite{ahn2022can}'s falls short in providing affordance grounding at the scene-scale, thus limiting the ability to reason about the potential actions a robot can perform within a given environment. 
Hence, following this thought, 
there are mainly two requirements for an affordance grounded lanugage model (AGLM): 
(1) \textbf{scene-scale perception}, and 
(2) \textbf{possible action}, \textbf{conditioned on the language-based instructions}. 
For example, 
when considering a smart home environment and asking the agent to ``\textit{turn off the lights in the living room.}'', 
scene-scale perception aims to make the agent aware of all (or only) the existing and relevant objects, such as ``\textit{bedlamps}'' and ``\textit{droplights}''. 
Secondly, possible action tasks the agent to determine the executable actions on the objects that can complete the instructions, such as ``\textit{\textbf{press} the switches.}''

For example, 
SayCan~\citep{ahn2022can} leverages large language models to generate a list of object-action proposals (i.e., task grounding) which are then scored by a value function connected to the environment (i.e., world-grounding). 
Similarly, ~\cite{chen2022openvocabulary} first construct a language queryable scene representation, NLMap,  through pre-exploration of a robotic agent and then use the a LLM to generate a list of relevant objects to be filtered and located. The object presence and location are finally used for LLM planning. 
~\cite{abramson2022improving} train an agent via behavioral cloning on the interactions of paired human players. They then collect human feedback on the learned agent to train a reward model, which is finally used to post-train the agent. That is, they achieve affordance grounding via behavioral cloning and RLHF. 
Code as Policies~\citep{code-as-policies} enables a language model to generate executable code directly. The generated codes can be executed with a python interpreter for affordance verification~\citep{ni2023lever}. 
LM-Nav~\citep{shah2022lmnav} converts the planning results of the language model to image form and then uses a Visual Navigation Model to convert them into executable instructions (i.e., action+distance). 
Grounded Decoding~\citep{huang2023grounded} integrates the high-level semantic understanding of LLMs with the reality-based practicalities of grounded models, enabling the generation of action sequences that are both knowledge-informed and feasible in embodied agent tasks like robotics. 
\cite{wake2023chatgpt} provide numerous examples of utilizing ChatGPT for generating executable action sequences to accomplish tasks assigned by users. 

Note that KB-in-the-loop, Model/Tool-in-the-loop, or Human-in-the-loop approaches can also be employed for modality grounding or affordance grounding~\citep{huang2022language, yao2022react, zeng2022socratic, huang2022inner, lynch2022interactive, abramson2022improving}. In these approaches, external objects or entities undertake these functions, such as utilizing humans to describe the visual scene for modality grounding~\citep{huang2022inner}.

\section{Interaction Interface\label{interface}}

In this section, we discuss the interfaces through which language models communicate with interactive objects. The interfaces include three types of languages: natural language, formal language, and machine language, as well as two special interfaces: edits and shared memory.

\subsection{Natural Language} \label{natural-language}

\begin{wrapfigure}[15]{r}{7cm}
\centering
\vspace{-10pt}
  \includegraphics[width=1 \linewidth]{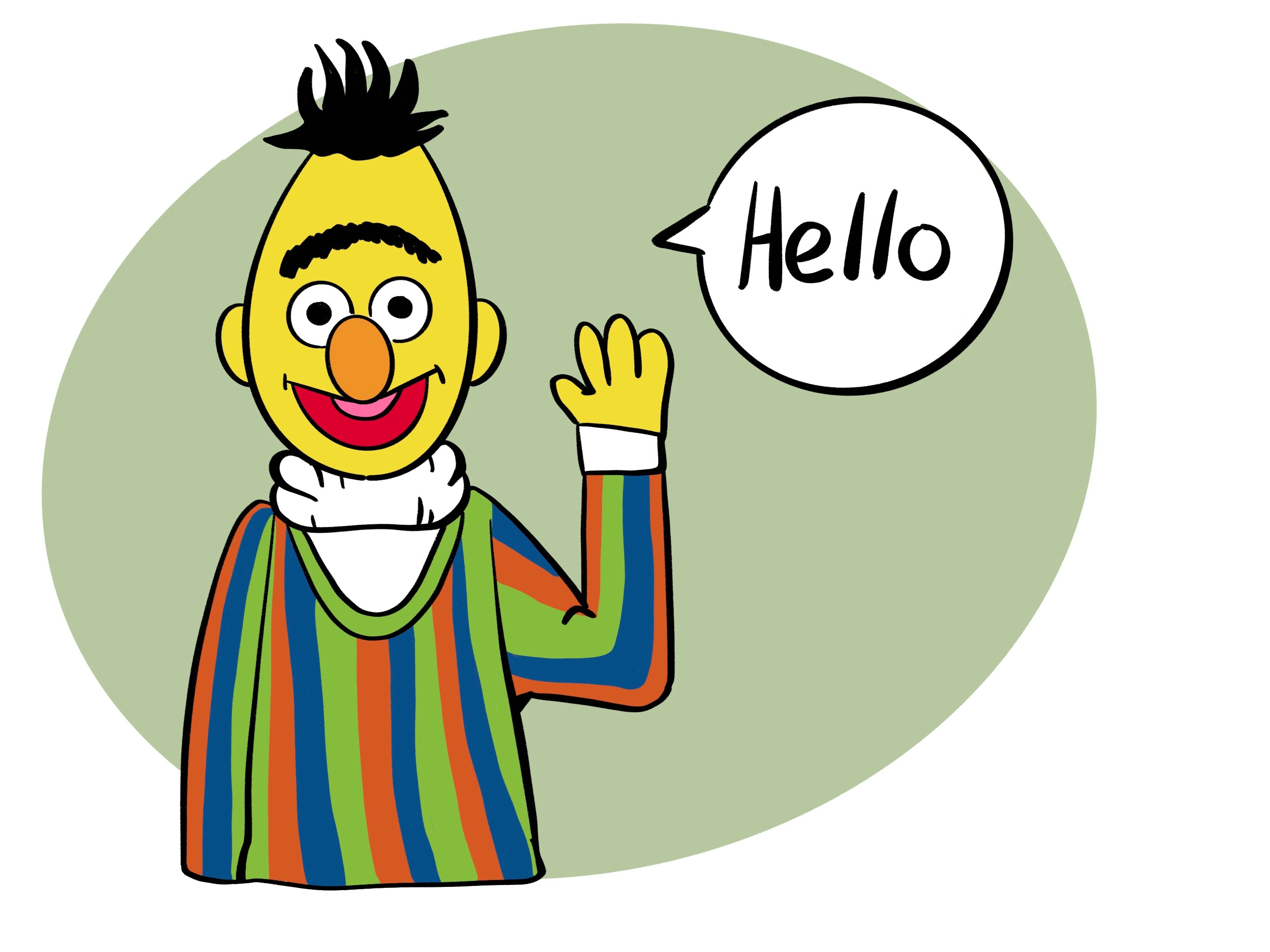}
        \caption{Interacting via Natural Language.}
        \label{fig:natural-language}
\end{wrapfigure}

Natural Language is the most common interaction interface. 
Communicating via this interface requires that the interactive objects can effectively understand and produce natural language. 
This interface is therefore commonly used in Model-in-the-loop~\citep{ai-chains, zeng2022socratic} and Human-in-the-loop~\citep{ouyang2022training, Evaluating_Human_Language_Model_Interaction}. 
Natural language interaction empowers users to express their needs with inherent expressiveness, enabling effective communication of their requirements without the need for specialized training. 
Additionally, this interaction interface facilitates a better understanding of the intermediate interaction process, leading to improved debuggability and interpretability of the interaction chain~\citep{ai-chains, Wei2022ChainOT, Evaluating_Human_Language_Model_Interaction}. 
Crucially, since LMs are primarily pre-trained on natural language, interacting with them through natural language instead of other language is the most effective way to activate and utilize the knowledge encoded in the LMs. 
This alignment between the LM training data and the interaction interface allows for optimal utilization of the knowledge contained within the LMs. 

However, interacting with a language model through natural language heavily relies on the organization and the utterance of the language, often necessitating intricate prompt engineering~\citep{liu-etal-2022-makes, gu2022robustness, calibrate-before-use, lu-etal-2022-fantastically, icl_survey, chen2022relation}. 
Organization of the language refers to the structure of a model's prompt, and can be categorized into \textbf{unstructural natural language} and \textbf{structural natural language}. 
Utterance, on the other hand, refers to the specific wording or language used to express a given prompt or query. 
Utterance is more flexible by nature and therefore difficult to determine an optimal one. 
Different utterances may produce different results as they differ from the activated pattern in the model parameters. 
Practically, suitable prompts can be discovered through manual or automatic search
~\citep{icl_survey, prompt-survey, wallace-etal-2019-universal, 10.1162/tacl_a_00324, li2021prefixtuning, zhang2022active, zhou2022large, liu-etal-2022-makes}. 
We refer the readers to §\ref{prompting} for more information. 

\paragraph{Unstructural Natural Language.} 
Unstructural natural language is a free-form text. 
When it serves as an output from the language model, it does not have specific categorization, and the content can be free-form responses such as answers to questions and textual feedback. 
When it serves as an input to the language model, in addition to the main input content, such as interaction messages and queries, it primarily takes three forms of auxiliary context: (1) few-shot examples, (2) task description, and (3) role assignment~\citep{naturalinstructions, supernaturalinstructions, camel}. 
Thereof, 
\begin{itemize}
    \item Input format example of few-shot prompting: 
    ``\textit{[Example-1]; [Example-2]; [Example-3]; [input]}'' e.g. ``\textit{sea otter → loutre de mer; plush girafe → girafe peluche; cheese →}'' for translation task. 
    \item Input format example of task description: ``\textit{[task description]: [input]}'' e.g. ``\textit{translate English to French: cheese →}''. 
    \item Input format example of role assignment: ``\textit{[role assignment]. [input]}'' e.g. ``\textit{Act as a python programmer: write codes to detect objects.}''\footnote{Role assignment can be considered a special type of task descriptions.}. 
\end{itemize}

For example, some recent work, including 
Natural Instructions~\citep{naturalinstructions} and Super Natural Instructions~\citep{supernaturalinstructions}, have built comprehensive collections of tasks and their corresponding instructions in natural language. 
Interactive Language~\citep{lynch2022interactive} enables humans to provide real-time instructions for the multimodal language model based on the current state of a given environment for robotic manipulation. 
Camel~\citep{camel} defines an inception prompt that comprises a task specifier prompt and two role assignment prompts, namely the assistant system prompt and the user system prompt, which are utilized for role-playing tasks. 

\paragraph{Structural Natural Language.} 
Structural natural language usually imposes explicit constraints on the text in terms of content or formatting. 
Such constraints can be imposed on either the input~\citep{zhong-etal-2022-proqa} or output~\citep{ahn2022can} of language models. 
For example, 
\cite{drissi2018hierarchical, sun2020summarize, yang2022re3} define the overall structure of the generated article via an Outline or Plan (e.g., ``\textit{(1) Introduction, (2) Related Work, (3) Method, (4) Experimental Results, ...}`` or a storyline). 
\cite{ahn2022can} and ~\cite{chen2022openvocabulary} unify the format of a generated text via a Template (e.g., ``\textit{pick up [object]}'') to facilitate parsing of the action and the object to be acted upon. 
ProQA~\citep{zhong-etal-2022-proqa} employs a prompt-based input schema that is designed in a structured manner, e.g., ``\textit{[Format]: <Extractive QA>; [Task]: <SQuAD>; [Domain]: <Wikipedia>; [Question]: In what Country is Normandy located? [Passage]: ...}''. This schema allows for efficient modeling of knowledge generalization across all QA tasks, while also preserving task-specific knowledge tailored to each individual QA task. Note that although ProQA incorporates certain soft prompts in its input schema (c.f., ~\S\ref{machine-language}), the main body of its instance still consists of natural language.

While unstructured natural language is a widely used interface for interaction due to its flexibility, simplicity, and readability, it suffers from certain drawbacks, including ambiguity, lack of coherence and parsability. 
Although these challenges can be partially addressed by employing structural natural language, all forms of natural language are inherently limited by its subjectivity and variability. 

\subsection{Formal Language}

To further unlock the benefits of structural language, such as unambiguity, coherence, and parsability, and to mitigate the inherent limitations of natural language mentioned above, formal language emerges as another important interaction interface. 
According to Wikipedia\footnote{\url{https://en.wikipedia.org/wiki/Formal_language}}, 
``\textit{a formal language consists of words whose letters are taken from an alphabet and are well-formed according to a specific set of rules.}'' 
Formal language is utilized in various domains such as mathematics, logic, linguistics, computer science, as well as other fields where precise and unambiguous communication is essential. 
Here are some examples of formal languages:

\begin{enumerate}
    \item Programming Languages: examples include C, Java, Python, and many others. These programming languages are used to write scripts or commands that computers can execute~\citep{cheng2022binding, code-as-policies, chen2022program, toolformer, art-tool}.
    \item Query Languages: examples include SQL and XQuery, which are used to retrieve and manipulate data stored in databases~\citep{cheng2022binding, li2023llm}.
    \item Mathematical Expressions: examples include boolean algebra, first-order logic, and equations. They are used to describe mathematical concepts and relationships~\citep{wu2022autoformalization, lu2021intergps, han2022folio}.
    \item Formal Grammars: examples include context-free grammars, regular grammars, recursive grammars, etc\footnote{\url{https://en.wikipedia.org/wiki/Formal_grammar}}. They are used to describe the syntactic structure of natural language~\citep{bai2021syntaxbert, sachan2020syntax, kadapter}.
    \item Others: for example, knowledge triples~\citep{relational-memory, sun2021ernie}, and regular expressions (regex, \citeauthor{locascio2016neural}).
\end{enumerate}

\begin{wrapfigure}[16]{r}{7cm}
\centering
 \includegraphics[width=1 \linewidth]{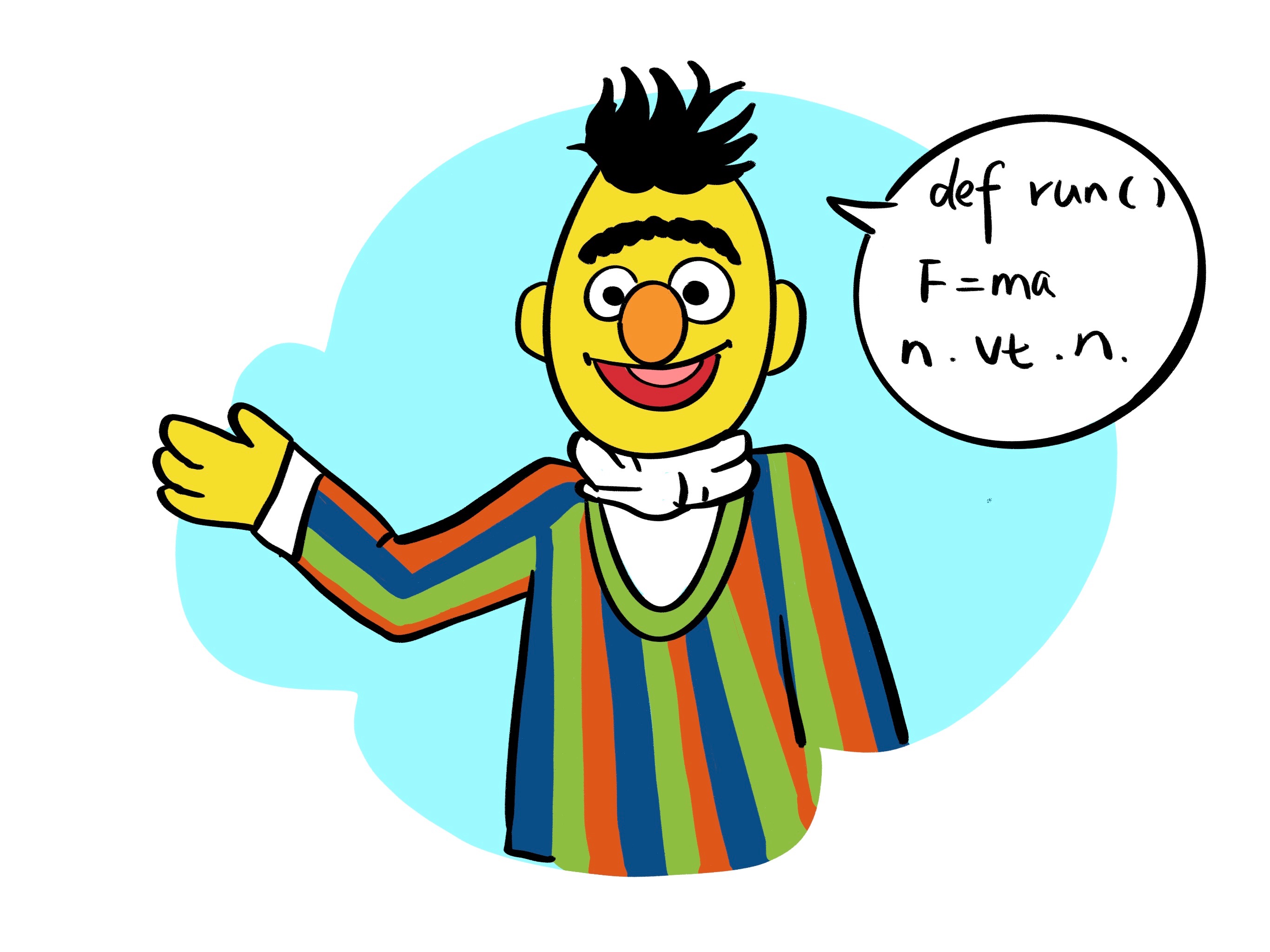}
        \caption{Interacting via Formal Language.}
        \label{fig:formal-language}
\end{wrapfigure}

The interactive objects that use formal language as an interaction interface usually include knowledge bases~\citep{relational-memory, li2023llm, cheng2022binding}, environments~\citep{code-as-policies}, and models/tools~\citep{liu2022mind, wu2022autoformalization, lu2021intergps, draft-sketch-prove, liang2023taskmatrixai}. 
For example, 
Mind's-Eye~\citep{liu2022mind} uses a text-to-code language model to generate rendering codes for the physical simulation engine. 
\cite{draft-sketch-prove} involve a three-step approach to creating mathematical proofs. This approach includes formulating an initial informal proof, converting it into a formal sketch, and then employing a standard prover to prove the conjectures. 
This allows for the automated transformation of informal mathematical issues into fully formalized proofs using natural and mathematical languages. 
Binder~\citep{cheng2022binding} first parses its input into programs (Python, SQL, etc.) given the questions and knowledge bases, and then executes them to get the results. 
K-Adapter~\citep{kadapter} incorporates linguistic knowledge into PLMs through the use of adapters~\citep{pmlr-v97-houlsby19a}, exemplifying the application of formal grammars as an interaction interface. 
In some specific cases, other interactive objects may also use formal language. 
For example, human developers can interact with a code-based language model~\citep{codex} using formal language. 
\cite{lahiri2022interactive} create an interactive framework to refine user intents through test case generations and user feedback. 

Compared to natural language, formal language offers distinctive advantages as an interaction interface, including: 
(1) It brings about precision and clarity, eradicating the ambiguity often associated with natural language. 
(2) Its structured syntax and rules make it directly parsable and easily interpretable by programs, enabling more efficient and accurate interaction with tools, for example. 
(3) It facilitates complex reasoning and logic-based operations more effectively, as codes or mathematical proofs are data formats that encompass a series of logical reasoning steps, which may provide opportunities to enhance models' reasoning abilities~\citep{yaofu-notion-blog, suris2023vipergpt}. 
However, the use of formal language may have certain limitations, including: 
(1) Limited accessibility: It often requires specialized knowledge or training for proper understanding and usage. And it relies on LMs specifically trained with formal language. 
(2) High sensitivity: E.g., even small errors in codes can render them non-executable. 
(3) Lack of expressiveness: It is unable to convey ideas in a nuanced and flexible manner.

\subsection{Edits\label{chap-edits}}

Text editing aims to reconstruct the textual source input to the target one by applying a set of edits, such as deletion, insertion, and substitution~\citep{malmi-etal-2022-text}. The motivation behind text editing is the recognition that source and target texts often share significant similarities in various monolingual tasks.
Instead of reproducing the source words~\citep{gu-etal-2016-incorporating,see-etal-2017-get,zhao-etal-2019-improving,Panthaplackel_Allamanis_Brockschmidt_2021}, text editing models reduce such copying to predicting a single keep operation. Also, edits are often facilitated with rich metadata about language editing, including the inserted or deleted spans and the word order.

\begin{wrapfigure}[15]{r}{7cm}
\centering
 \includegraphics[width=1 \linewidth]{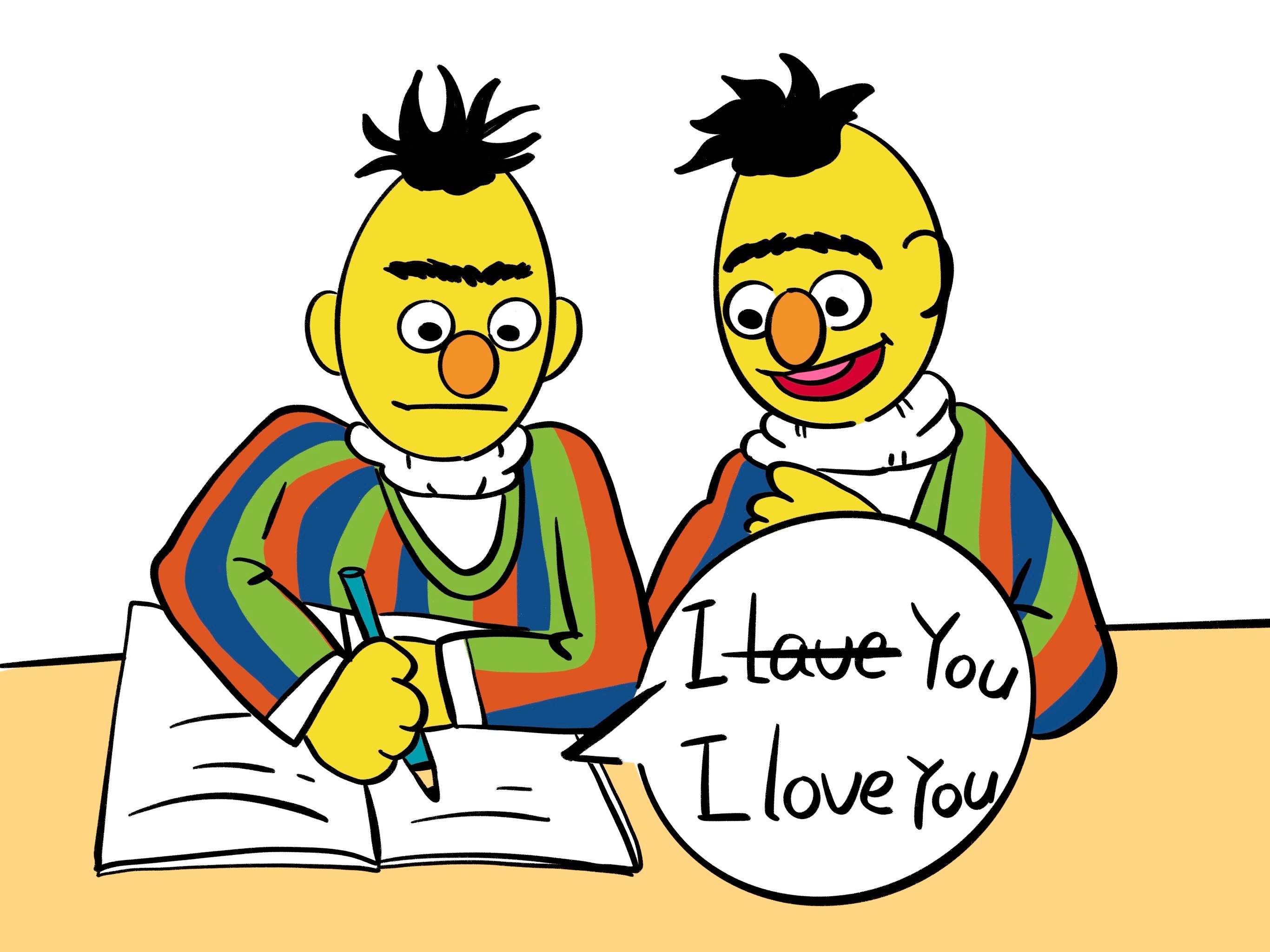}
        \caption{Interacting via Edits.}
        \label{fig:edits}
\end{wrapfigure}

Learning from the editing of textual data is gaining increasing attention, given its success in code pre-training~\citep{zhang2022coditt5}, image editing~\citep{ravi2023preditor}, drug design~\citep{corso2022diffdock}, and other areas.
Previous cognition-related research has proven that mechanical editing operations require less cognitive effort compared to correcting transfer errors, which have no references to the source text version~\citep{lacruz2014cognitive}, and that iterative editing procedure plays an important role in improving students' writing abilities~\citep{vardi2012impact, gollins2016framework}.
Moreover, these editing-related cognitive phenomena have shined upon various NLP topics. 
By addressing some limitations of the dominant sequence-to-sequence approaches~\citep{sutskever2014sequence}, such as a relatively high computational requirement~\citep{mallinson-etal-2020-felix}, text editing has found its wide array of applications~\citep{malmi-etal-2019-encode,mallinson-etal-2020-felix,stahlberg-kumar-2020-seq2edits} such as automatic post-editing~\citep{berard-etal-2017-lig,xu-etal-2022-bilingual}, data-to-text generation~\citep{kasner-dusek-2020-data}, grammatical error correction~\citep{awasthi-etal-2019-parallel, zhou-etal-2020-improving-grammatical, hinson-etal-2020-heterogeneous,omelianchuk-etal-2020-gector}, punctuation restoration~\citep{che2016punctuation,kim2019deep,alam-etal-2020-punctuation,shi21_interspeech}, sentence simplification~\citep{dong-etal-2019-editnts,agrawal-etal-2021-non}, human value alignment~\citep{liu2022second, zhang2023corgi}, style transfer~\citep{reid-zhong-2021-lewis}, and sequence-to-sequence pre-training~\citep{zhou-etal-2021-improving-sequence}.

Similar to the pattern of repeated revisions made by humans to a manuscript until it is finalized, a complete process of text editing can be decomposed to multiple iterative rounds of editing, rather than one-pass edit
~\citep{ge-etal-2018-fluency,levenshtein-transformer,pmlr-v97-stern19a,kumar-etal-2020-iterative,shi-etal-2020-recurrent,shi-etal-2022-text,faltings2023interactive}. 
On account of this, edits can be treated as one kind of interaction interface. 
Typically, text editing can be conducted through interaction between the editing model and itself, with the outputs of the previous iterations as the input of the current one until the text is fully edited to be returned~\citep{schick2022peer,kasner-dusek-2020-data,kim-etal-2022-improving,madaan2023selfrefine}. 
Meanwhile, text editing can also be conducted through interaction among multiple different models or modules~\citep{narayan-gardent-2014-hybrid,mallinson-etal-2022-edit5, mallinson-etal-2020-felix,malmi-etal-2020-unsupervised}.
For example, an edit can be split into tasks, such as sequence tagging and masked language modeling, for models to cooperate. Specifically, a tagger first attaches an edit operation to each token. Afterwards, a masked language model fills in the placeholders for insertion and substitution operations to complete the edit~\citep{mallinson-etal-2020-felix,malmi-etal-2020-unsupervised}. 
Moreover, the participation of a code interpreter~\citep{dong-etal-2019-editnts,shi-etal-2020-recurrent}, environment~\citep{shi-etal-2022-text}, and user simulator~\citep{faltings2023interactive} can control the editing better and provide additional supervision signals.

Recent research proves that editing-based models can be expanded to various NLP downstream tasks by retrieving or generating prototypes, i.e., original text to be edited~\citep{kazemnejad-etal-2020-paraphrase, malmi-etal-2022-text}. Additionally, text editing models have shown impressive performance in low-resource settings and can get rid of the typical autoregressive mechanism, thus improving inference speed~\citep{mallinson-etal-2020-felix, awasthi-etal-2019-parallel}.
However, it is still under-explored how to automatically generate prototypes for general NLG tasks so as to expand the text editing paradigm to them~\citep{guu-etal-2018-generating}, which hinders the broad use of edits as an interaction interface.

\subsection{Machine Language\label{machine-language}}

In some cases, the communication language between the language model and interactive objects is not human-readable. 
This communication interface is referred to as Machine Language, 
as it can only be understood and processed by computers (e.g., models, tools). 
We can break down this type of interaction interface into two categories: 
\textbf{discrete machine language} and \textbf{continuous machine language}.

\newpage
\begin{wrapfigure}[15]{r}{7cm}
\centering
\vspace{-10pt}
 \includegraphics[width=1 \linewidth]{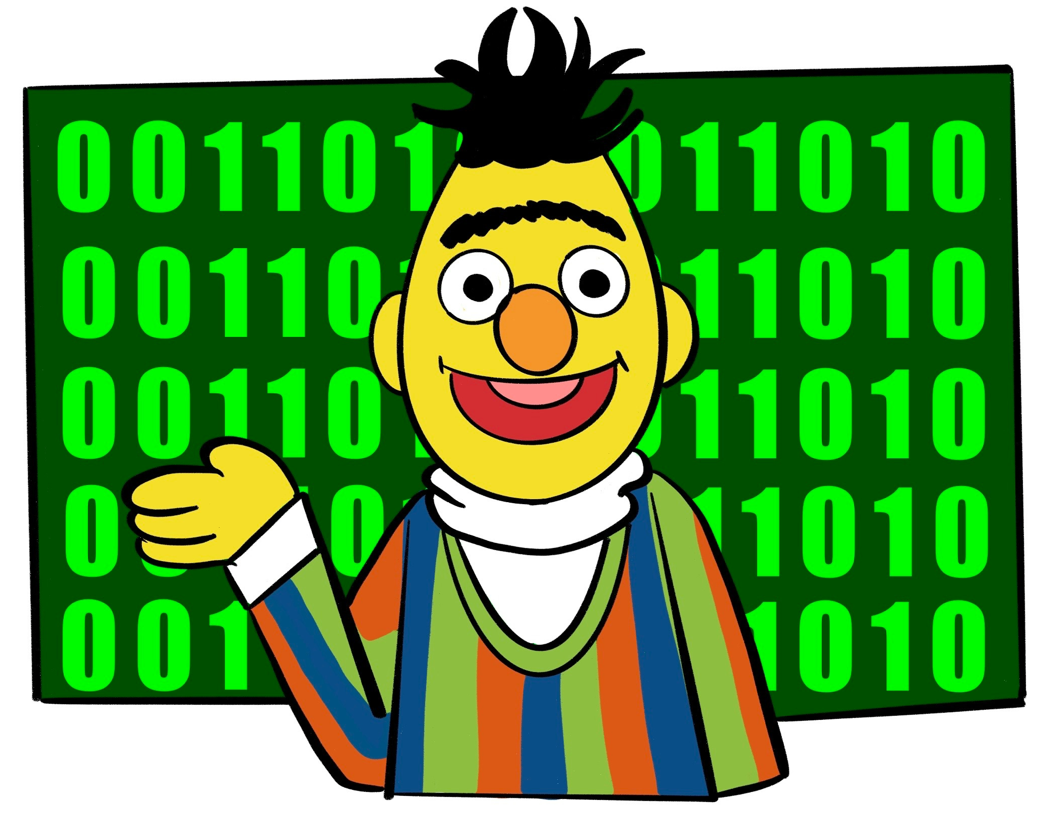}
        \caption{Interacting via Machine Language.}
        \label{fig:machine-language}
\end{wrapfigure}

\paragraph{Discrete Machine Language.} 
It refers to an interaction interface that is not readable by humans and quantized. 
For example, 
OFA~\citep{one-for-all} and BEiT-3~\citep{beit3} treat images as a form of  ``\textit{foreign language}''. 
That is, 
the sequence of image patches is obtained through image quantization and discretization techniques~\citep{vq-vae, vqgan, vit-vqgan, beit2}. This process allows the generation or understanding of an image token sequence that cannot be directly readable by humans but can be processed by models such as VQ-VAE~\citep{vq-vae} or VQGAN~\citep{vqgan, vit-vqgan}. 
Similarly, the hidden states inside the language model can also be discretized into discrete machine language in a similar manner. 
For example, 
~\cite{trauble2023discrete, liu2021discretevalued, machine-language} have demonstrated that discretized, human-unreadable hidden states can lead to better generalization and robustness.

\paragraph{Continuous Machine Language\label{continuous-machine-language}.} 
It refers to an interaction interface through which the language models communicate with interactive objects using continuous scalars or vectors in a dense space. 
For example, 
Flamingo~\citep{alayrac2022flamingo} encodes and re-samples images into dense and continuous vectors and then passes them to a language model via cross attention. 
BLIP-2~\citep{li2023blip2} encodes and maps images into numerous soft tokens
\footnote{Soft tokens, also known as soft prompts, refer to learnable parameters that are concatenated to the input prompt of a language model. Please refer to \cite{prompt-survey}.} 
and passes them to a language model as prefixes of text inputs. 

Note that metric signals, such as scalar rewards and ranking scores~\citep{christiano2017deep, ouyang2022training, ramamurthy2023is}, can also be regarded as a form of machine language employed by language models. In particular, if these signals belong to a discrete set of numbers (e.g., $\in \mathbb{Z}$), they can be classified as a form of discrete machine language. On the other hand, if they are represented as continuous values (e.g., $\in \mathbb{R}$), they can be classified as a form of continuous machine language.

\subsection{Shared Memory}
\begin{wrapfigure}[13]{r}{7cm}
\centering
\vspace{-20pt}
 \includegraphics[width=1 \linewidth]{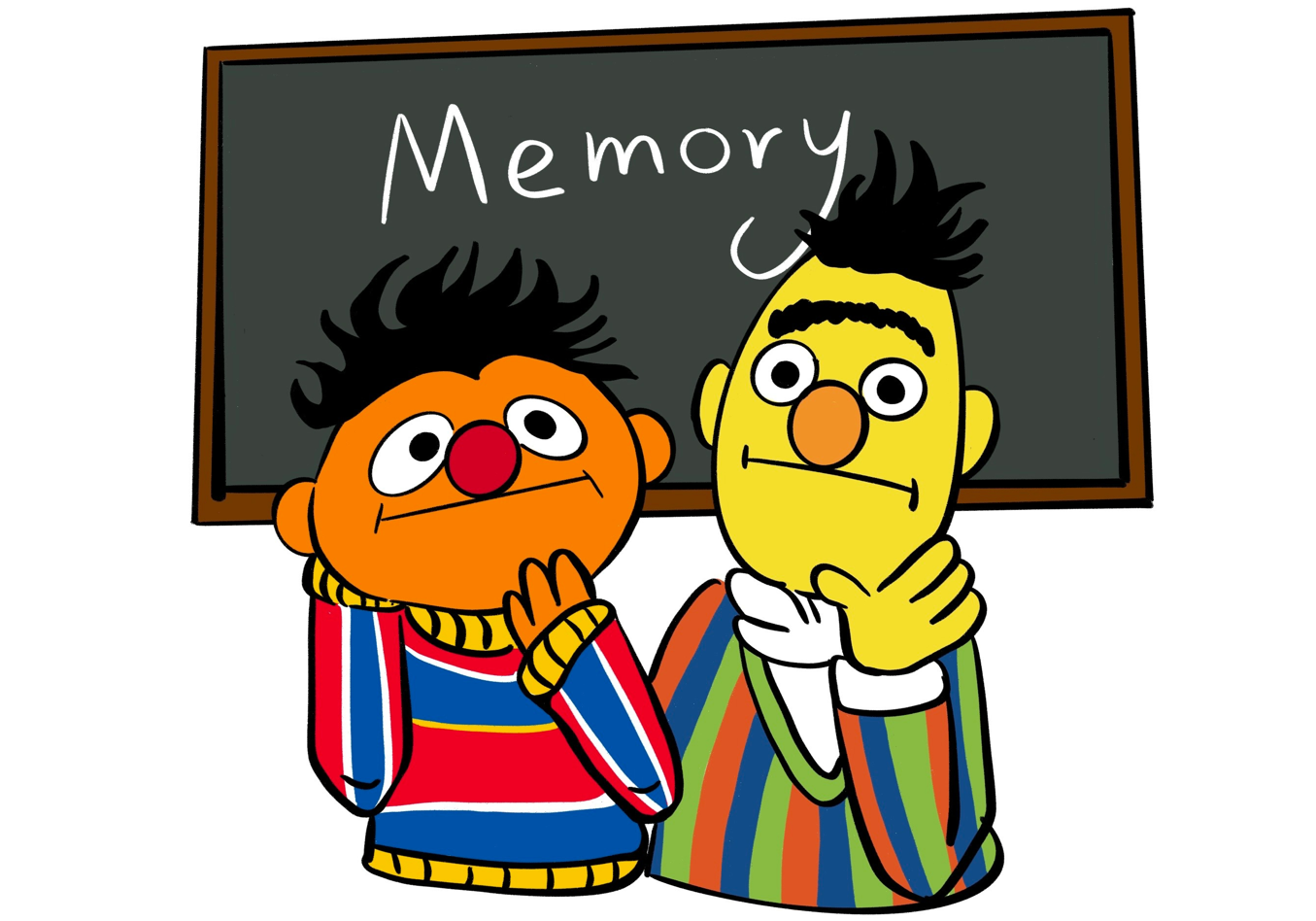}
        \caption{Interacting via Shared Memory.}
        \label{fig:memory}
\end{wrapfigure}

The interaction interfaces discussed earlier focus on direct communication between language models and interactive objects. However, there is also a form of indirect communication facilitated through shared information units, commonly referred to as shared memory~\citep{workspace, zeng2022socratic, madaan2022memory, dalvi2022towards, ryoo2022token}. 
That is, the message receiver does not directly receive the message from the sender, but instead retrieves it from a memory pool where the message has been pre-written by the sender. 
Depending on the form in which the message is stored and utilized, this type of interaction interface can be classified into two categories: \textbf{hard memory} and \textbf{soft memory}.

\paragraph{Hard Memory.} 
Hard memory often utilizes a human-readable history log to store shared information. 
For example, 
Socratic Models~\citep{zeng2022socratic} employs a communication mechanism where Vision-Language Models (VLM), Audio-Language Models (ALM), and Language Models (LM) interact through a history log written in natural language. This log records the complete history of states perceived by each model~\citep{zeng2022socratic}. 
MemPrompt~\citep{madaan2022memory} edits human prompts to GPT-3 with user feedback memory for better Human-LM interaction. 
\cite{dalvi2022towards} augment a question-answering model with a dynamic memory of user feedback for continual learning.

\paragraph{Soft Memory.} 
Soft memory typically employs queryable and human-unreadable memory slots to store shared information. 
These memory slots utilize a continuous machine language for efficient storage and retrieval of information. 
For example, 
Shared Global Workspace~\citep{workspace} stores the information from multiple modules in a shared sequence of working memory slots to facilitate inter-module communication and coordination. 
Token Turing Machines~\citep{ryoo2022token} use an additional memory unit to store historical state information to aid long-horizon robotic manipulation.

Utilizing memory as an indirect interaction interface provides several advantages over direct communication. 
It allows interactive objects to retrieve messages from earlier moments, enabling them to access past information. Memory enables the storage of a large volume of information, facilitating high-throughput communication. 
However, memory can become noised or outdated, 
leading to potential confusion or errors. 
Retrieval from memory can be time-consuming, impacting the efficiency of the interaction. 
It can also introduce unpredictability and uncertainty into the interaction. 
Therefore, careful design is crucial to ensure the effective and efficient utilization of memory. 

\section{Interaction Methods\label{methods}}

This section aims to explore the methodologies employed by language models for understanding and processing interaction messages.
We begin with a quick tour through the pre-trained language models (§\ref{plms}). 
Next, we divide interaction methods into five categories: 
prompting without model training (§\ref{prompting}), 
fine-tuning which involves updating models' parameters (§\ref{fine-tuning}), 
active learning (§\ref{active-learning}), 
reinforcement learning (§\ref{rl}) as well as imitation learning (§\ref{imitation-learning}). 
Finally, we propose to re-frame and formalize these methods in a unified manner, i.e., interaction message fusion (§\ref{fusion}). 

\subsection{Pre-trained Language Models\label{plms}}

\begingroup
\setlength{\tabcolsep}{2pt}

\begin{table}[!htp]\centering
\caption{Overview of PLMs.}\label{tab:plm}
\rowcolors{1}{lightgray!20}{}
\resizebox{\textwidth}{!}{%
\scriptsize
\begin{tabular}{lrrrrr}
\toprule
\rowcolor{gray!30}
\textbf{Model} & \textbf{Architecture} & \textbf{Strategy} & \textbf{\#Parameters} & \textbf{Characteristics} \\
\midrule
BERT~\citep{devlin2018bert} &Enc & MLM, SRC & \makecell[r]{Base110M/Large340M} & MLM, NSP \\
RoBERTa~\citep{liu2019roberta} &Enc & MLM & \makecell[r]{Base123M/Large354M} & \makecell[r]{Dynamic Mask, No NSP }\\
XLNet~\citep{yang2019xlnet}&Enc/Dec & CausalLM & \makecell[r]{Base110M/Large340M} &Permutation AR LM \\
SpanBERT~\citep{joshi2020spanbert} &Enc & MLM & \makecell[r]{Base110M/Large340M} &Span Mask \\
ERNIE~\citep{sun2019ernie} &Enc & MLM &Base110M & Entity Mask, Phrase Mask \\
ERNIE-2.0~\citep{sun2020ernie} &Enc & MLM, SRC & \makecell[r]{Base110M/Large340M} & \makecell[r]{Learning lexical, syntactic, and semantic\\ information across Multi-Tasks Learning} \\
ALBERT~\citep{lan2019albert} &Enc & MLM, SRC & \makecell[r]{Base12M/Large18M/\\XL60M/XXL235M}& \makecell[r]{Embedding Decomposeing, Parameters Share,\\SOP (sentence order prediction)} \\
DistilBERT~\citep{sanh2019distilbert} &Enc & MLM &66M & Teacher-Student, Dynamic Mask, No NSP Task \\
ELECTRA~\citep{clark2020electra} &Enc & MLM & \makecell[r]{Small14M/Base110M/\\Large335M} & \makecell[r]{Token Generator, Discriminator to predict\\ original or replaced} \\
SqueezeBERT~\citep{iandola2020squeezebert} &Enc & MLM, SRC &62M &Replace FC layers with Convolutions \\
GPT~\citep{gpt} &Dec & CausalLM &117M & Decoder-based Model \\
GPT-2~\citep{radford2019language} &Dec & CausalLM &1.5B &More parameters and data than GPT \\
BART~\citep{lewis2019bart} &Enc-Dec & \makecell[r]{Seq2Seq} &Base140M, Large406M &Arbitrary Noise \\
PEGASUS~\citep{zhang2020pegasus} &Enc-Dec & \makecell[r]{Seq2Seq} & Base223M, Large568M & GSG (gap-sentences generation) \\
UniLM~\citep{dong2019unified} &Enc/Dec & \makecell[r]{PrefixLM} &340M & \makecell[r]{Unified for Bidirectional, Unidirectional, \\and Seq2Seq LM} \\
\bottomrule
\end{tabular}
}
\end{table}
\endgroup

\begingroup
\setlength{\tabcolsep}{2pt}
\begin{table}[!htp]\centering
\caption{Overview of LLMs.}\label{tab:llm}
\rowcolors{1}{lightgray!20}{}
\resizebox{\textwidth}{!}{%
\scriptsize
\begin{tabular}{lrrrrr}
\rowcolor{gray!30}
\toprule
\textbf{Model} &\textbf{Architecture} &\textbf{Pre-training} &\textbf{\#Parameters} &\textbf{Characteristics} \\\midrule

T5~\citep{2020t5} &Enc-Dec & \makecell[r]{Seq2Seq} & \makecell[r]{Base220M/Small60M/\\Large770M/3B/11B}  & \makecell[r]{Unified NLP tasks with\\ the same input-output format} \\
mT5~\citep{xue2020mt5} &Enc-Dec & \makecell[r]{Seq2Seq} & \makecell[r]{Base580M/Small300M/\\Large1.2B/XL3.7B/\\XL13B} &Multilingual T5 \\
ExT5~\citep{aribandi2021ext5} &Enc-Dec & \makecell[r]{Seq2Seq} & Base220M/Large770M &T5 with Multi-Task Learning \\
FLAN-T5~\citep{chung2022scaling} &Enc-Dec & Seq2Seq &8B/62B/540B &Scaling and Instruction Fine-tuning T5 \\
ERNIE-3.0~\citep{sun2021ernie} &Enc-Dec & \makecell[r]{MLM,\\ CausalLM,\\ SRC} &10B & \makecell[r]{Multi-Task Learning, \\External Knowledge Enhanced} \\
ERNIE-3.0 Titan~\citep{wang2021ernie} &Enc-Dec & \makecell[r]{MLM,\\ CausalLM,\\ SRC} &260B &Large Scale of Ernie 3.0 \\
GPT-3~\citep{brown2020language} &Dec & CausalLM &175B &100X parameters compared with GPT-2 \\
PANGU-$\alpha$~\citep{zeng2021pangu} &Dec & CausalLM &2.6B/13B/200B & Query Layer to induce expected output \\
FLAN~\citep{flan} &Dec & CausalLM &137B & Instruct Tuning \\
Gopher~\citep{rae2021scaling} &Dec & CausalLM & \makecell[r]{44M/117M/417M/\\1.4B/7.1B/280B} &RMSNorm, RoPE \\
InstructGPT~\citep{ouyang2022training} &Dec & CausalLM &1.3B/6B/175B & Instruct, GPT, RLHF \\
PaLM~\citep{chowdhery2022palm} &Dec & CausalLM &8B/62B/540B & \makecell[r]{SwiGLU, Parallel Layer, \\Multi-Query Attention, \\Shared Input-Output Embeddings, No Bias} \\
UL2~\citep{tay2022unifying} & \makecell[r]{Dec,\\Enc-Dec} & \makecell[r]{CausalLM,\\ Seq2Seq} & 1B/20B & \makecell[r]{Unified Denoising Objectives for\\ both Enc-Dec and Dec Architecture} \\
PaLM-2~\citep{google2023palm2} & \makecell[r]{Dec,\\Enc-Dec} & \makecell[r]{CausalLM,\\ Seq2Seq} & 1.04B/3.35B/10.7B & \makecell[r]{Multi-lingual and Multi-domain Training Data,\\ More Efficient Model Architecture} \\
OPT~\citep{zhang2022opt} &Dec & CausalLM & \makecell[r]{125M/350M/1.3B/2.7B/\\6.7B/13B/30B/\\66B/175B} & Open Pre-trained Transformer \\
Galactica~\citep{GALACTICA} &Dec  & CausalLM & \makecell[r]{125M/1.3B/6.7B/\\30B/120B} & \makecell[r]{High-quaility Scientific Training Data, \\ Prompt Pre-training} \\
GLM-130B~\citep{zeng2022glm} &Enc-Dec & \makecell[r]{CausalLM,\\ MLM} & \makecell[r]{Base100M/Large340M/\\410M/515M} & \makecell[r]{2D Positional Encoding, \\Autoregressive Blank Infilling, \\Multi-Task Instruction Pre-Training} \\
Bloom~\citep{scao2022bloom} &Dec & CausalLM & \makecell[r]{560M/1.1B/1.7B/\\3B/7.1B/176B} & \makecell[r]{ALiBi Positional Embedding, \\Embedding LayerNorm} \\
FLAN-PaLM~\citep{chung2022scaling} &Dec & CausalLM & \makecell[r]{Base250M/Small80M/\\Large780M/XL3B/XXL11B} & Scaling and Instruction Fine-tuning PaLM \\
LLaMA~\citep{touvron2023llama} &Dec & CausalLM &6.7B/13B/33B/65B &Pre-normalization, SwiGLU, RoPE \\
\bottomrule
\end{tabular}
}
\end{table}
\endgroup
Pre-trained language models (PLMs), especially large language models (LLMs), have demonstrated their tremendous potential to serve as the cornerstone of advancing language intelligence. 
Transformer~\citep{transformer}, BERT~\citep{devlin2018bert}, GPT-3~\citep{brown2020language} and ChatGPT are recognized as four major milestones of utilizing pre-trained language models for various NLP tasks, which also frame the roadmap of AI development. 
PLM is usually based on Transformer and can be categorized along two dimensions: (1) architectures, (2) pre-training strategies~\citep{tay2022unifying}.

\paragraph{Architectures.} 
There are overall three types of architectures: 
(1) \textbf{encoder-only}, 
where the model takes input tokens and produces a fixed-dimensional representation of the input text~\citep{devlin2018bert, liu2019roberta, sun2019ernie}, 
(2) \textbf{encoder-decoder}, 
where the model first generates a fixed-dimensional representation of the input text with an encoder, and then autoregressively generates tokens based on this representation with a decoder~\citep{lewis2019bart, 2020t5}, and 
(3) \textbf{decoder-only}, 
where the model directly generates tokens in an autoregressive manner based on the input text as context, utilizing only a decoder~\citep{gpt, radford2019language, brown2020language}. 
The encoder-only architecture is especially well-suited for discriminative tasks, such as text classification~\citep{adhikari2019docbert}. 
On the other hand, the encoder-decoder architecture is particularly suitable for sequence-to-sequence tasks, such as machine translation~\citep{mbart}. 
Lastly, the decoder-only architecture is particularly well-suited for generative tasks, such as story generation~\citep{gpt-2-story-gen}.

\paragraph{Pre-training Strategies.} 
LMs typically employ self-supervised training objectives for pre-training, including: 
(1) \textbf{CausalLM} (causal language modeling), where the model predicts the next token based on the preceding tokens from left to right~\citep{gpt, radford2019language, brown2020language}. 
(2) \textbf{PrefixLM} (prefix language modeling), where the model predicts the next token using a bidirectionally encoded prefix as well as the previous tokens from left to right~\citep{dong2019unified}. 
(3) \textbf{MLM} (masked language modeling), where the model predicts the masked span of the input~\citep{devlin2018bert}. 
(4) \textbf{Seq2Seq} (sequence-to-sequence), where the model decodes the output from left to right based on the encoded input~\citep{lewis2019bart, 2020t5}. 
(5) \textbf{SRC} (sentence relationship capturing), which includes tasks such as Next Sentence Prediction~\citep{devlin2018bert} and Sentence Order Prediction~\citep{lan2019albert}, aimed at capturing relationships between sentences. 
Other pre-training objectives, such as Right-to-Left Language Modeling~\citep{dong2019unified} and Permutation Language Modeling~\citep{yang2019xlnet}, are less commonly used.

We briefly introduce the representative PLMs in Table \ref{tab:plm}, LLMs in Table \ref{tab:llm}, and Multimodal Foundation Models (MFMs) in Table \ref{tab:mfm}. 
We refer the readers to \cite{prompt-survey}, \cite{from-bert-to-chatgpt}, and \cite{llm-survey} for more information.

\begin{table}[!htp]\centering
\caption{Overview of MFMs.}\label{tab:mfm}
\rowcolors{1}{lightgray!20}{}
\scalebox{0.85}{
\scriptsize
\begin{tabular}{lrrrr}\toprule
\rowcolor{gray!30}
\textbf{Model} &\textbf{Modality} &\textbf{\#Parameters} &\textbf{Characteristics} \\\midrule
RT-1~\citep{rt1} & Robotic & 35M & \makecell[r]{End-to-End Robotic Transformer, mapping Text and \\Image to Action} \\
VIMA~\citep{jiang2022vima} & Robotic & \makecell[r]{2M/4M/9M/20M/\\43M/92M/200M} & \makecell[r]{leverage text-image prompt to produce motor\\ actions auto-repressively} \\
LAVA~\citep{lynch2022interactive} & Robotic & N/A & Real-time Speech and Natural Language Guidance   to the Robots \\
PALM-E~\citep{driess2023palme} &Robotic &562B & \makecell[r]{Embodied Multi-modal adds Robotic or Object states\\ with Image and Text} \\
Data2Vec~\citep{baevski2022data2vec} &Text/Image/Audio &N/A & \makecell[r]{Unified framework predicts latent representations \\instead of modality-specific targets} \\
CLIP~\citep{clip} &Text/Image &428M &Jointly learn text and image representation interactively \\
VLMo~\citep{bao2022vlmo} &Image/Multimodal &130M & \makecell[r]{Unified various modalities by MOME Transformer, \\trained jointly with ITC, ITM and MLM}\\
Flamingo~\citep{alayrac2022flamingo} &Image/Multimodal &3B/9B/80B & \makecell[r]{Few-shot in-context learning of visual \\and text multi-modal tasks} \\
CoCa~\citep{yu2022coca} &Image/Multimodal & \makecell[r]{Base383M/Large787M/\\2.1B} & \makecell[r]{Unified single-encoder, dual-encoder and encoder-decoder \\and trained with contrastive and captioning loss} \\
PaLI~\citep{chen2022pali} & Image/Multimodal &3B/15B/17B & \makecell[r]{Joint training large scale of mixed mulit-modal\\ and multilingual tasks} \\
FLAVA~\citep{singh2022flava} &Text/Image/Multimodal &350M & \makecell[r]{Unimodal, Cross-Modal, and Multi-Modal \\Foundational Model trained with \\MMM, ITM, MIM and MLM }\\
OFA~\citep{one-for-all} &Text/Image/Multimodal & \makecell[r]{Tiny33M/Medium93M/\\Base182M/Large472M/\\Huge930M} & \makecell[r]{Unified architectures, tasks, and modalities by instruction \\based pre-training and fine-tuning} \\
BEiT-3~\citep{beit3} &Text/Image/Multimodal &1.9B & \makecell[r]{General multimodal foundation model on text, \\image and text-image pair with \\MDM (Masked Data Modeling)} \\
BLIP~\citep{li2022blip} &Text/Image/Multimodal &446M & \makecell[r]{Use a synthetic caption producer and a noise caption\\ filter boostrappingly train a unified multi-modal\\ model with ITC, ITM and LM loss} \\
BLIP-2~\citep{li2023blip2} & Text/Image/Multimodal &474M/1.2B & \makecell[r]{Bridge the gap between a frozen image encoder and \\a frozen LM in two stages by a Querying Transformer} \\
KOSMOS-1~\citep{mllm} &Text/Image/Multimodal &1.6B &Instruct and Multi-modal Transformer \\
GPT-4~\citep{gpt4} &Text/Image/Multimodal &N/A &Multi-modal supported ChatGPT \\
\bottomrule
\end{tabular}
}
\end{table}

\subsection{Prompting\label{prompting}}

According to \cite{DBLP:journals/corr/abs-2210-02406}, 
prompting refers to the interaction methods that focus on calling a model via prompts, without involving any parameter updating\footnote{This definition is a bit different from that of ~\cite{prompt-survey}. We align this definition as ``Tuning-free Prompting'' in ~\cite{prompt-survey}'s categorization. Additionally, we put ``Promptless Fine-tuning'', ``Fixed-prompt LM Tuning'', ``Prompt+LM Tuning'' in §\ref{fine-tuning} and ``Fixed-LM Prompt Tuning'' in §\ref{param-efficient}.}. 
This line of research stems from in-context learning~\citep{brown2020language, icl_survey}, a significant capability of large language models. 
In-Context Learning (ICL) refers to the approach that allows large language models to learn from examples provided in context~\citep{brown2020language}. 
Moreover, the task description can also be incorporated within the context, accompanied with few-shot examples~\citep{sanh2021multitask, flan, naturalinstructions, supernaturalinstructions}. 
Prompting is one of the simplest ways to incorporate interactive messages. However, making it effective can still be tricky, as we will discuss below. 

Note that in this subsection, the discussion focuses on large-scale generative language models, as prompting is challenging to implement with small language models, which may necessitate fine-tuning with prompts~\citep{ite-prompting}.

In the following subsections, 
prompting methods are classified into three categories according to their characteristics and objectives: 
(1) Standard Prompting with straightforward task descriptions and demonstrations (i.e., examples) as context for instruction-following; 
(2) Elicitive Prompting with the context which can stimulate the language model to generate intermediate steps for reasoning; 
and (3) Prompt Chaining, which cascades multiple language model runs for complex reasoning and pipelined tasks.

\subsubsection{Standard Prompting\label{standard-prompting}}

\begin{wrapfigure}[14]{r}{7cm}
\centering
\vspace{-10pt}
 \includegraphics[width=1 \linewidth]{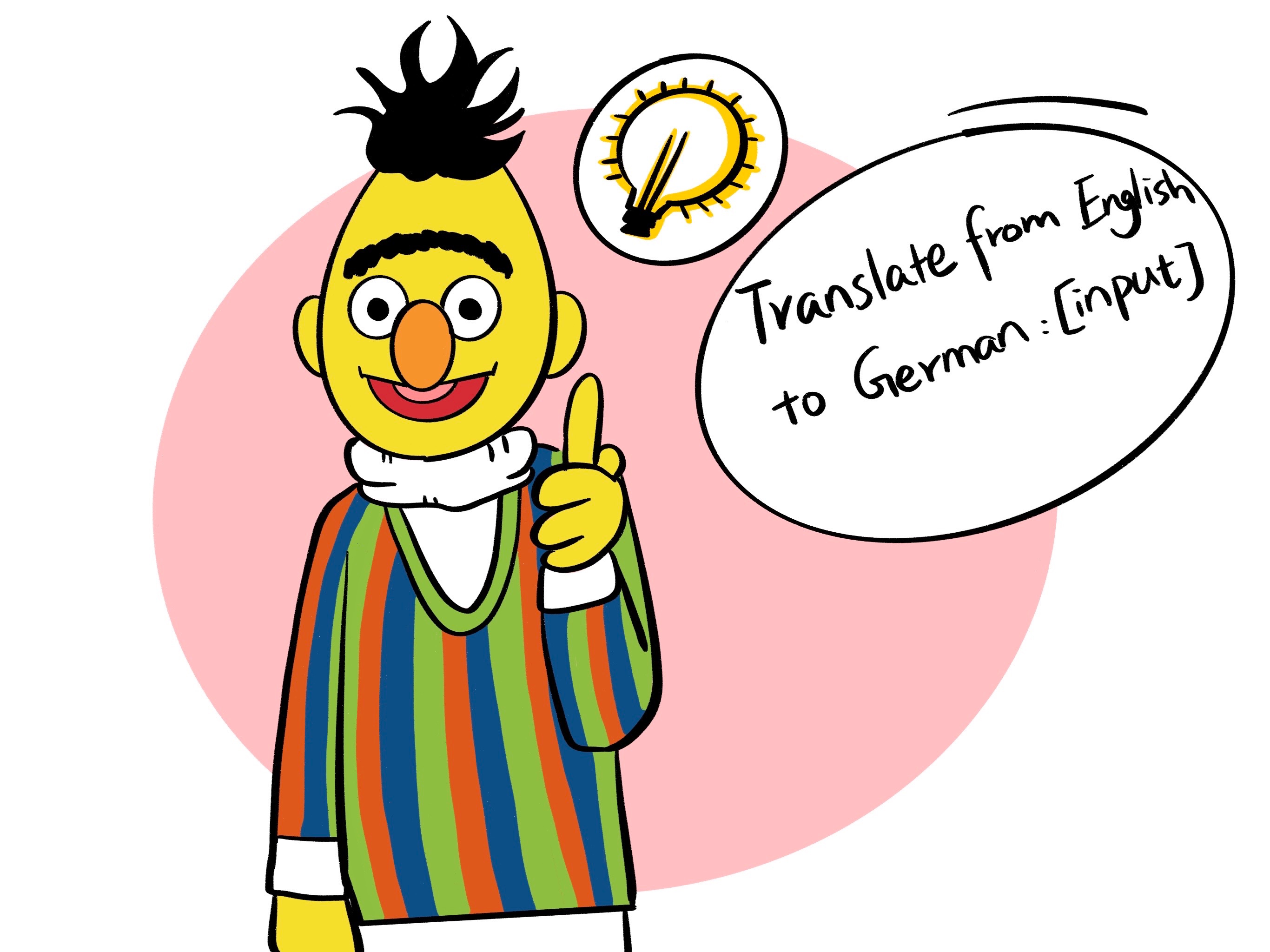}
        \caption{Standard Prompting.}
        \label{fig:standard-prompting}
\end{wrapfigure}

Standard prompting represents the most elementary form of In-Context Learning. 
The prompting context primarily comprises a concise, answer-focused task description, along with few-shot examples, as elucidated in Section §\ref{natural-language}.
In Natural Instructions~\citep{naturalinstructions} and Super-Natural Instructions~\citep{supernaturalinstructions}, the fundamental structure of a context, or instruction, is composed of: task definition, several positive examples accompanied by explanations (demonstrations), and numerous negative examples with clarifications. 
Despite its simplicity, various approaches to standard prompting continue to be proposed, as large language models tend to be context-sensitive, often resulting in a lack of robustness~\citep{liu-etal-2022-makes, gu2022robustness, calibrate-before-use, lu-etal-2022-fantastically, icl_survey, chen2022relation}.

This line of research endeavors to enhance the organization of instructions to improve the performance of ICL~\citep{icl_survey}, which enables a language model to better understand and respond to the interaction messages. 
In accordance with ~\cite{icl_survey}, this primarily entails optimizing the subsequent factors: (1) instance selection; (2) instance processing; and (3) instance combination.

\paragraph{Instance Selection.} 
In order to find useful examples, various unsupervised prompt retrieval methods can be utilized, including distance metrics~\citep{liu-etal-2022-makes}, mutual information~\citep{sorensen-etal-2022-information}, and n-gram overlap~\citep{agrawal2022incontext}, which have been discussed in~\cite{icl_survey}. 
Additionally, \cite{learning-to-retrieve-prompts} and \cite{cheng2023uprise} utilize learned retrievers to identify the most relevant demonstrations to the input. 
\cite{zhang2022active} select demonstrations using reinforcement learning. 
\cite{li2023finding} propose \textit{InfoScore}, a metric designed to evaluate the informativeness of examples, which facilitates example selection using feedback from language models. It employs an iterative diversity-guided search algorithm to improve and assess the examples. 
Most studies along this line build upon the premise that an increased relevance of demonstrations directly correlates with enhanced ICL performance~\citep{liu-etal-2022-makes}.
However, \cite{si2022reliable} find that using randomly sampled demonstrations leads to similar results with GPT-3~\citep{brown2020language} compared to in-distribution demonstrations. 
\cite{working-memory} reveal that controllability and robustness in LLMs can be improved by incorporating counterfactual and irrelevant contexts during fine-tuning. 

\paragraph{Instance Processing.} 
The processing of context involves four main types: expansion, filtering, edit and formatting. 
For example, 
SuperICL~\citep{xu2023small} expands in-context examples by incorporating labels, predicted by a small plug-in model, and their associated confidence scores to augment the context for large language models. 
\cite{zhou2022large} employ LLMs for instruction generation, example generation, and filtering through a scoring model. 
\cite{honovich2022instruction} generate task descriptions based on examples. 
GrIPS~\citep{GrIPS} employs a gradient-free, edit-based approach to conduct instruction search (processing). In particular, it follows an iterative process of modifying the base instruction at the phrase-level and subsequently evaluating the candidate instructions to identify the optimal one. 
ProQA~\citep{zhong-etal-2022-proqa} uses a structured schema to format the context. 

\paragraph{Instance Combination.} 
The order and structure of demonstrations in a given context also play a crucial role~\citep{liu-etal-2022-makes, lu-etal-2022-fantastically, ye2023context, icl_survey}. 
For example, 
~\cite{liu-etal-2022-makes} and ~\cite{lu-etal-2022-fantastically} sort examples in the context according to their distance and entropy metrics with the input, respectively, as mentioned in ~\cite{icl_survey}. 
Batch prompting~\citep{cheng2023batch} enables LLMs to perform inference on multiple samples in a batch, thus reducing token and time costs while maintaining the overall performance. 
Structured prompting~\citep{hao2022structured} involves encoding multiple groups of examples into multiple LM replica, which are then merged using rescaled attention. 
This process allows LMs to incorporate and contextualize 1000+ examples. 
ICIL~\citep{ye2023context} puts multiple task instructions composed of task definitions and groups of examples together in the context to improve LLMs' zero-shot task generalization performance.

Note that although the diverse approaches mentioned in this part are mainly designed for general-purpose in-context learning, they can be used as methods for interaction message communication. 
During the interaction with language models, 
determining the most appropriate way to organize context for interaction messages via elaborate prompt engineering is crucial for performance gain. 
For example, 
in the scope of KB-in-the-loop, 
\cite{lazaridou2022internet}, \cite{izacard2022atlas}, and \cite{ram2023incontext} work on how to feed the retrieved knowledge into language models via ICL; 
in the scope of env-in-the-loop, 
\cite{weir2022oneshot} demonstrate how to generate task instructions and enable cross-environment transfer to help agents generalize their execution. 

\subsubsection{Elicitive Prompting\label{elicitive-prompting}}

Extending standard prompting, elicitive prompting improves the abilities of LLMs, such as reasoning and planning, by providing them with extra step-by-step guidance in context. 

\paragraph{Few-Shot Demonstrations.} 
Typical chain-of-thought~\citep{Wei2022ChainOT} uses few-shot examples with reasoning steps to elicit reasoning as shown below:
\begin{tcolorbox}[colback=white!95!gray,colframe=gray!50!black,rounded corners,label={elicitivebox}]
\textbf{Question}: If a rectangle has a width of 5 units and a length of 8 units, what is its perimeter?

\textbf{Answer}: The perimeter of a rectangle is the sum of the lengths of all its sides. In this case, the rectangle has two sides with a length of 5 units and two sides with a length of 8 units. Therefore, its perimeter is 2 x (5 + 8) = \textcolor{zekundarkgreen}{26 units}.

\textbf{Question}: If I need to be at work by 9:00 am, and it takes me 20 minutes to drive there, what time should I leave my house?

\textbf{Answer}: \textcolor{gray}{[to be generated]}
\end{tcolorbox}

\begin{wrapfigure}[13]{r}{.45\textwidth}
\centering  
\vspace{-7pt}
 \includegraphics[width=.8 \linewidth]{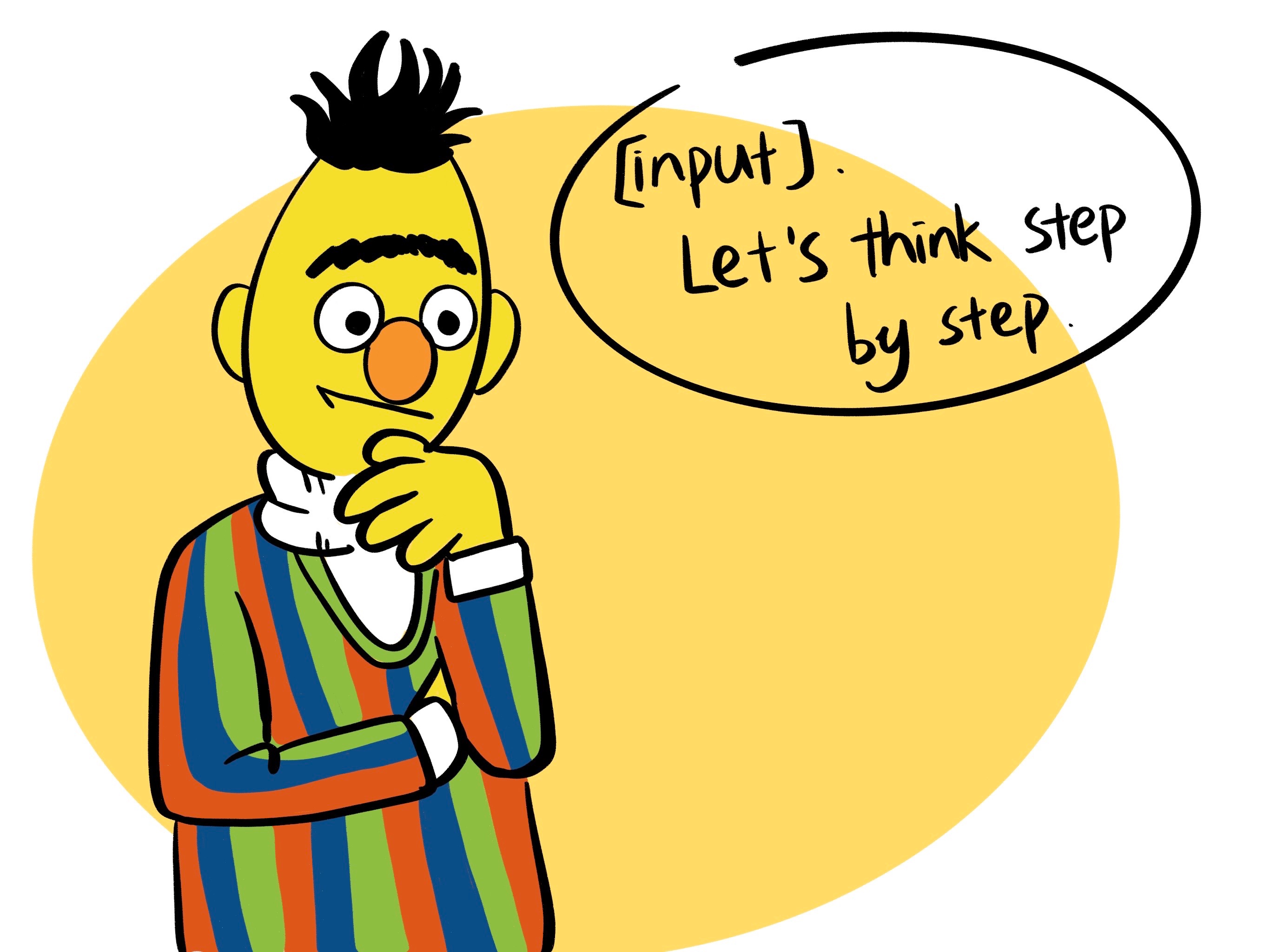}
        \caption{Elicitive Prompting.}
        \label{fig:elicitive-prompt}
\end{wrapfigure}

For example, 
Scratchpads~\citep{nye2021show} and CoT~\citep{Wei2022ChainOT} are two representative techniques for elicitive prompting. 
They explicitly describe the reasoning steps in the few-shot examples, significantly improving math reasoning abilities compared with standard prompting. 
Least-to-most prompting~\citep{zhou2022least} aims to tackle complex tasks that CoT struggles with.
It achieves this by decomposing a complex problem into smaller and more manageable ones with few-shot demonstrations. 
Other follow-up works focus on how to improve the robustness of CoT, such as majority voting on results~\citep{wang2022self}, perplexity check~\citep{fu2022complexity}, or retrieving CoTs from pre-defined clusters~\citep{zhang2022automatic}. 

\paragraph{Other Forms of Instructions.} 
According to recent studies, it may not be necessary to rely solely on human-written, step-by-step rationales for eliciting prompts, as other forms of instructions may be useful. 
For example, zero-shot CoT~\citep{kojima2022large} uses a simple phrase ``\textit{Let's think step by step.}'' to induce the CoT-style reasoning in zero-shot settings:
\begin{tcolorbox}[colback=white!95!gray,colframe=gray!50!black,rounded corners,label={elicitivebox2}]
\textbf{Question}: If I need to be at work by 9:00 am, and it takes me 20 minutes to drive there, what time should I leave my house?

\textbf{Answer}: Let's think step by step: \textcolor{gray}{[to be generated]}
\end{tcolorbox}
The format of answers~\citep{marasovic2021few} and task descriptions~\citep{naturalinstructions} have also been explored to serve as elicitive prompts. 
In addition to text-form CoT, Program-of-Thought (PoT)~\citep{chen2022program}, Program-aided Language Model (PAL)~\citep{gao2022pal}, and ViperGPT~\citep{suris2023vipergpt} leverage program-form CoTs to obtain reliable reasoning performance in many tasks that programs can solve. 
PoT, PAL, and ViperGPT offer advantages over text-based CoT since they deliver verified, stepwise results by executing the programs. 
Vanilla CoT, on the other hand, cannot verify results.
Furthermore, through specially-designed prompts (e.g., ``\textit{Search[query]}'', ``\textit{<API> Calculator(735 / 499) → 1.47 </API>}''), humans can unlock tool-using abilities of language models, such as web-searching~\citep{toolformer, yao2022react}, calculators~\citep{toolformer}, physical simulation~\citep{liu2022mind}, etc (c.f., §\ref{sec-acting}).

Note that in the scope of interactive natural language processing, elicitive prompting can be used to enhance reasoning and planning capabilities of language models during interactions with other objects~\citep{yao2022react, ai-chains, wu2022promptchainer, yang2023mmreact, zhang2022automatic, reasoning-survey}. 
Furthermore, the idea of elicitive prompting is usually instantiated within the scope of model/tool-in-the-loop (§\ref{model-tool-in-loop}), which will be discussed in detail in the next part (§\ref{prompt-chaining}).

\subsubsection{Prompt Chaining\label{prompt-chaining}}

\begin{wrapfigure}[13]{r}{7cm}
\centering
\vspace{-10pt}
 \includegraphics[width=1 \linewidth]{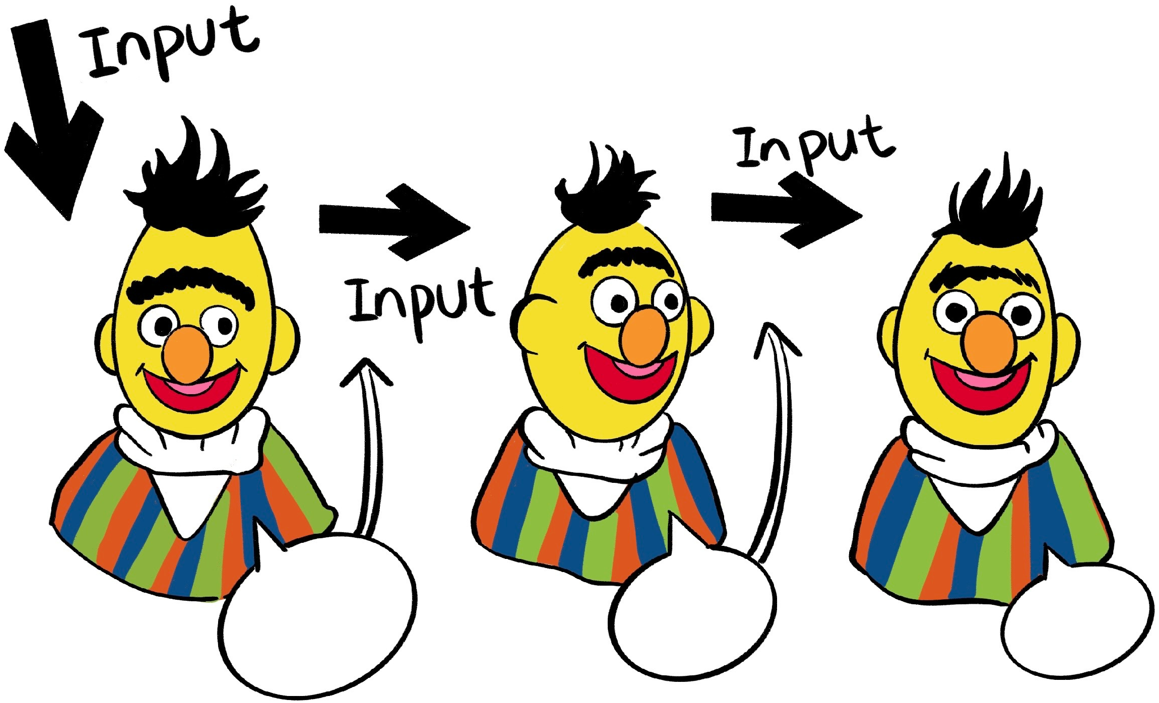}
        \caption{Prompt Chaining.}
        \label{fig:prompt-chain}
\end{wrapfigure}

An increasing number of studies are using multi-stage chain-of-thought to improve multi-hop reasoning capabilities. In this approach, LMs are cascaded and can be prompted via different contexts, allowing for more complex reasoning. 
This is in contrast to typical elicitive prompting, which generally only performs one stage of chain-of-thought via In-Context Learning for reasoning. 
This approach is intuitive as it can aid in generating precise reasoning steps by conducting multiple model runs with different yet interdependent prompts~\citep{reasoning-survey}. 
In contrast, elicitive prompt relies on a single model run with only one context~\citep{reasoning-survey}. 

By decomposing the task and cascading language models for different reasoning steps or sub-tasks, prompt chaining can not only perform multi-hop reasoning~\citep{lm-cascades, reasoning-survey}, but also work well in pipelined tasks such as peer review writing~\citep{ai-chains} and advertisement generation~\citep{wu2022promptchainer}. 
Prompt chaining is one of the fundamental methods for model/tool-in-the-loop natural language processing (c.f., §\ref{model-tool-in-loop}).

\begin{figure}[htbp]
  \centering
  \begin{subfigure}{0.25\textwidth}
    \includegraphics[width=\textwidth]{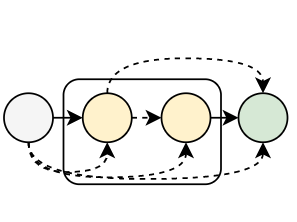}
    \caption{Sequential.}
    \label{fig:sequential}
  \end{subfigure}
  \begin{subfigure}{0.25\textwidth}
    \includegraphics[width=\textwidth]{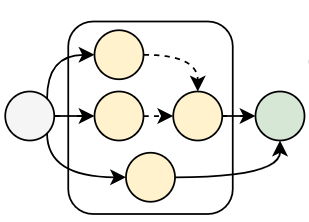}
    \caption{Branched.}
    \label{fig:branched}
  \end{subfigure}
  \begin{subfigure}{0.25\textwidth}
    \includegraphics[width=\textwidth]{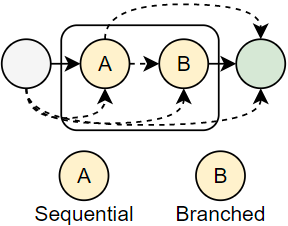}
    \caption{Decomposed.}
    \label{fig:decomposed}
  \end{subfigure}
  \begin{subfigure}{0.18\textwidth}
    \includegraphics[width=\textwidth]{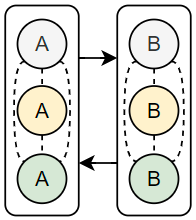}
    \caption{Interactive.}
    \label{fig:interactive}
  \end{subfigure}
  \caption{Examples of prompt chaining schemes. Gray, yellow, and green circles refer to input queries, intermediate reasoning steps, and final responses, respectively. Solid and dashed arrows represent indispensable and optional (i.e., none, single, or multiple) conditional probabilities ($P(Y|X)$), respectively. The rounded rectangle refers to a looping block.}
  \label{fig:chain}
\end{figure}

LM Cascades~\citep{lm-cascades} has presented some works in this line, including sequential reasoning mechanisms~\citep{nye2021show, Wei2022ChainOT, creswell2022selectioninference}, reasoning procedures with verifiers or tools~\citep{math-verifier, nakano2021webgpt, liu2022mind}, and multi-agent interacting question-answering~\citep{srivastava2022beyond}\footnote{\url{https://github.com/google/BIG-bench/tree/main/bigbench/benchmark_tasks/twenty_questions}}. 
\cite{reasoning-survey} investigated the enhancement of reasoning through language model prompting, focusing on strategy enhancement and knowledge enhancement. The study explored various aspects, such as prompt engineering, process optimization, external engines, and both implicit and explicit knowledge. 
However, to the best of our knowledge, no survey has systematically examined the structure of prompt chaining. 
Hence, in this part, we divide the prompt chaining schemes into four categories according to their topology, as shown in Figure \ref{fig:chain}. 
Furthermore, instead of fixed prompt chaining schemes, users can customize them (\S\ref{customization} Customization). And they can also be constructed automatically (\S\ref{automatization} Automatization).

\paragraph{Sequential.} 
The nodes are arranged in a straight line, where each node takes as input the specific outputs of the previous nodes, including the initial input query. 
For example, 
Self-Ask~\citep{self-ask} and Successive Prompting~\citep{DBLP:conf/emnlp/DuaG0G22} construct the reasoning chain via sequential question generation (QG) and question answering (QA) nodes. 
~\cite{ite-prompting} further enable smaller language models to construct the similar QG-QA chain via a learned context-aware prompter. 
Selection-Inference~\citep{creswell2022selectioninference} begins by utilizing the selection module to choose a group of relevant facts based on the given question. Subsequently, the inference module generates new facts by utilizing this subset of facts.
Multimodal-CoT~\citep{zhang2023multimodal} addresses the visual reasoning problem through a two-step processing consisting of rationale generation and answer inference conditioned on the image, question, and the generated rationale. 
Mind’s-Eye~\citep{liu2022mind} first generates the rendering codes for an intermediate environment simulator to get grounded rationales, and then generates the final answer based on the simulation results. 
Note that when the number of looping blocks becomes zero, it is reduced to standard prompting (§\ref{standard-prompting}). 
When the intermediate reasoning steps and the final answer are simultaneously prompted in a single model run, it is reduced to single-stage CoT (§\ref{elicitive-prompting}).

\paragraph{Branched.} 
The nodes are arranged in a tree-like structure, where a node's output may serve as input to multiple other nodes, and a node's input may come from multiple other nodes' outputs. 
For example, \cite{perez-etal-2020-unsupervised} decompose one multi-hop question into many single-hop sub-questions and then aggregates their answers to get the final answer. 
\cite{wang2022self} first generate a range of reasoning paths through sampling from the language model’s decoder and then aggregates the most consistent answer in the final answer set by computing the likelihood of the reasoning paths. 
Ask Me Anything~\citep{arora2022ask} first reformats the question into diverse possible ones with different in-context demonstrations, which are then answered respectively. Finally, the answers are aggregated into the final answer via a learned probabilistic graphical model. 
Tree of Thoughts~\citep{yao2023tree} constructs the language model's reasoning chain in a tree form, enabling evaluation of states via heuristic approaches, and exploration of potential solutions via Breadth-first search (BFS) or Depth-first search (DFS), significantly improving the model's problem-solving capabilities. 
\cite{prompt-survey} have investigated multi-prompt learning, including prompt ensembling, prompt composition, and prompt decomposition, which can all be viewed as in this line.

\paragraph{Decomposed.} 
The overall process is linear, but some nodes can be broken down further into nested or hierarchical chains recursively that follow any of the four schemes described. 
For example, \cite{zhou2022least} first call an LM to reduce the problem into multiple sub-problems and then iteratively call the language model to solve these sub-problems step-by-step (i.e., the first node serves for problem reduction while the second node is another sequential chain). 
Decomposed Prompting~\citep{DBLP:journals/corr/abs-2210-02406} decomposes the question into a sequence of sub-questions, each being answered immediately after generation by utilizing another prompt chaining block or just standard prompting. 
This approach facilitates the hierarchical and recursive decomposition of tasks.

\paragraph{Interactive.} 
The nodes are split into multiple groups, each with their own functions. 
These groups communicate with each other in an alternating fashion to construct a chain of interactions. 
For example, Socratic Models~\citep{zeng2022socratic} and Inner Monologue~\citep{huang2022inner} partition the groups of nodes according to the modality. They utilize LLMs for planning and reasoning while making use of VLM for observation. 
\textit{ChatGPT Asks, BLIP-2 Answers}~\citep{chatgpt-blip-2} use ChatGPT to ask questions about an image, while BLIP-2~\citep{li2023blip2} is used to answer these questions. 
This communication through question-answering is performed interactively to generate the image captions. 
The visual reasoning path in ~\citep{see-think-confirm} can be divided into two groups, one for answer and explanation generation, and another for explanation verification via a multimodal classifier.
The frameworks of ~\cite{math-verifier,self-verification} can be viewed as interaction between the thought generation group and the thought verification group. 
MindCraft~\citep{bara-etal-2021-mindcraft} first assigns two agents with different skills (i.e., recipe knowledge) and then lets them interact through question-answering to accomplish the task in the \textit{MineCraft} environment.

\paragraph{Customization\label{customization}.} 
The prompt chaining scheme depends on several factors related to the nature of the problem, such as its complexity, structure, and the availability of relevant data. 
Hence, due to their diversity and variability, a fixed prompt chaining scheme may not satisfy the users’ needs. 
An intuitive solution is to enable users to create or modify the prompt chaining scheme on their own accord, which can further enhance the debuggability and configurability of the system~\citep{ai-chains, wu2022promptchainer}. 
Generally, this feature is more important for pipelined tasks such as peer review writing, brainstorming, personalized flashcard creation, writing assistant, etc~\citep{ai-chains, wu2022promptchainer}. 
For example, 
AI Chains~\citep{ai-chains} defines a set of primitive operations such as classification, factual query, information extraction, split points, compose points, etc. 
These primitive operations are implemented with different instructions fed into the language model. 
An interactive interface then shows the prompt chaining schemes, which users can customize to construct a pipeline. 
PromptChainer~\citep{wu2022promptchainer} also introduces an interactive interface that facilitates the visual programming of chains. 
It divides the nodes into three types: LLM nodes, such as generic LLM and LLM classifier; Helper nodes, such as model output evaluation, data processing, and generic JavaScript; and Communication nodes, such as data inputs, user actions, and API calls. 
The workflow defined by the graph of nodes is also transparent and configurable. It has been shown that this approach can assist users in building a satisfactory pipeline for applications such as music chatbots, advertisement generators, image query generators, and writing assistants~\citep{wu2022promptchainer}. 

\paragraph{Automatization\label{automatization}.} 
The prompt chaining schemes can also be constructed automatically. 
For example, 
ReAct~\citep{yao2022react} and MM-ReAct~\citep{yang2023mmreact} determine whether a thought should be generated or a tool should be called automatically based on the context. 
ToolFormer~\citep{toolformer} can determine which tools to utilize or whether to use tools based on the context, as it is trained on tool-use prompted data. 

\subsection{Fine-Tuning\label{fine-tuning}}

Fine-tuning refers to the process of updating the parameters of a model. 
The ongoing interaction provides an increasing amount of interaction messages that can be used to update the models' parameters, resulting in better language model adaptation to interactions like instruction following~\citep{flan, sanh2021multitask, aribandi2021ext5, xu2022zeroprompt,ouyang2022training, yaofu-notion-blog} 
and grounding~\citep{xie2022unifiedskg, sharma2021skill, emboded-bert, episodic-transformer}. 
This line of research also explores how to make effective use of this new data 
without catastrophic forgetting~\citep{gururangan-etal-2020-dont, he-etal-2021-effectiveness, time-aware, jang2021towards, jin2021lifelong, qin-etal-2022-elle}, 
how to ensure generalization to new tasks~\citep{flan, sanh2021multitask, aribandi2021ext5, xu2022zeroprompt, ouyang2022training, yaofu-notion-blog}, and 
how to adapt the language model more efficiently~\citep{prompt-survey, delta-tuning,li2021prefixtuning, lester-etal-2021-power, pmlr-v97-houlsby19a, hu2022lora}.

In this subsection, we discuss four commonly employed fine-tuning-based methods: 
(1) Supervised Instruction Tuning, which aims to adapt language models for instruction following and to enhance their task generalization abilities (§\ref{instruction-tuning}); 
(2) Continual Learning, which aims to infuse new data into language models without catastrophic forgetting (§\ref{contiunal}); 
(3) Parameter-Efficient Fine-Tuning, which focuses on the efficient adaptation of language models (§\ref{param-efficient}); and 
(4) Semi-Supervised Fine-Tuning, which further tackles the problem of unlabeled data, as in some cases, the interaction message may not provide adequate supervision to train the model~\citep{li2023blip2, alpaca, zelikman2022star, llmteacher, huang2022large} (§\ref{semi-supervised}).

\subsubsection{Supervised Instruction Tuning\label{instruction-tuning}}

Supervised instruction tuning involves fine-tuning a pre-trained language model using data that provides task instruction supervision. 
 Various studies ~\citep{2020t5,aribandi2021ext5, xu2022zeroprompt, sanh2021multitask,xie2022unifiedskg,li2022consisttl,weller2022use,ouyang2022training,flan,chung2022scaling,flan-collection,iyer2022opt, glaese2022improving,zhang2023chinese} have been conducted in this area. 
 These methods fine-tune a pre-trained model by using supervised instructions on a multitask mixture, covering various tasks and inducing zero-shot generalization capabilities. 

\begin{wrapfigure}[15]{r}{7cm}
\centering
 \includegraphics[width=1 \linewidth]{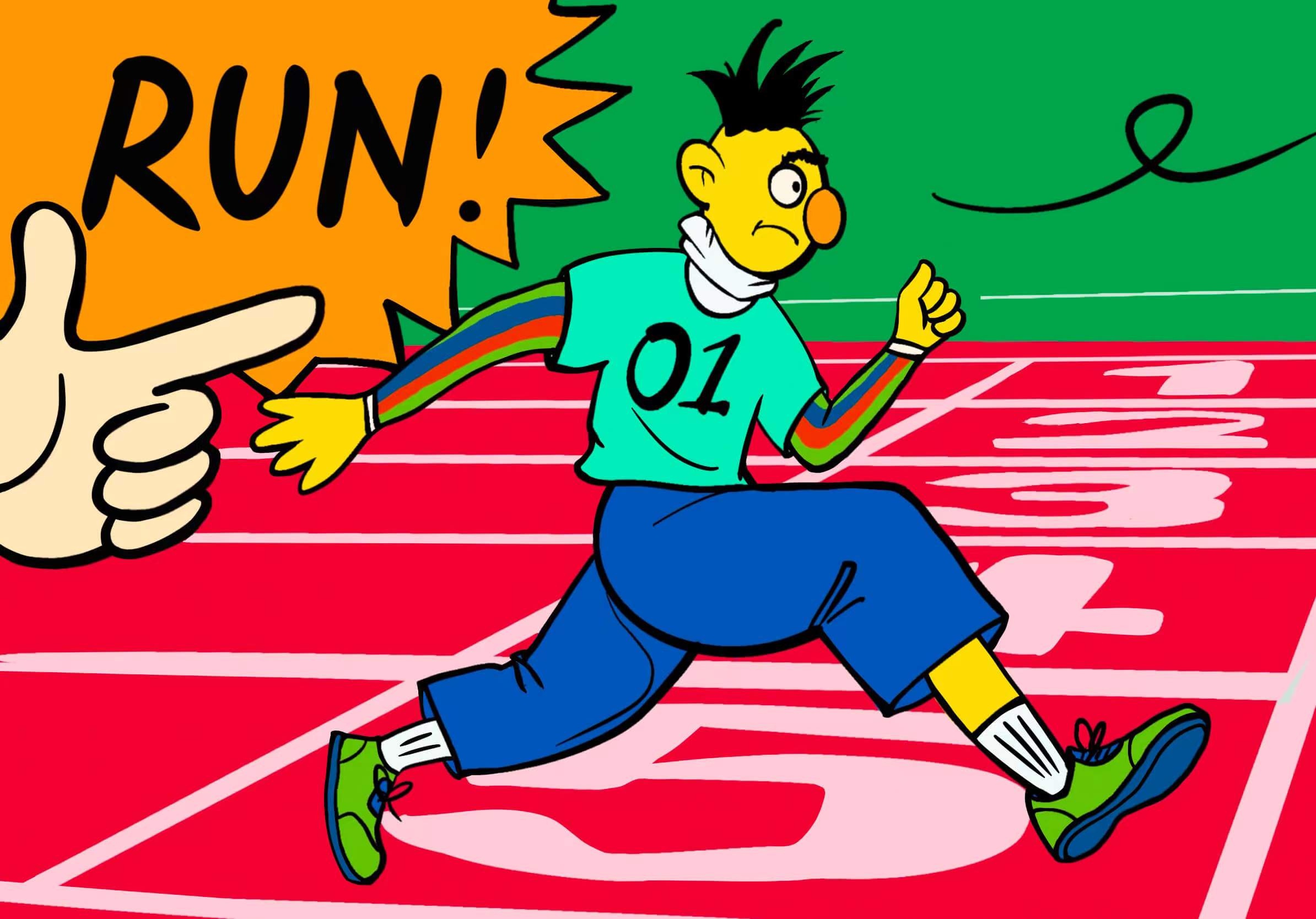}
        \caption{Supervised Instruction Tuning.}
        \label{fig:instruction}
\end{wrapfigure}

The first line of work, investigated by researchers such as ~\cite{2020t5, aribandi2021ext5, xu2022zeroprompt, sanh2021multitask, wang2022super, xie2022unifiedskg, muennighoff2022crosslingual, li2022consisttl, weller2022use, flan}, focuses on providing instructions to language models as part of the input. 
Typically, these instructions are prepended to the input and contain specific details about the task the model is expected to perform. 
These models explore different aspects, such as training and evaluation data, model architectures (decoder-only v.s. encoder-decoder), instruction formatting, task mixtures, and other related factors. 
The discussed studies offer conclusive evidence that fine-tuning language models on multiple NLP tasks and incorporating instructions allow these models to generalize to unseen tasks and better understand and respond to user queries~\citep{yaofu-notion-blog}. 
As demonstrated in ~\cite{kaplan2020scaling} and \cite{wei2022emergent}, scaling up language models leads to improvements in performance. The researchers also study the impact of different scales of instruction data, aiming to better understand the influence of the amount and diversity of this kind of training data~\citep{zhang2023chinese}.

Subsequently, OpenAI releases the InstructGPT~\citep{ouyang2022training} and develops a series of GPT-3.5 variants, all of which are built upon the foundation of GPT-3~\citep{brown2020language}. 
These variants include \textit{code-davinci-002} and \textit{text-davinci-002}, which only involve supervised instruction tuning, as well as \textit{text-davinci-003} and \textit{gpt-3.5-turbo}, which are refined through both supervised fine-tuning and reinforcement learning from human feedback (RLHF). 
These modifications enhance the models' alignment with human intent, resulting in more truthful and less toxic responses from the language models.
Along with this line, DeepMind's Sparrow~\citep{glaese2022improving} and Anthropic's Claude\footnote{\url{https://www.anthropic.com/index/introducing-claude}} also use instruction tuning and RLHF to teach models to produce answers that align with human values~\citep{liu2023perspectives}. 
Furthermore, apart from scaling up the instructional fine-tuning process by increasing the number of tasks and the model size, \cite{chung2022scaling,flan-collection} improve the process by jointly integrating chain-of-thought data during instruction tuning. 
They fine-tune the T5~\citep{2020t5} and PaLM~\citep{chowdhery2022palm} into FLAN-T5 and FLAN-PaLM models~\citep{flan-collection}, resulting in robust performance across a diverse range of natural language processing tasks, including translation, reasoning, and question answering.

Apart from supervised instruction fine-tuning using existing instruction datasets or human-annotated instruction datasets, recent studies have also highlighted semi-supervised approaches for creating instruction-following data generated by LLMs~\citep{wang2022selfinstruct, alpaca, xu2023baize, peng2023instruction, zhou2023controlled, honovich2022unnatural}. 
This synthetic data can be utilized for fine-tuning PLMs~\citep{alpaca, xu2023baize}. 
We refer the readers to §\ref{semi-supervised} for more information.

Supervised instruction tuning can be regarded as one of the crucial steps in interactive natural language processing. By fine-tuning language models with supervised instructions, their ability to comprehend and respond to a diverse range of queries can be enhanced, enabling them to perform tasks such as question-answering and task completion with greater precision and efficiency.

\subsubsection{Continual Learning\label{contiunal}}

LMs that have been pre-trained on static data may become outdated and no longer aligned with new domains or tasks~\citep{toolformer, qin-etal-2022-elle}. 
Therefore, it is beneficial to utilize interaction messages accumulated over time to fine-tune LMs. 
This guarantees that the LMs are up-to-date with the newest information and perform optimally in novel scenarios~\citep{dalvi2022towards}. 
Although typical fine-tuning is an effective approach to updating an LM, it can suffer from catastrophic forgetting~\citep{catastrophic-forgetting}. 
As data size increases, attempting to incorporate new knowledge into a fixed-sized LM may result in losing previous knowledge. 
Continual Learning (CL) is a promising solution to this problem. 
It seeks to continuously integrate knowledge from novel sources without expunging prior learning~\citep{continual-comparative}.

\begin{wrapfigure}[15]{r}{7cm}
\centering
 \includegraphics[width=1 \linewidth]{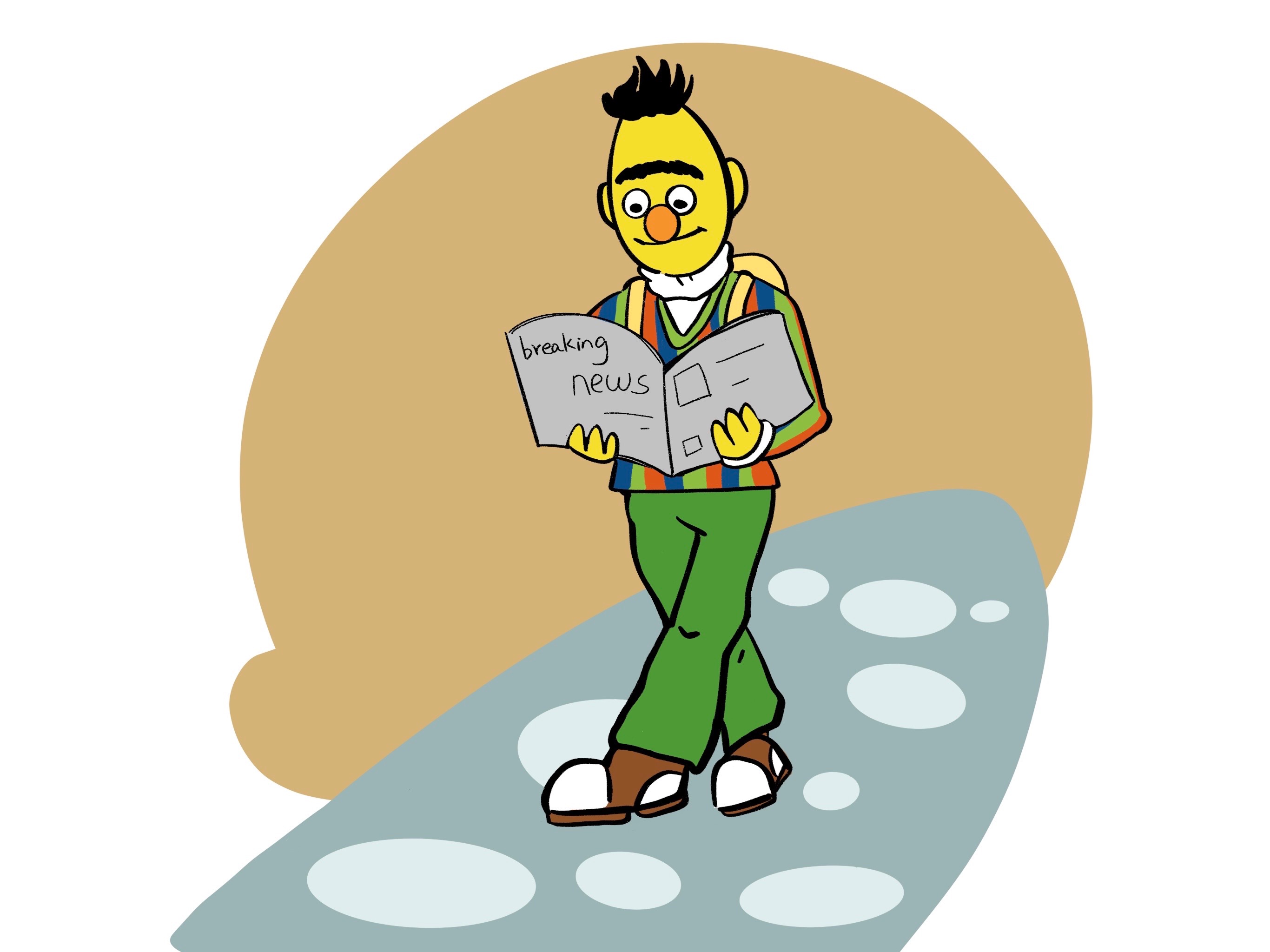}
        \caption{Continual Learning.}
        \label{fig:continual}
\end{wrapfigure}

First of all, we present a mathematical formalization of continual learning. 
Consider a language model that is presented with a sequence of $n$ tasks $(T_1,...,T_n)$. 
For each task $T_k$, the model is provided with a set of $N$ i.i.d. training examples $(x_i , y_i)^N_{i=1}$. 
Assuming that the language model is parameterized by $\theta$ and is aware of the task identity during both training and inference, 
the overall learning objective across all tasks is: 
\begin{align}
    \max _{\theta} \sum_{k=1}^n \sum_{(x, y) \in T_k} \log p(y \mid x; \theta)
\end{align}
In the continual learning setting, 
it involves sequentially optimizing the loss for each task $T_k$ via fine-tuning:  
\begin{align}
    \max_\theta \sum_{x, y \in T_k} \log p(y \mid x; \theta) \text{  for $k$ in \{1,...,n\}}
\end{align}
Such a na\"ive continual fine-tuning may cause catastrophic forgetting, ultimately leading to a decline in overall performance on earlier tasks after the learning of new tasks~\citep{mccloskey1989catastrophic}.

In this part, we briefly introduce recent CL methods that aim to alleviate the forgetting phenomenon, which can be categorized into three groups: 
(1) Regularization, (2) Rehearsal, and (3) Modularization, 
following ~\cite{continual-survey, biesialska-etal-2020-continual}. 

\paragraph{Regularization.} 
Regularization has been widely adopted to inhibit catastrophic forgetting by penalizing the model when it deviates significantly from its previous state. 
This is typically accomplished by determining the essential parameters for the prior task and domain data, and subsequently integrating a regularization term into the loss function, which encourages the model to preserve this knowledge while learning new tasks~\citep{continual-survey}. 
For example, \cite{kirkpatrick2017overcoming} propose Elastic Weight Consolidation (EWC), which involves measuring important parameters using a Fisher information matrix, derived from the magnitude of the gradient update step corresponding to each parameter. 
Following this work, \cite{chen2020recall} introduce a pre-training simulation mechanism that enables the memorization of knowledge for pre-training tasks, eliminating the requirement for pre-training data access. 
\cite{li2022lpc} introduce a novel approach for continual learning in pre-trained models, which calibrates both the parameters and logits. This approach helps in retaining the acquired knowledge while also facilitating the learning of new concepts. 
As a concurrent work, \cite{li2022overcoming} propose to selectively memorize important parameters from previous tasks through a recall optimization mechanism facilitated by regularization, which is evaluated on interactive dialogue generation datasets.

\paragraph{Rehearsal.} 
A rehearsal-based approach to continual learning involves replaying previous task data or synthetic previous task data while learning new tasks. 
For example, 
~\cite{he2021analyzing} conduct preliminary experiments on interactive dialog models, assuming access to the pre-training corpus during fine-tuning. 
Throughout the training process, they mix random subsets of the pre-training corpus based on a mix-ratio that anneals towards the target task. 
ELLE~\citep{qin-etal-2022-elle} employs a memory replay mechanism, where several data from previous tasks is mixed with current task data for tuning. 
Moreover, the synthetic previous task data generated by the old model can be utilized to train the new model as memory replay~\citep{cappellazzo2022exploring}.

\paragraph{Modularization.} 
Modularization refers to separating model parameters into distinct modules or sub-networks, where each module is responsible for performing a specific task or set of tasks~\citep{continual-survey, biesialska-etal-2020-continual, pfeiffer2023modular}. 
The modularization can be achieved by adapter-based methods, which introduce new parameters corresponding to new tasks; Alternatively, it can be based on partial tuning, which freezes or prunes previous task parameters as discussed in \cite{pfeiffer2023modular,biesialska-etal-2020-continual} and \S\ref{param-efficient}. 
For example, \cite{lee2022plug} propose a plug-and-play approach for incorporating the target knowledge into new parameters by applying multiple large-scale updates on plug-in modules, thereby avoiding the risk of forgetting the previously acquired source knowledge in old parameters. 
In SupSup~\citep{wortsman2020supermasks}, a network is first initialized as a base network, and task-specific sub-networks are separately learned for different tasks and acquired with a network-level mask. During testing, the task identity can be provided or inferred using gradient-based optimization, allowing the appropriate sub-network to be retrieved.

Continual learning is crucial for interactive NLP, enabling the model to adapt to the dynamic and diverse nature of user inputs, tasks, and environments. 
Despite considerable efforts to address the issue of catastrophic forgetting, the problem persists, especially in large language models, as they impose substantial computational requirements. 
Additionally, various other questions, such as evaluation for CL and sample efficiency, remain unanswered. 
Researchers persistently strive to improve the efficiency and effectiveness of continual learning for NLP, and advancements in this field will significantly impact the future of interactive NLP systems.

\subsubsection{Parameter-Efficient Fine-Tuning\label{param-efficient}}

\begin{wrapfigure}[13]{r}{7cm}
\centering
\vspace{-40pt}
 \includegraphics[width=1 \linewidth]{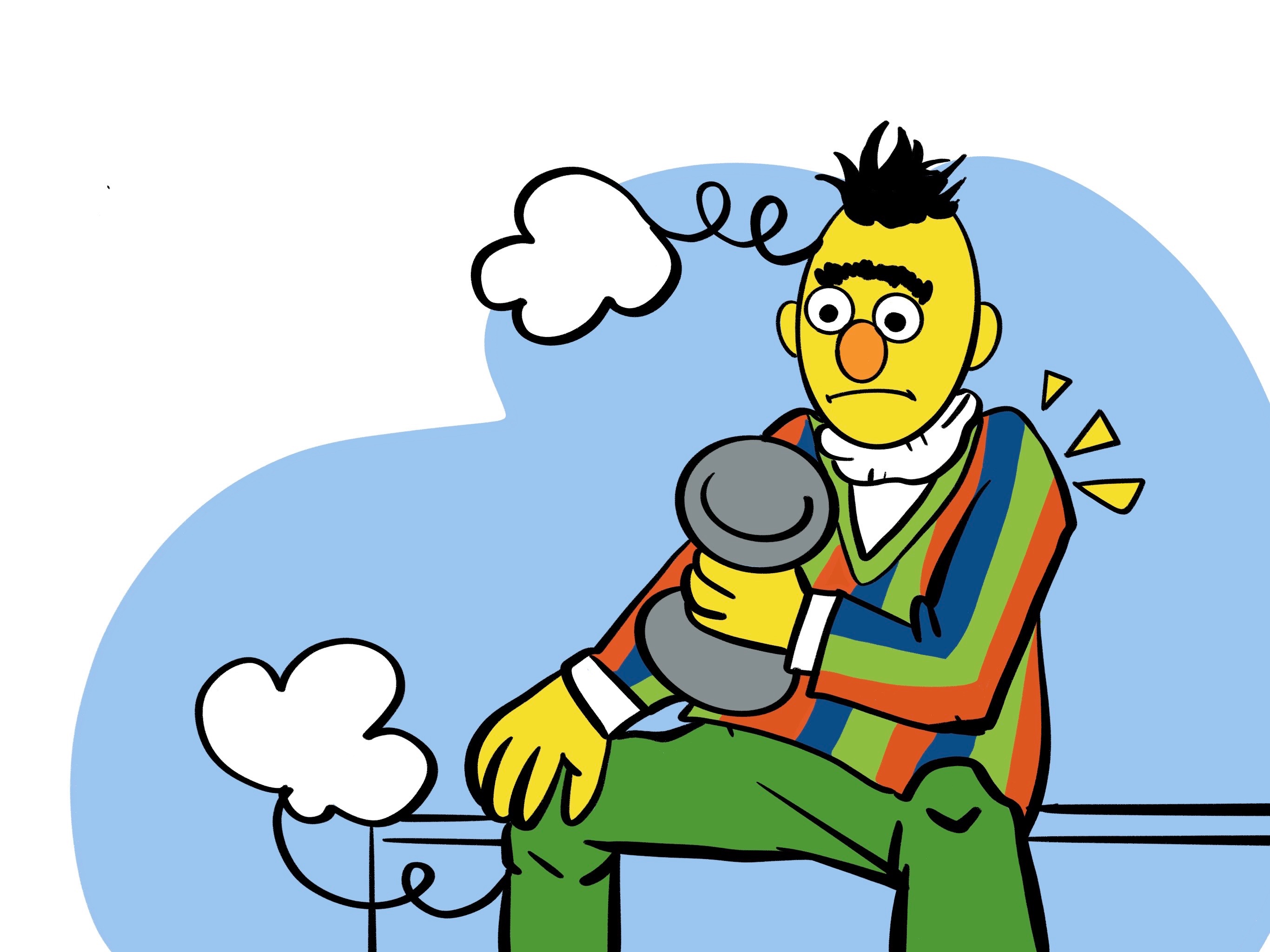}
        \caption{Parameter-Efficient Fine-Tuning.}
        \label{fig:param-efficient}
\end{wrapfigure}

When the size of a pre-trained language model grows larger and larger, it becomes more and more difficult to fine-tune the model, especially when the size of interaction message data is limited. This is because very large models often require impractical amount of GPU memory to fine-tune, and 
over-fitting easily happens when the training data is limited. 
Thus, to avoid these issues, various parameter-efficient tuning methods are proposed. Parameter-efficient fine-tuning, as its name suggests, only updates a small number of parameters compared to the number of parameters of the full model. By reducing the number of trainable parameters during fine-tuning, parameter-efficient fine-tuning methods consume much less GPU memory and substantially diminish the risk of overfitting. Therefore, parameter-efficient fine-tuning methods have become the de facto practice for fine-tuning LLMs. 

Parameter-efficient fine-tuning can be formalized as:

\begin{align}
    \mathcal{L}_{\theta_0 \subsetneq h(\theta)}(LM(x; h(\theta)))
\end{align}
where $\theta_0$ is the set of tunable parameters, $h(\theta)$ is the change of model parameters by introducing additional modules, pruning, and reparameterization~\citep{delta-tuning}.

Following Delta-Tuning~\citep{delta-tuning}, parameter-efficient fine-tuning methods for iNLP can be divided into two main categories based on whether they introduce additional modules or parameters. 
The first category includes \textbf{partial fine-tuning} techniques, which select a small number of parameters of the pre-trained model during fine-tuning. 
In contrast, the second category of parameter-efficient fine-tuning methods keeps the model frozen and adds extra parameters that are updated during fine-tuning. 
These newly added and tuned modules are called adapters, belonging to \textbf{adapter-based methods}.

\paragraph{Partial Fine-tuning.} 
This fine-tuning method, also known as ``specification-based tuning'', updates only a strict subset of the model parameters while keeping the remaining parameters unchanged during fine-tuning~\citep{delta-tuning}. 
This approach does not add any new parameters to the model, meaning that the number of parameters remains the same (i.e., $|h(\theta)|=|\theta|$). 
Instead, the method specifies explicitly or implicitly which parts of the model's parameters should be optimized by indicating $\theta_0$ and $|\theta_0| << |\theta|$~\citep{delta-tuning}. 
For example, 
we can choose a few layers to freeze and only update the remaining layers. 
~\cite{howard-ruder-2018-universal} initially freeze all layers except for the last layer, which contains less general knowledge, and then gradually unfreeze the remaining layers during the fine-tuning process. 
Moreover, 
BitFit~\citep{DBLP:conf/acl/ZakenGR22} optimizes only the bias terms of a pre-trained model, specifically the ``query'' and ``middle-of-MLP'' bias terms, resulting in a significant reduction in the number of tunable parameters while still maintaining high performance on several benchmarks. 
Diff pruning~\citep{DBLP:conf/acl/GuoRK20} involves learning a delta vector to be applied to the initial pre-trained model parameters. It uses a differentiable approximation of the \textit{L0}-norm penalty which facilitates the delta vector becoming more sparse, thereby resulting in a more parameter-efficient approach to fine-tuning. 
Similarly, 
\cite{head-pruning} encourage the model to prune less important attention heads through regularization, effectively reducing the number of attention heads that need to be fine-tuned. 
SupSup~\citep{wortsman2020supermasks} selectively updates the critical weights of a PLM for specific tasks by learning a sub-network called supermask. 
It learns a superposition of supermasks with gradient-based optimization for an unseen task during inference.

\paragraph{Adapter-based Methods.} 
Adapter tuning~\citep{NIPS2017_e7b24b11} inserts small modules called \textit{adapters} to a model: $\theta \subsetneq h(\theta)$ and $|h(\theta)|-|\theta| << |\theta|$~\citep{delta-tuning}. 
With adapters, we can completely freeze the model and only optimize the newly introduced adapters, i.e., $\theta_0 = h(\theta) - \theta$. 
Conventional adapters follow \cite{pmlr-v97-houlsby19a}'s practice of placing a two-layer feed-forward neural network with a bottleneck after each sub-layer within a Transformer layer, including both the multi-head attention sub-layer and the feed-forward network sub-layer. 
In addition, recent works have introduced many other parameterizations of adapters. 
The most representative ones include Prefix Tuning~\citep{li2021prefixtuning}, Prompt Tuning~\citep{lester-etal-2021-power}, LoRA~\citep{hu2022lora}, and Compacter~\citep{karimi2021compacter}. 
As pointed out by~\cite{he2022towards}, these approaches can be unified into a single framework, wherein a module is added as a residual to specific components of the original computational graph in the Transformer architecture. 
As a result, they can all essentially be considered as different variants of adapters. 
For example, 
Prefix Tuning~\citep{li2021prefixtuning} introduces learnable tokens to the input, which are prepended to the keys and values of self-attention, with the model weights being frozen. 
LoRA~\citep{hu2022lora} introduces a module of low-rank trainable parameters, while the pre-trained model's weights keep frozen. 
In addition to low-rank approximation, Compacter~\citep{karimi2021compacter} also utilizes parameterized hypercomplex multiplications layers~\citep{zhang2021beyond}.  
Compared to partial fine-tuning, adapter-based methods incorporate task-specific additional modules, offering increased flexibility, modularity, compositionality, and shareability~\citep{pfeiffer2020AdapterHub, pfeiffer2023modular}. 
These features also allow adapters to function as a form of task representation. 
For example, 
recent works~\citep{zhou-etal-2022-efficiently, vu-etal-2022-spot} show that the additional parameters introduced in adapter-based approaches reveal inter-task similarities and transferability between tasks.
Specifically, ~\cite{zhou-etal-2022-efficiently} show that intermediate task transferability can be effectively predicted by calculating the cosine similarity between adapter parameters for two tasks under the same backbone model. 

\paragraph{Applications in iNLP.} 
Parameter-efficient fine-tuning has been successfully applied in interactive NLP scenarios since it substantially reduces the parameters required for training and storage. 
For example, 
in the model-in-the-loop setting, 
BLIP-2~\citep{li2023blip2} uses prompt tuning-like method to implement vision model-language model interaction, where the interaction interface is learnable soft tokens which is mapped from the output of a vision model. 
In the KB-in-the-loop setting, 
K-Adapter~\citep{kadapter} uses trainable adapters to infuse knowledge retrieved from knowledge bases into pre-trained language models while keeping the model parameter frozen. 
ROME~\citep{meng2022locating} and MEMIT~\citep{meng2022memit} first locate the factual knowledge on some critical MLP layers of the language model via causal tracing. And then modify their weights to write memories into the model. 
\cite{liang2022transformer} also investigate adapter tuning techniques to train a Transformer-based policy model through imitation learning for robotic manipulation (environment-in-the-loop). 
Additionally, parameter-efficient fine-tuning can be applied to improve the modularity of models, leading to better out-of-distribution generalization, knowledge composition, prevention of catastrophic interference, and continual learning~\citep{pfeiffer2020AdapterHub, qin-etal-2022-elle, pfeiffer2023modular}, which are beneficial for iNLP.

\subsubsection{Semi-Supervised Fine-Tuning\label{semi-supervised}}

According to~\cite{Liang2005SemiSupervisedLF,semi-supervised-book-zien,brief-super-weak}, semi-supervised learning aims to use both labeled data and unlabeled data to train a model. 
Semi-supervised learning can be used for interactive natural language processing in that interaction messages are without sufficient supervision in some cases such as lack of instructions~\citep{wang2022selfinstruct, alpaca}, misaligned image-text signals~\citep{li2022blip}, etc. 
We can formulate semi-supervised fine-tuning as follows: 

\begin{wrapfigure}{r}{.5\textwidth}
\centering
\vspace{-15pt}
 \includegraphics[width=.8 \linewidth]{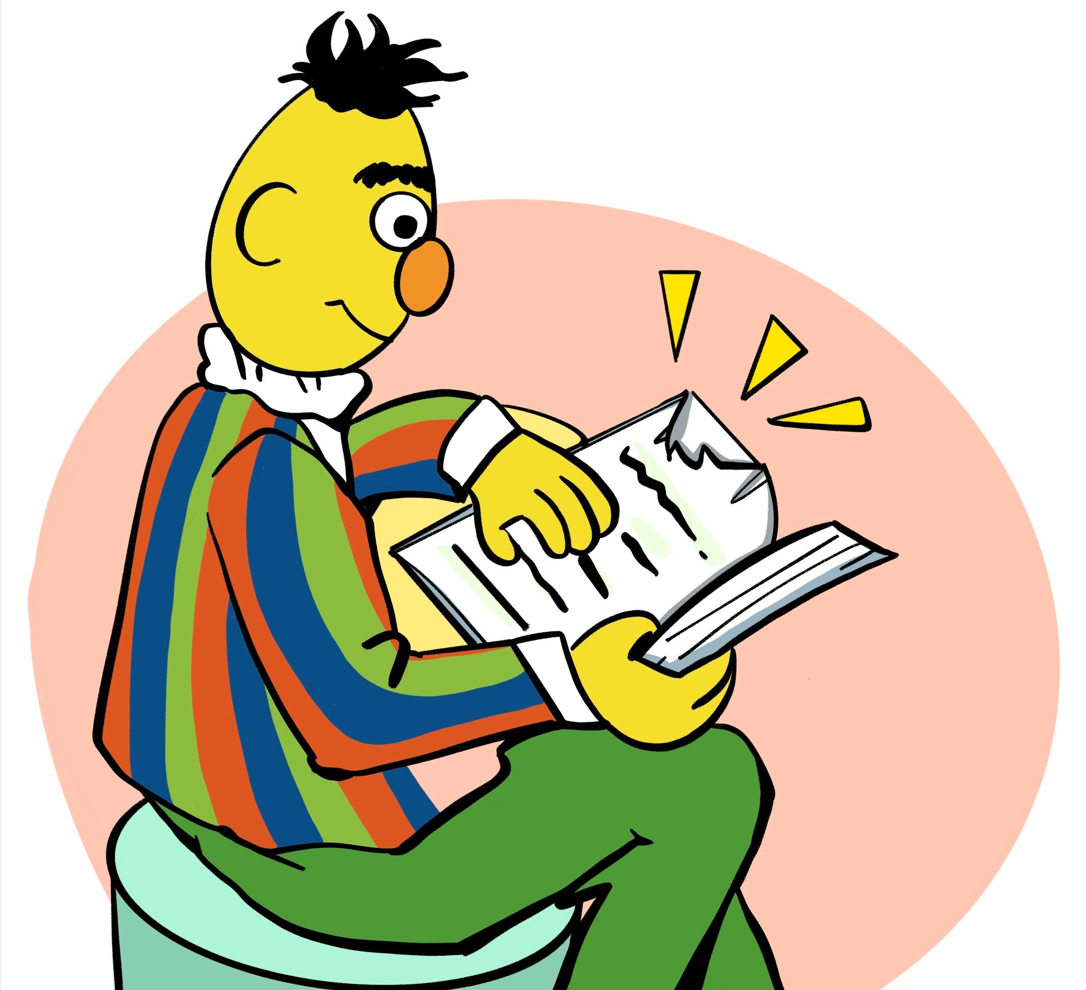}
        \caption{Semi-Supervised Fine-Tuning.}
        \label{fig:semi-supervised}
\end{wrapfigure}

Let $D_l$ be the set of labeled data points, $D_u$ be the set of unlabeled data points, and $D = D_l \cup D_u$ be the full dataset. 
Denote $P(y\mid x; \theta)$ as the language model with initial parameters $\theta_0$. 
The initial training of the model on the labeled data can be formalized as: 

\begin{align}
    \theta_1=\operatorname{argmax}_{\theta} \frac{1}{\left|D_l\right|} \sum_{(x, y) \in D_l} \log P(y \mid x; \theta)
\end{align}

The model is then used to make predictions on the unlabeled data: 

\begin{align}
    y_u=\operatorname{argmax}_y P\left(y \mid x_u; \theta_1\right)
\end{align}
where $y_u$ can refer to both generated labels (for sample matching) or output distribution (for distribution matching)~\citep{fu2023specializing}. 
The top $s$ fraction of the most confident predictions (with high probability) is added to the labeled data set: 

\begin{align}
    D_l^{\prime}=\left\{\left(x_u, y_u\right) \mid P\left(y_u \mid x_u; \theta_1\right)>\text { threshold }\right\}
\end{align}
which is finally used to 
train the model $P(y\mid x)$ in turn (\textbf{self-training}) or 
another model $P^\prime(y\mid x)$ (\textbf{semi-supervised knowledge distillation}).

\paragraph{Self-Training.} 
Self-Training uses the model-generated data to train the model itself, which is also known as bootstrapping. 
For example, 
BLIP~\citep{li2022blip} employs a captioner to generate novel captions derived from the given image, and utilizes a filter to eliminate noisy generations. 
Subsequently, the models are fine-tuned using the refined data obtained after filtering.
~\cite{huang2022large} generate multiple reasoning paths with the help of a language model, and then majority voting~\citep{wang2022self} is used to predict the most reliable answer. 
The reasoning paths with this answer are then used as synthetic training data for language model self-training.
~\cite{zelikman2022star} aim to train the language model to perform CoT reasoning in a semi-supervised setting, where the rationales are not always available. They first generate an intermediate rationale, and if it induces an incorrect answer, the model attempts to produce an inversely rationalized explanation. Subsequently, the model is fine-tuned using the question, the newly generated rationale, and the answer. 
Alpaca~\citep{alpaca} uses a set of 175 seed tasks, each comprising a single instruction and a single example, to enable the large language model to generate additional instructions and examples through In-Context Learning for self-training. 
This technique is known as Self-Instruct~\citep{wang2022selfinstruct}. 
TALM~\citep{parisi2022talm} uses a self-play way to improve performance. 
It generates a tool input based on the task input, and then incorporates the output of the tool and the tool input to produce the final task output. 
These synthetic data generated through self-play are employed to iteratively fine-tune the language model.

\paragraph{Semi-Supervised Knowledge Distillation.} 
Similar to self-training, semi-supervised knowledge distillation also leverages a model to annotate unlabeled data for model tuning.
However, instead of generating the synthetic data with the trainable model itself, semi-supervised knowledge distillation uses another teacher model for data generation, and the student model is trained with this new data. 
For example, 
~\cite{llmteacher} use large language models to generate chain-of-thought data, which is then used to fine-tune a smaller language model. 
~\cite{fu2023specializing} first use a LLM to generate responses for unlabeled questions and then use its output distribution to train the smaller student language model for specialization. 
~\cite{shridhar2022distilling} propose a method to disassemble a LLM into two smaller models, namely a problem decomposer and a problem solver, using a distillation approach.

\subsection{Active Learning\label{active-learning}}

\begin{wrapfigure}[14]{r}{7cm}
\centering
\vspace{-25pt}
 \includegraphics[width=1 \linewidth]{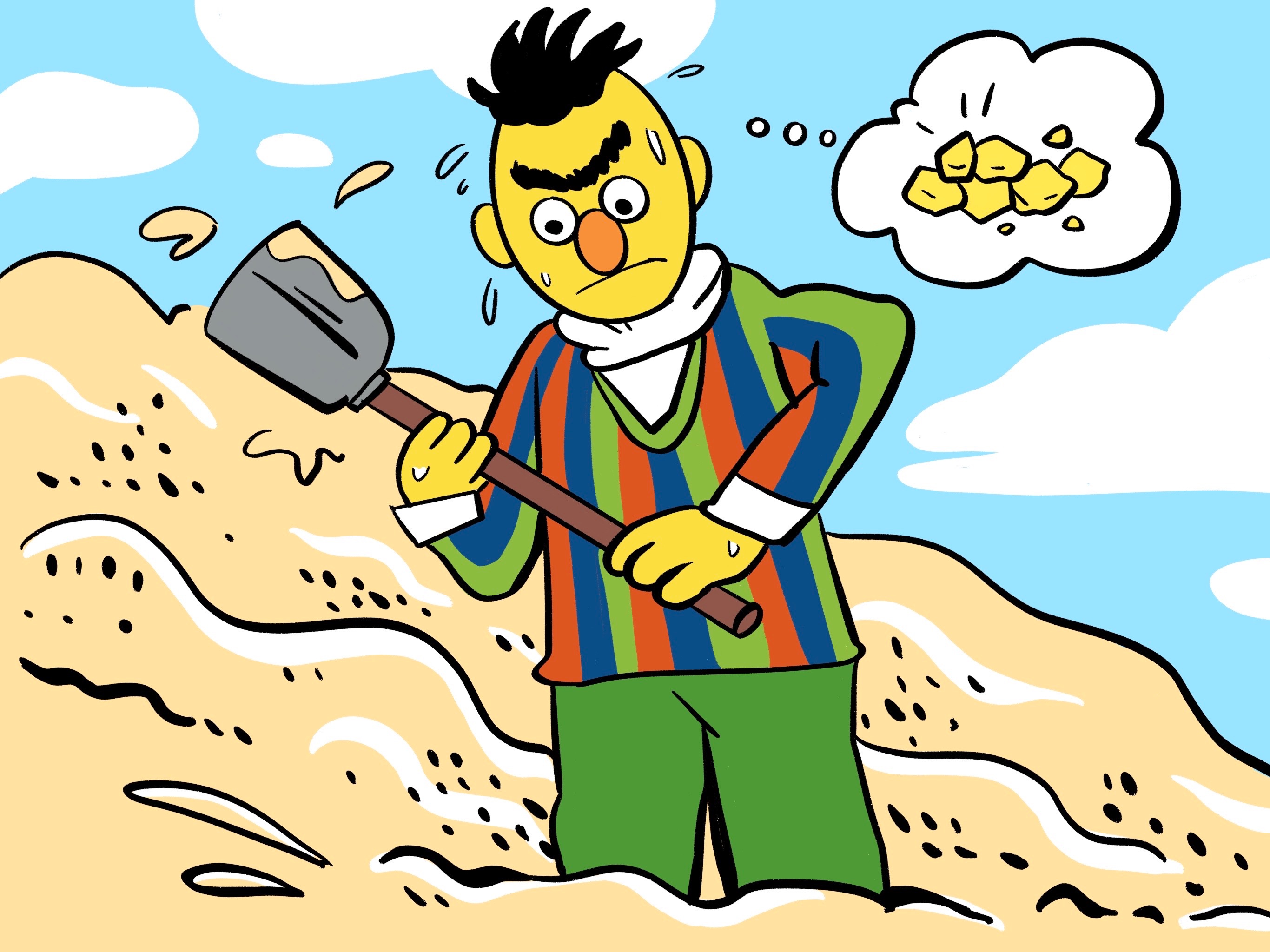}
        \caption{Active Learning.}
        \label{fig:active}
\end{wrapfigure}

Active learning (AL) is a machine learning approach where an algorithm selects a subset of unlabeled data points iteratively to be manually labeled by a human annotator or automatically labeled by an automated labeling system. The goal of this approach is twofold: (1) to obtain a larger and more desirable set of labeled data points that can be used to further train the model, and (2) to maximize the model's performance gain with minimal data expansion, thereby improving sample efficiency. The human annotator or automated labeling system is commonly referred to as the Oracle ($O(\cdot)$), which is an (interactive) object or a function that can label the data points~\citep{ren2021survey}. While traditional AL research has primarily focused on finding an optimal query strategy ($Q(\cdot)$) with humans as the oracle~\citep{ren2021survey, wang2021putting}, recent work has shown that other interactive objects such as LMs can also serve as oracles~\citep{dossou-etal-2022-afrolm}. 
Furthermore, AL involves selecting data points from an unlabeled data pool, which can include sources such as the internet~\citep{li2023internet} and knowledge graph~\citep{seo2021active}. 
The presence of interactive objects in the loop highlights the strong relationship between active learning and iNLP.

Formally, let $D_u$ be the unlabeled dataset. We aim to find the optimal subset $Q(D_u) \subseteq D_u$ of samples to be labeled. Subsequently, when we construct a newly labeled set $D_l=O(Q(D_u))$ or then add them to the existing labeled dataset, we aim to minimize the loss of the model $P(y|x; \theta)$ on the labeled dataset. This process is repeated iteratively. Thus, given an unlabeled data pool such as corpus, internet~\citep{li2023internet}, and knowledge graph~\citep{seo2021active}, the main effort on active learning is twofold: (1) find the optimal \textbf{query strategy} $Q(\cdot)$, and (2) choose a suitable \textbf{oracle} $O(\cdot)$.

\paragraph{Query Strategy.} 
\cite{zhang-etal-2022-survey} summarize two major concerns of AL query strategy design: informativeness and representativeness. 
Informativeness-based query strategies aim to identify unlabeled data points that can provide the maximum amount of additional information when being labeled, with the objective of maximizing the information gained at each iteration. 
Common informativeness-based query strategies include uncertainty sampling-based strategies~\citep{schroder-etal-2022-revisiting}, disagreement based-strategies~\citep{shelmanov-etal-2021-active}, and performance based-strategies~\citep{zhang-plank-2021-cartography-active,shen2021active}.
Representativeness-based query strategies aim to account for correlations among samples to avoid sampling bias and excessive weighting of outliers. 
Common representativeness-based query strategies include density based-strategies~\citep{zhao-etal-2020-active} and batch diversity based-strategies~\citep{yu-etal-2022-actune}. 
We refer the readers to ~\cite{zhang-etal-2022-survey} for more details. 
Traditional AL query strategies rely on statistical metrics~\citep{tong2000support,settles-craven-2008-analysis} that may lack representation richness compared to neural representations, particularly those based on PLMs. As such, there is a growing interest in exploring how to effectively leverage PLMs and LLMs in the design of AL query strategies. 
For example, 
several useful empirical conclusions on the combination of BERT~\citep{devlin2018bert} and traditional AL query strategies are illustrated in \cite{ein-dor-etal-2020-active}. 
\cite{seo2022active} propose a query strategy that is based on a task-independent triplet loss, which leverages task-related features provided by task classifiers and PLM-based knowledge features to enhance batch sample diversity. 
ALLSH~\citep{zhang2022allsh} designs a query strategy that is guided by local sensitivity and hardness utilizing a PLM-based representation, in order to improve the performance of prompt-based few-shot fine-tuning on various NLP tasks.
However, adapting PLM-based representations to AL query strategies poses several delicate challenges. 
For example, 
\cite{schroder-etal-2022-revisiting} point out that adapting state-of-the-art LLMs for query strategies can lead to prohibitive costs, even outweighing expected savings. They also conduct preliminary experiments that explore the combination of transformer-based models with several traditional uncertainty-based AL strategies. 
\cite{margatina-etal-2022-importance} suggest that a PLM-based representation that is fine-tuned using poor training strategies can be detrimental to AL performance. They highlight the significance of effectively adapting PLMs to the specific downstream tasks during the AL process. 
These challenges suggest that there is a vast research space for the NLP community to explore more efficient and effective AL query strategies. 
This effort may be in line with the roadmap of retriever methods (c.f., ~\S\ref{para:retrieval}), incorporating metrics that are specific to active learning. 
For example, \cite{zhang2022active} situate the RL-based retrieval method mentioned in ~\S\ref{para:retrieval} within the context of active learning (AL), with the ``additional gain'' upon acquiring a label for an example being the AL-specific metric. 

\paragraph{Oracle.} 
The oracle applied for labeling are usually humans in most previous NLP AL research~\citep{zhang-etal-2022-survey, wang2021putting}, including annotation of coreference resolution~\citep{li2020active}, word sense disambiguation~\citep{zhu-hovy-2007-active}, non-literal language identification~\citep{birke-sarkar-2007-active}, and especially small corpora construction with intensive professional knowledge or cost~\citep{peshterliev2018active,quteineh-etal-2020-textual,griesshaber-etal-2020-fine, maekawa2022low}. 
PLM, as an alternative, has shown considerable potential in playing a role as an oracle for AL. 
For example, 
\cite{yu2022actune} propose to actively annotate highly uncertain samples with pseudo-labels generated by PLMs and use low uncertainty data points for self-training. 
\cite{dossou-etal-2022-afrolm} adopt PLM to generate new sentences to enrich the corpora of low-resourced African languages for pre-training a multilingual model. 
\cite{wang2022selfinstruct} formulate the data augmentation problem as a relabeling mechanism. They iteratively collect a batch of data points with removed instruction or input-output example for relabeling, and then leverage PLM and ICL to generate new instruction labels or input-output labels to train the model. 
Furthermore, to the best of our knowledge, despite the scarcity of related work, other interactive objects such as knowledge bases, tools, and environments may have the potential to serve as oracles for active learning. 
For example, 
as a special type of knowledge base, ontology may contain structured information that can facilitate labeling~\citep{ye2022ontologyenhanced}, making it a potential candidate to serve as an oracle. 
ReAct~\citep{yao2022react} and PoT~\citep{chen2022program} integrate web searching and code execution, respectively, into the answer generation process. 
During the AL process, we can intuitively leverage these tools to label data points. 
MineDojo~\citep{fan2022minedojo} finds action guidance from an internet-scale knowledge base when faced with a difficult open-ended task (i.e., task without guidance label) in the \textit{MineCraft} environment. 
NLMap-SayCan~\citep{chen2022openvocabulary} equips the LLM with an open-vocabulary and queryable scene representation which can actively propose involved objects and their locations in the environment to label incomplete and non-reifiable planning of the LLM. 
These work highlight the potential for using knowledge bases, tools, and environments for labeling. 

\subsection{Reinforcement Learning\label{rl}}

\begin{wrapfigure}[14]{r}{7cm}
\centering
\vspace{-20pt}
 \includegraphics[width=1 \linewidth]{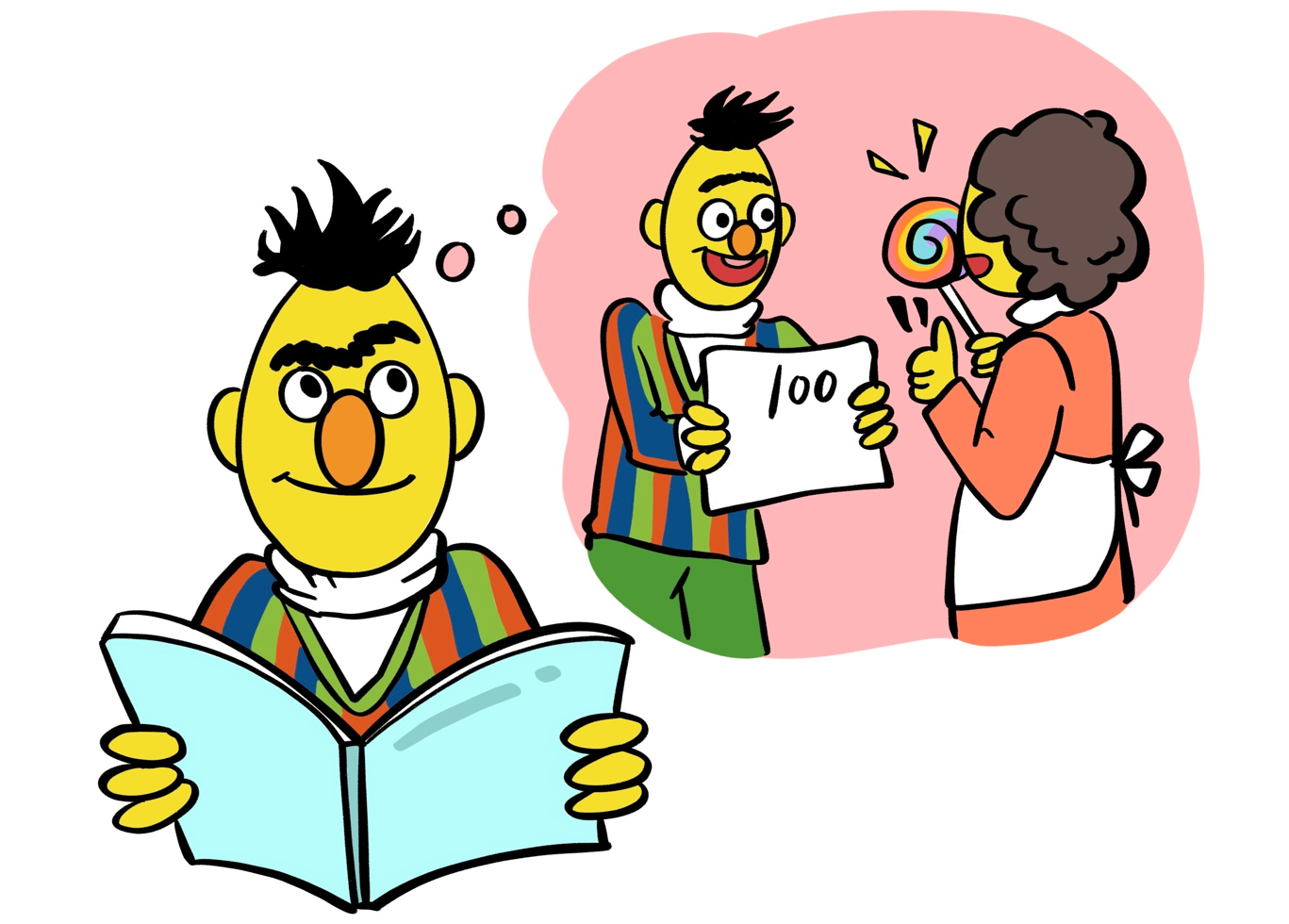}
        \caption{Reinforcement Learning.}
        \label{fig:rl}
\end{wrapfigure}

With the advent of reinforcement learning from environment feedback (RLEF)~\citep{ahn2022can, chen2022openvocabulary}, human feedback (RLHF)~\citep{christiano2017deep, ouyang2022training}, and artificial intelligence feedback (RLAIF)~\citep{bai2022constitutional,liu-etal-2022-aligning}, there has been a surge in reinforcement learning for natural language processing. 
In the context of Reinforcement Learning (RL), generated text or tool-use can be treated as an ``action'', while both surrounding text and non-text inputs can be viewed as ``observations''. 
The ``reward'' for the language model mainly includes human preference feedback~\citep{christiano2017deep, ouyang2022training} and affordance grounding feedback~\citep{ahn2022can, chen2022openvocabulary}. 
Therefore, our survey focuses on examining the interaction of language models with environments and humans. 
We also refer the readers to ~\cite{yang2023foundation} for more information. 
In the following part, we will provide a brief introduction to RL for iNLP, focusing on two aspects: (1) feedback loop, and (2) reward modeling.

\subsubsection{Feedback Loop} 
\begin{figure}[ht]
    \centering
    \begin{subfigure}[b]{0.35\textwidth}
        \includegraphics[width=\textwidth]{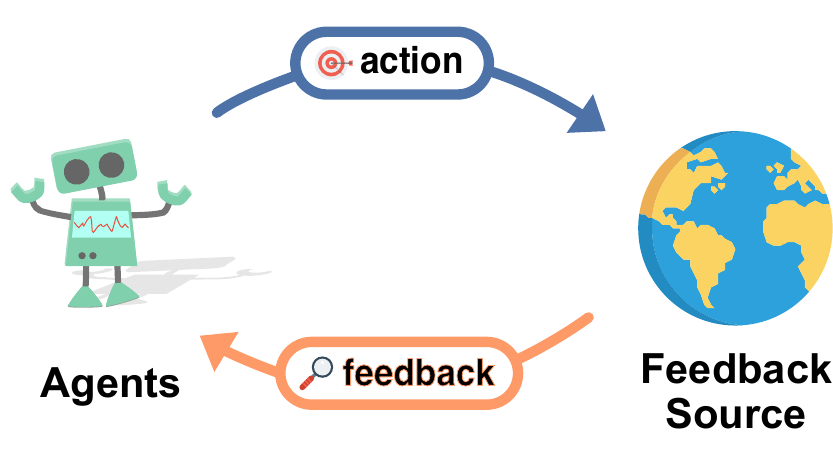}
        \vspace{10mm}
        \caption{Online RL.}
        \label{fig:rl-loop-online}
    \end{subfigure}
    ~ 
    \quad
    \begin{subfigure}[b]{0.5\textwidth}
        \includegraphics[width=\textwidth]{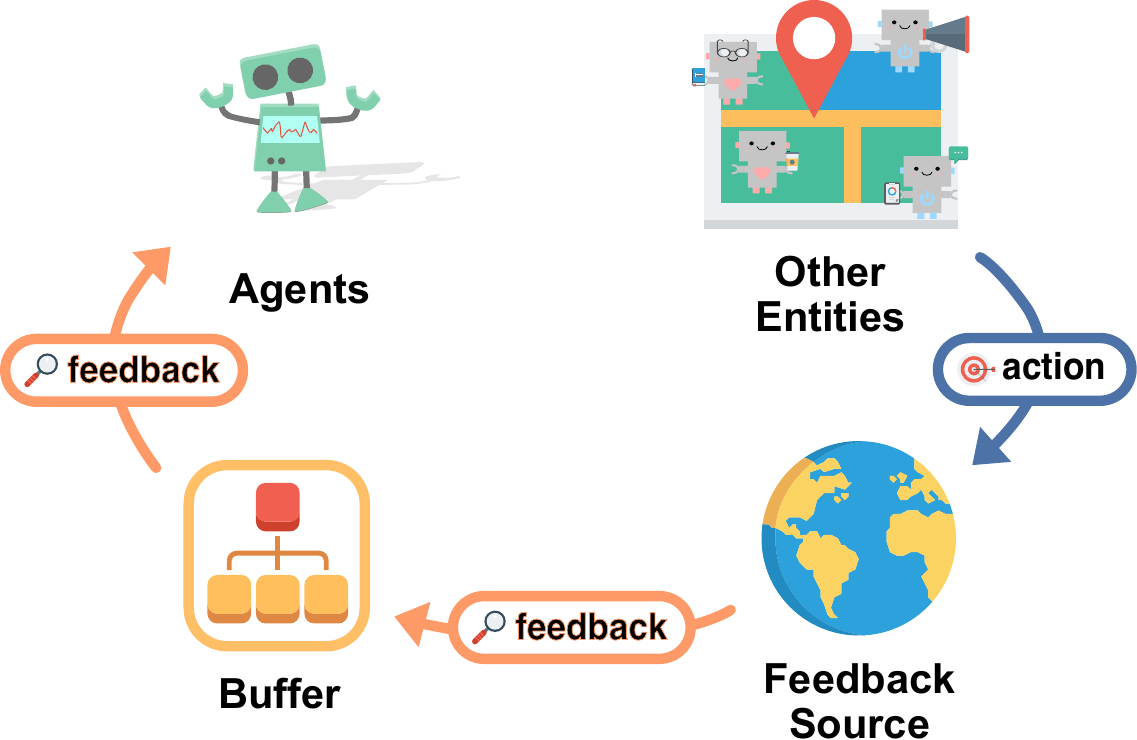}
        \caption{Offline RL.}
        \label{fig:rl-loop-offline}
    \end{subfigure}
    \caption{Online RL uses the actions of the training agent to generate feedback in real-time, or synchronously~\citep{prudencio2023survey}. This means that the agent interacts with the environment, takes actions, and immediately learns from the consequences of those actions. On the other hand, Offline RL, also known as batch RL, uses feedback from actions that were generated by other entities, not the training agent itself~\citep{levine2020offline}. This feedback is calculated and stored in a buffer, allowing the training agent to learn from it asynchronously.}
    \label{fig:rl-procedure}
\end{figure}

There are two main approaches to build RL frameworks with language models: online reinforcement learning and offline reinforcement learning. 
In these approaches, language models serve as RL agents or policies. 
As shown in Figure \ref{fig:rl-procedure}, 
online reinforcement learning involves updating models with real-time synchronous rewards during training~\citep{carta2023grounding,yu2022multimodal}, whereas offline reinforcement learning leverages rewards derived from a static data source~\citep{li2022pre}. 
The choice between online RL and offline RL depends primarily on practical scenarios. Generally, online RL is more suitable for LM-environment interaction as obtaining feedback from environments is a more automated process~\citep{yang2023foundation}. On the other hand, offline RL is more practical for LM-human interaction since human feedback may not always be readily available~\citep{fernandes2023bridging}. 
Consequently, while there are studies on online RL for LM-human interaction~\citep{wang2021putting} and offline RL for LM-environment interaction~\citep{yang2023foundation}, our survey mainly focuses on online RL for LM-environment interaction and offline RL for LM-human interaction\footnote{For the counterpart, we refer readers to~\cite{wang2021putting} and~\cite{yang2023foundation}.}.

\paragraph{Online RL.} 
Online RL empowers language models to tackle specific tasks by learning from real-time feedback provided by external entities such as the environment or more intricate reward models~\citep{carta2023grounding, fan2022minedojo, huang2022inner}. This feedback, similar to other RL models, is typically used as rewards for the online RL models. 
Standard online RL allows language models to learn from immediate rewards after each step. These dense rewards are often scalar values~\citep{carta2023grounding} or boolean values~\citep{huang2022inner} derived from the agent's state, observation, and outputs. 
For example, in tasks that necessitate continuous movement and multiple actions, RL agents can learn from real-time feedback during the process. 
\cite{abramson2022improving} sample multiple discrete time points during the agent's continuous movement and provides real-time binary feedback (positive or negative) based on the relative progress of the task. 
Moreover, in tasks with a long horizon, where each task involves multiple actions, sparse rewards can be employed to avoid the inefficiency of dense rewards. 
These rewards are given once per episode instead of after each step, as seen in studies like \cite{ahn2022can,yao2022webshop,yuan2023plan4mc}. 
For large-scale, production-ready online services, language models can be trained with buffered feedback and re-deployed periodically to minimize learning costs~\citep{help-harm}. 

\paragraph{Offline RL.} 
According to recent research~\citep{ouyang2022training, li2022pre}, offline reinforcement learning agents can utilize PLMs as the policy and share the same reward mechanism as online reinforcement agents while using different learning procedures. 
Specifically, they asynchronously apply feedback by fine-tuning the PLM rather than updating it in real time. 
During such a procedure, offline reinforcement learning with language models can utilize several algorithms
\footnote{Please note that these algorithms are not specific to offline RL but are frequently utilized in offline RL settings within the context of iNLP.}
to align language models with human preference. 
These algorithms include 
Proximal Policy Optimization (PPO)~\citep{schulman2017proximal, ouyang2022training}, 
Advantage Actor-Critic (A2C)~\citep{A2C, glaese2022improving}, 
Implicit Language Q-Learning (ILQL)~\citep{snell2022offline}, 
Trust Region Policy Optimization (TRPO)~\citep{schulman2015trust}, 
Natural Language Policy Optimization (NLPO)~\citep{ramamurthy2023is}, 
among others. 
These RL methods usually introduce alignment costs, resulting in performance degradation for other tasks, known as alignment tax~\citep{alignment-tax, ouyang2022training}. 
To address this issue, 
\cite{korbak2023pretraining} employ offline reinforcement learning during model pre-training, leading to better alignment while preserving superior performance. 
RAFT~\cite{dong2023raft} demonstrates that employing early stopping can find a more favorable balance between text quality and preference reward. 
Another line of works focus on improving the optimization algorithms such as computational efficiency~\citep{dong2023raft}, robustness~\citep{yuan2023rrhf}, and training stability~\citep{ramamurthy2023is}.

\subsubsection{Reward Modeling} 
Reinforcement learning agents are trained using rewards that are computed based on feedback from external entities, which primarily include interactive environments, and humans. 
Various feedback sources and mechanism contribute to the development of distinct reward models.

\paragraph{RL from Environment Feedback.} 
As discussed in ~\S\ref{Sec:Env-in-the-loop}, reinforcement learning from environment feedback (RLEF) facilitates affordance grounding of language models~\citep{ahn2022can}. 
For tasks with clear and easy-to-compare evaluation metrics, such as shopping and object arrangement tasks~\citep{yao2022webshop,fan2022minedojo,yu2022multimodal}, reinforcement learning agents can be optimized using absolute reward evaluated by corresponding environments and other evaluation models. 
Typically, reinforcement learning agents can be trained with binary reward functions, receiving positive reward if they conducted correct actions, and negative reward otherwise~\citep{huang2023grounded,huang2022inner}. 
Further, ~\cite{goyal2021pixl2r} involve evaluators that map visual environment and natural language descriptions to scale reward. 
Additionally, reinforcement learning agents can also receive rewards that are more complex and specifically handcrafted. 
For example, \cite{yao2022webshop} involve a carefully-designed reward function that considers various attributes of the chosen product and text-based instruction-product similarity, thereby reducing the need for human-in-the-loop evaluation. 
For tasks where it's hard to build such efficient and sensitive evaluation metrics that maps agents' states and actions to absolute scores and establish totally-ordered relations among all actions, such as text generation tasks, reinforcement learning agents can be optimized using comparative reward, such as the relative relationship between generated actions, where the policy is rewarded whether the generated action is better than the previous one~\citep{DBLP:conf/iclr/ZhouGXW020}. 
In addition to considering the relative relationship between actions within the same state and input, reinforcement learning agents in long-horizon tasks can also be rewarded based on the relative progress made in reducing the distance between the current state and the target state~\citep{yuan2023plan4mc}.

\paragraph{RL from Human Feedback\label{para-rlhf}.} 
Reinforcement Learning from Human Feedback (RLHF) is receiving increasing attention as a crucial post-training procedure for LLMs~\citep{ouyang2022training, glaese2022improving, yaofu-notion-blog, fernandes2023bridging}. 
For general-purpose text generation tasks, which is an open-ended task without deterministic answers, reward models are often preferred than handcrafted reward functions due to their higher sensitivity and better performance~\citep{fan2022minedojo,li2022pre, augmented-lm}. 
Specifically, the reward models for offline RL are trained with a small set of human feedback that is cost-effective to gather, and subsequently are used to reward reinforcement learning agents~\citep{bai2022constitutional, kiseleva2022iglu}. 
Similar to environment feedback, the reward mechanism of human feedback also varies among tasks. 
In embodied tasks, the reward model can observe the agent's history and provide binary feedback~\citep{abramson2022improving}. 
In natural language generation tasks, reward models simulate human annotators to label outputs and provide numeric scores~\citep{ouyang2022training, learning-to-summ, help-harm}. 
In more complicated tasks like reasoning tasks, reward models learn to follow specific rules and provide more complex and structural feedback~\citep{glaese2022improving}. 
Note that feedback isn't limited to simple binary or numeric forms. It can also be generated in natural languages. Language models can process this type of feedback, make adjustments, and ultimately produce corrected outputs~\citep{chen2023improving,scheurer2023training}. 
Further, the use of a reward model trained on human preference data can also be considered as ``RL from Model Feedback''. 
For example, 
~\cite{bai2022constitutional} propose ``RL from AI Feedback'' (RLAIF), where agents can self-adjust their outputs by providing self-feedback and self-prompts, resulting in higher-quality outputs. 
These AI feedback mechanisms essentially provide indirect human feedback as they are trained on data annotated by humans. 

\subsection{Imitation Learning\label{imitation-learning}}

\begin{wrapfigure}[20]{r}{7cm}
\centering
\vspace{-10pt}
 \includegraphics[width=0.7 \linewidth]{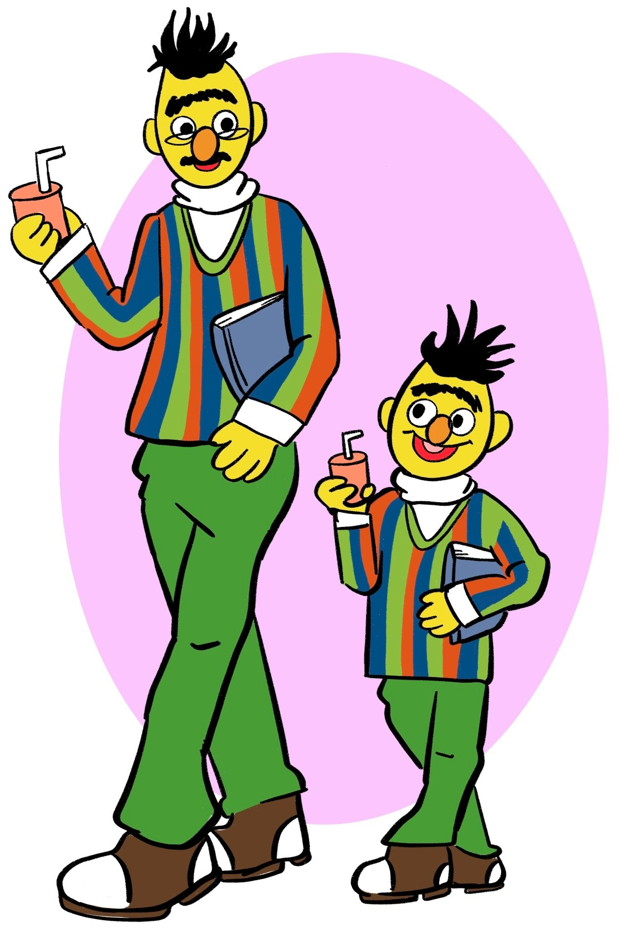}
        \caption{Imitation Learning.}
        \label{fig:imitation}
\end{wrapfigure}

Imitation learning enables an agent to learn a policy~\citep{pomerleau1988alvinn,10.1162/neco.1991.3.1.88,ross2011reduction} or a reward function~\citep{ng2000algorithms} by mimicking an expert behavior represented by given demonstrations. 
In contrast to reinforcement learning, a primary advantage of imitation learning is that it does not depend on a manually designed reward function or a learned reward model, but solely relies on behavior demonstrations. 
Such an advantage becomes especially prominent when accessing an expert at a low cost is possible, leading to a scalable generalization to various fields such as autonomous driving~\citep{bansal2018chauffeurnet}, computer control~\citep{pmlr-v162-humphreys22a}, game playing~\citep{10.1162/neco.1991.3.1.88,silver2016mastering,baker2022video}, human–agent interaction~\citep{team2021creating}, robotic learning~\citep{pmlr-v164-jang22a,lynch2022interactive,karamcheti2023language}, skill acquisition~\citep{10.1109/ICRA.2018.8461249,RoboImitationPeng20}, and even surgery~\citep{Tanwani_M2V_ICRA_20}. 
It has also been applied widely to tasks in NLP, including paraphrase generation~\citep{du-ji-2019-empirical}, textual adversarial attack~\citep{chen-etal-2021-multi}, text editing~\citep{agrawal-carpuat-2022-imitation,shi-etal-2022-text}, and text generation~\citep{hao2022teacher}.

In imitation learning, the goal is to learn a policy $\pi_\theta$ parameterized by $\theta$ that mimics the behavior of an expert policy $\pi_E$ in a given task. 
The behavior of the expert policy is represented by a set of demonstrations $D = \{(s_1, a_1), (s_2, a_2), ..., (s_T, a_T)\}$, where $s_t$ is the state at time step $t$ and $a_t$ is the action taken by the expert policy $\pi_E$ in that state. The objective of imitation learning is typically formulated as: 

\begin{align}
    \max _\theta \sum_{i=1}^T \log \pi_\theta\left(a_i \mid s_i\right)
\end{align} 

Imitation learning for interactive natural language processing can be divided into offline and online imitation learning.

\paragraph{Offline Imitation Learning.} 
Demonstrations can be collected and stored offline as pairs of state observations and corresponding expert actions. 
Directly training models on such datasets in a supervised learning manner frames the basic type of imitation learning, namely behavior cloning~\citep{pomerleau1988alvinn}. 
Numerous supervised text generation approaches can be classified into this group by reformatting the autoregressive decoding into a Markov decision process at either token~\citep{levenshtein-transformer,agrawal-carpuat-2022-imitation} or sequence~\citep{shi-etal-2022-text,faltings2023interactive} level. 
The learning process from the local expert's demonstrations can be considered an offline interaction. 
The expert policy is encoded in the data as an offline source to acquire the correct action given the current state. 
The model learns through offline interaction with the expert, and can later perform the task or behavior independently. 

\paragraph{Online Imitation Learning.} 
For some cases where the model can consult an expert, imitation learning can be conducted through an online interaction for additional supervision and evaluation. 
When incorporating online interaction or online imitation learning~\citep{ross2011reduction}, the trained model can be updated on-the-fly using feedback from the expert, thus improving its performance over time. 
In particular, the expert response can be simulated by a separate system that knows the goal state, which proves beneficial in scenarios where accessing a human expert is infeasible or impractical~\citep{faltings2023interactive}. 
This approach also provides the advantage of reinforcing or discouraging specific behaviors to aligning models with human expectations by adjusting the expert simulator.

Imitation learning often suffers from exposure bias, leading to distribution shifts and error accumulation in sequential decision-making tasks~\citep{ross2011reduction}, such as robotic control or text generation~\citep{6795228}. 
When the model is only exposed to the prior trajectories generated by the expert policy, it rarely experiences the state updated by the action of its own policy. 
This mismatch can lead to a distribution shift where the model may encounter states it has not seen during training, thus exacerbating error accumulation. 
To address exposure bias in imitation learning, researchers have proposed methods that alternate between the policy of the expert and that of the model being trained. 
Upon the model predicts actions, the expert provides feedback or corrections, allowing for fine-tuning~\citep{ross2011reduction}. 
These approaches share the same principle with interactive natural language processing which involves both offline and online interaction. 
How to eliminate exposure bias in language generation, especially when the online interaction with experts is limited, remains an area of active exploration~\citep{arora-etal-2022-exposure}. 
Furthermore, imitation learning has other limitations, including its reliance on the quality of expert demonstrations and the need for a large amount of demonstration data, impeding its broader application in interactive natural language processing.

\subsection{Interaction Message Fusion\label{fusion}}

In this subsection, we strive to provide a comprehensive and unified framing of the interaction message fusion methods, including all the methods presented in this section. 
Note that this framing also systematically categorizes the knowledge integration methods as mentioned in §\ref{Sec:KB-in-the-loop}.

\begin{figure}[h]
    \centering
    \includegraphics[width=300.0pt]{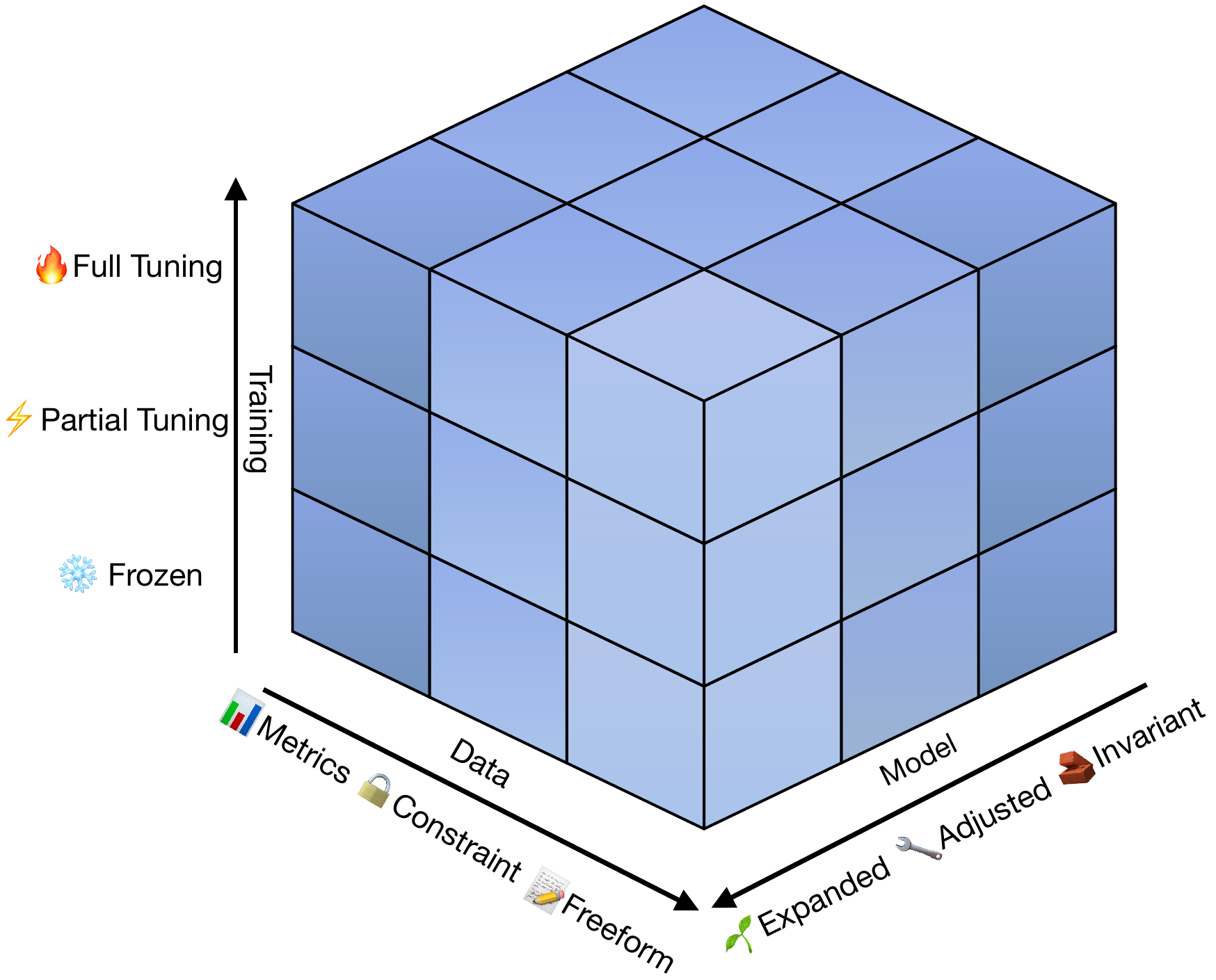}
    \caption{The three dimensions of interaction message fusion methods.}
    \label{fig:fusion}
\end{figure}

As illustrated in Figure \ref{fig:fusion}, the interaction message fusion methods can be divided into three dimensions, each having three categories. 
Thus, we have a total of 3 * 3 * 3 = 27 clusters of ways to incorporate interaction messages into language models. 
The following will present the basic definition and examples for each dimension. 

\paragraph{Along the data dimension.} 
(1) ``Metrics'' refers to simple signals such as scalar rewards or ranking scores. 
(2) ``Constraint'' means that the interaction message is in form of keywords, templates, skeletons, constrained vocabularies and the like for message formatting, conditioning, refinement as well as data augmentation, etc. 
(3) ``Freeform'' involves providing an unconstrained and unstructured text as an interaction message. 
For example, 
\begin{enumerate}
    \item \textbf{Metrics}: Most reinforcement learning methods are related to this (§\ref{rl}) as they rely on reward signals. For instance, RLHF~\citep{christiano2017deep, ouyang2022training} utilizes a feedback mechanism wherein the language model outputs are scored to indicate their quality and safety. 
    Several methods employ heuristics to filter interaction messages~\citep{li2022blip, hu2021knowledgeable}, which can also be considered related, since the boolean signals are leveraged. 
    \item \textbf{Constraint}: The constraint on the input end involves some conditional signals such as skeletons~\citep{wu2018response, cai-etal-2019-skeleton}, and outlines~\citep{yang2022re3}. 
    For instance, Re3~\citep{yang2022re3} generates an outline to condition long-text story generation. The constraint on the output end involves constrained decoding~\citep{shin-etal-2021-constrained, hokamp-liu-2017-lexically}, output refinement via a verbalizer which transforms the output~\citep{hu2021knowledgeable}, etc.
    \item \textbf{Freeform}: Most of the methods deal with freeform data, as defined by its unstructured nature.
\end{enumerate}

\paragraph{Along the model dimension.} 
(1) ``Model-Invariant'' approach does not modify any part of the model architecture. 
(2) ``Model-Adjusted'' approach makes changes to the existing modules of the model architecture. 
(3) ``Model-Expanded'' approach involves adding new modules to the model architecture. 
For example, 
\begin{enumerate}
    \item \textbf{Model-Invariant}: 
    Prompting methods (~\S\ref{prompting}) belong to this line. 
    Most of the tuning methods are also in this line of research. In most cases, the interaction message is concatenated or inserted to the input along sequence length dimension so that it is unnecessary to change the model architecture~\citep{guu2020realm, izacard2022atlas, art-tool, xie2022unifiedskg}. 
    We can also only modify some critical mediating modules of the language model to incorporate knowledge~\citep{meng2022locating, meng2022memit} with the model architecture unchanged. 
    \item \textbf{Model-Adjusted}: 
    VaLM~\citep{valm} replaces one layer of self-attention block with cross-attention block to fuse retrieved visual knowledge. 
    K-BERT~\citep{kbert} uses a superposition of text embedding and knowledge embedding (i.e., token embedding+soft-position embedding+segment embedding) and an attention mask matrix to fan in the knowledge.
    \item \textbf{Model-Expanded}: 
    K-Adapter~\citep{kadapter} adds additional adapters to the model for knowledge enhancement as mentioned in ~\S\ref{param-efficient}. 
    Retro~\citep{retro} adds cross-attention modules to the model for retrieval augmentation. 
    RelationLM~\citep{relational-memory} uses an additional GRU function to fuse knowledge into the language model. 
    KELM~\citep{lu2021kelm} uses additional hierarchical knowledge enhancement module to incorporate heterogeneous information from knowledge graphs and text into PLMs. 
    This module employs a relational Graph Neural Network (GNN) to dynamically representing inputted knowledge, an attention mechanism to resolve knowledge ambiguity, and a self-attention mechanism for further interactions between knowledge-enriched tokens.
\end{enumerate}

\paragraph{Along the training dimension.} 
(1) ``Frozen'' means that all the model parameters are fixed and cannot be tuned. 
(2) ``Partial Tuning'' means that only a part of the model parameters are tuned. 
(3) ``Full Tuning'' refers to updating all model parameters. 
For example, 
\begin{enumerate}
    \item \textbf{Frozen}: This category mainly involves prompting methods (~\S\ref{prompting}). 
    Besides, some constrained decoding-based methods~\citep{shin-etal-2021-constrained, hokamp-liu-2017-lexically} also do not update model parameters. 
    Another example is KPT~\citep{hu2021knowledgeable}, which refines the output verbalizer using metrics that consider the frequency and relevance of knowledgeable words. 
    \item \textbf{Partial Tuning}: For those which introduce additional modules or leverage a strict subset of model parameters to fan in external information~\citep{meng2022memit, meng2022locating, kadapter}, partial tuning is a trivial way as mentioned in ~\S\ref{param-efficient}. For instance, 
    K-Adapter~\citep{kadapter} only tunes the adapter modules with the main body of the language model frozen. 
    SpoT's generic variant~\citep{vu-etal-2022-spot} learns a source prompt on some source tasks with the language model frozen and then uses it to initialize another learnable soft prompt of the target task for knowledge transfer. 
    \item \textbf{Full Tuning}: Most methods require full tuning of the model parameters to integrate external information into the language model. For instance, 
    ToolFormer~\citep{toolformer} trains all the language model parameters on API call inserted corpus to enable the language model to use tools. 
    Kepler~\citep{wang2019kepler} tunes all the model parameters with both knowledge embedding loss and masked language modeling loss. 
    KnowBERT~\citep{knowbert} embeds structured knowledge from multiple knowledge bases into PLMs. 
    It uses a combined entity linker to obtain relevant entity embeddings and further enhances contextual word representations through word-to-entity attention, where the model parameters are also fully trained. 
\end{enumerate}

In summary, integrating external information into language models has emerged as a rapidly evolving research area in recent years, with numerous strategies available for its implementation.
By breaking down the interaction message fusion methods into three dimensions along the data, model, and training dimensions, we provide a systematic categorization of the existing methods, which can help researchers and practitioners better understand and design new approaches. 
It is worth noting that these dimensions are not mutually exclusive, and different methods can combine different categories from each dimension. 
The selection of an appropriate category relies on the specific task, data, and resources, as well as the trade-off between performance and computational efficiency.

\section{Evaluation\label{eval}}

Evaluation is undoubtedly important in tracking the NLP progress~\citep{DBLP:journals/csur/SaiMK23}, and even helps to outline future directions of NLP, e.g., Turing test~\citep{elkins2020can, srivastava2022beyond}. Existing surveys have comparatively elaborated on both automatic evaluation and human evaluation methods from general ~\citep{DBLP:journals/csur/SaiMK23} to task-specific metrics, e.g., evaluating controllable text generation~\citep{ctg_survey}. However, interactive NLP is slightly different from the general NLP tasks, as it devotes greater attention to the quality and effectiveness of model interactions with humans, environments, etc. In the interest of brevity and simplicity, this section will shed light on the relationship between iNLP and evaluation, yet does not retrace nor depict the historical evolution of general NLP evaluation. For further information, please refer to related NLP evaluation surveys~\citep{khapra-sai-2021-tutorial,DBLP:journals/csur/SaiMK23}. 
Please note that our survey primarily focuses on generative natural language processing, i.e., natural language generation (NLG).

Interactive scenarios occur naturally in various general natural language generation (NLG) tasks, such as dialogue~\citep{dialogues_1,dialogues_2,dialogues_3} and question answering~\citep{qa_1,qa_2}. However, a large number of existing NLG metrics mainly evaluate the non-interactive performance of an NLG model~\citep{Evaluating_Human_Language_Model_Interaction}, which ignores evaluating the interactive quality and effectiveness happening in the model inference stage. Specifically, these metrics focus on comparing the difference between the model completion with the pre-specified human reference to determine whether the performance is beyond previous models, such as Meteor~\citep{meteor}, Rouge~\citep{rouge}, and CIDEr~\citep{cider}. While they are convenient and time-saving, the interaction quality may not be improved if the model only resorts to feedback from such evaluation metrics, resulting in it being hardly fit for usage in the real world. For example, BLEU~\citep{bleu} is one of the most commonly used metrics for NLG systems. 
Instead of directly telling the generation quality of the system, it focuses on comparing the differences between the model output and reference by n-gram matching. In this case, a higher lexical similarity leads to a higher score. Thus, a sentence whose expression is not similar to the reference will get a low score, even if it meets the user's given preferences~\citep{DBLP:conf/acl/SellamDP20}. Although interaction evaluation is still in its fledgling stage, researchers are aware of such insufficiency in current NLG evaluation and are diverting their attention to the interaction performance of a NLG model.

In this section, we make the attempt to summarize recent progress of interaction evaluation. We split interaction evaluation into four types according to the interactive objects.

\subsection{Evaluating Human-in-the-loop Interaction} 
Evaluating human-in-the-loop aims to evaluate the performance of human-system interaction, which can be divided into two technical routes: general and task-specific metrics.

\paragraph{General Metrics.} 
These metrics are agnostic to specific tasks and primarily aim to evaluate the general performance of PLMs. The key challenge is to evaluate the alignment between the language model and humans, that is, whether the NLG model satisfies certain human preferences. To solve this problem, \cite{ouyang2022training} give a specific interpretation of this nebulous concept--a model is aligned if it is helpful (i.e., helps to solve the task), honest (i.e., ensures the authenticity of information), and harmless (i.e., conforms to ethics). In detail, they evaluate the helpfulness property by preference rating from human labelers, while the honesty property is automatically evaluated by a hallucination dataset and the TruthfulQA dataset~\citep{lin2021truthfulqa}. As for the harmlessness property, they use both Real Toxicity Prompts~\citep{realtoxicityprompts} and CrowS-Pairs datasets~\citep{nangia-etal-2020-crows} to evaluate the bias and toxicity of the NLG model. 
To further expand the dimensions of interaction evaluation, HALIE~\citep{Evaluating_Human_Language_Model_Interaction} proposes a human-language model interaction evaluation framework, which includes a user interface to facilitate the interaction between humans and language models. Based on this system, HALIE introduces metrics that expand the non-interaction evaluation in three dimensions: (1) \textbf{Targets}, evaluating extra interactive processes except model completions, such as user edits; (2) \textbf{Perspectives}, adding human evaluation involving users who come into direct interaction with PLMs; (3) \textbf{Criteria}, highlighting human preference when evaluating the model completions, e.g., enjoyment, and helpfulness. Comprehensive experiments in HALIE demonstrate a significant disparity between human evaluations conducted by third-party annotators and those provided by interactive system users. This disparity underscores the crucial need to investigate interaction-based evaluation methodologies.

\paragraph{Task-specific Metrics.} 
Task-specific metrics are concerned with designing a customized interaction evaluation method for generation tasks, including task-specific feature measurement when evaluating human-model interaction. 
Such task-specific metrics are most commonly found in dialogue system~\citep{dialogue_human_in_the_loop_2}, question answering~\citep{qa_human_in_the_loop_1}, creative writing~\citep{lee2022coauthor}, etc. 
For example, dialogue system is interested in assessing the quality of human-model interaction in specific conversation scenarios~\citep{dialogue_human_in_the_loop_2,dialogue_human_in_the_loop_3} and evaluating the performance gain of adding human feedback in the model inference stage~\citep{dialogue_human_in_the_loop_1}. In particular, the metrics of task-oriented dialogue interaction evaluation largely consist of three items: 
(1) the success rate of tasks~\cite{testoni-bernardi-2021-interplay}, such as successfully booking flights or movies; 
(2) dialogue turn size~\citep{Lu2020}, fewer numbers indicate that the dialogue agent is better at completing a task with less time cost; 
(3) the accuracy in the dialog state track (DST)~\citep{henderson-etal-2014-second}, i.e., determining if the key information is accurate throughout the conversation. 
Additionally, human experience and knowledge have also been employed in the evaluation of question-answering systems. 
For instance, \cite{qa_human_in_the_loop_1} involve human authors in the process to enhance the generator's ability to create a diverse set of adversarial questions. 
This approach enables a more comprehensive evaluation of the robustness of question-answering systems. 
As for creative writing, 
CoAuthor~\citep{lee2022coauthor} introduces a dataset capturing interactions between writers and LLM, and demonstrates its usefulness in exploring LLM's language, ideation, and collaboration capabilities for creative and argumentative writing. 

Despite the advancements made in Human-LM interaction evaluation metrics, the development of a comprehensive general benchmark remains largely unexplored \citep{wang2021putting,DBLP:journals/fgcs/WuXSZM022}. A unified benchmark and metric play a crucial role in evaluating models from an interactive perspective; however, designing and collecting such a benchmark pose significant challenges. One of the primary difficulties stems from the specificity of interaction evaluation, which necessitates detailed tracking and recording of the interactive process rather than merely collecting model completions. 
In addition, the interactive nature of human-in-the-loop evaluation can lead to data inconsistencies due to inherent individual diversity. This further complicates the task of constructing a general benchmark. Building such a benchmark necessitates careful consideration in terms of the user interface, means, and interaction process.

\subsection{Evaluating KB-in-the-loop Interaction} 
Evaluation of KB-in-the-loop interaction naturally arises in knowledge-augmented NLG models as a means of assessing their ability to acquire knowledge for enhanced generation. 

\paragraph{Knowledge Acquisition.} 
As discussed in \S\ref{Sec:KB-in-the-loop}, knowledge acquisition can be achieved through retrieving from external knowledge sources (e.g., knowledge graph, database, and even web browsing). 
Therefore, these KB-in-the-loop methods based on knowledge retrieval are interested in evaluating the effectiveness and efficiency of the knowledge retrieval process. 
For example, 
REALM~\citep{guu2020realm} conducts ablation experiments comparing different retrievers used for knowledge acquisition, using ``top-5 recall'' metrics. 
Atlas~\citep{izacard2022atlas} analyzes how the frequency of the text in the correct answer option appearing in retrieved passages is influenced by the number of retrieved passages. 
Such an evaluation is useful for analyzing and improving the intermediate component of KB-in-the-loop NLG systems.

\paragraph{Knowledge-Enhanced Generation.} 
Another line of research focuses on evaluating the models' capabilities for knowledge-enhanced generation. 
The most common and important way lies in hallucination detection, which utilizes extra knowledge graphs or databases to detect factual errors~\citep{hallucination-survey}. 
For example, Knowledge F1~\citep{knowledge_f1} measures the word overlap between the model’s completion and the fact knowledge used to ground the dialogue. 
Several fact-checking benchmarks have been presented in recent years to facilitate this area. For example, BEGIN~\citep{BEGIN} introduces three types of knowledge-grounded dialogue responses based on the evidence from Wikipedia. In this case, factual knowledge is in the form of paragraph-level text. 
In contrast, DialFact~\citep{dialfact} utilizes manually annotated instance-level knowledge snippets as evidence. 
The Attributable to Identified Sources (AIS) framework~\citep{ais} assesses whether the statements produced by an NLG system can be attributed to a specific underlying source. 
Another type of evaluation method focuses on the comparison of the differences between humans and models when choosing to add additional knowledge. 
For example, WebGPT~\citep{nakano2021webgpt} evaluates model completions by comparing them with human-written answers based on the same web-browsing environment to determine whether the model's retained knowledge is different from that of users. 
GopherCite~\citep{menick2022teaching} computes the ratio at which the model produces plausible and supported claims (i.e., evidence) in human evaluation.

\subsection{Evaluating Model/Tool-in-the-loop Interaction} 

As discussed in \S\ref{model-tool-in-loop}, model/tool-in-the-loop encompasses three types of operations: (1) thinking, (2) acting, and (3) collaborating. 
Thus, evaluating model/tool-in-the-loop interaction involves assessing the following aspects: (1) (multi-stage) chain-of-thought capability, (2) tool-use ability, and (3) collaborative behavior analysis.

\paragraph{Chain-of-Thought Capability.\label{eval-cot}} 
Chain-of-Thought (CoT)~\citep{Wei2022ChainOT} is regarded as one of the most crucial emergent abilities, which tends to become more pronounced as model size scales up~\citep{kaplan2020scaling, wei2022emergent}. 
It is frequently triggered by elicitive prompts (\S\ref{elicitive-prompting}) and can be instantiated through prompt chaining (\S\ref{prompt-chaining}). 
Typically, CoT aims at improving models' multi-step reasoning abilities~\citep{Wei2022ChainOT, reasoning-survey}. 
Thus, some reasoning-intensive benchmarks such as GSM8K~\citep{math-verifier}, DROP~\citep{dua-etal-2019-drop}, and ScienceQA~\citep{scienceqa}, can be used to evaluate CoT capability. 
We refer the readers to ~\cite{reasoning-survey} for a more comprehensive survey on the reasoning tasks. 
In our survey, we introduce additional perspectives focused on intermediate steps or reasoning trajectories for evaluating the CoT abilities. 
For example, we can assess the error propagation between reasoning steps. 
~\cite{DBLP:conf/emnlp/DuaG0G22}'s successive prompting technique encounters three primary error sources: incorrect answer predictions, incorrect question decomposition, and out-of-scope reasoning types. By understanding and addressing these errors happened in the intermediate reasoning steps, we can develop more robust models that mitigate the impact of errors throughout the multi-step reasoning process. 
Another approach is to analyze the effects of varying the number of reasoning steps or the number of reasoning branches. 
For example, 
~\cite{zhou2022least} conduct experiments to investigate the influences of different numbers of reasoning steps. 
~\cite{wang2022self} compare the influence of different numbers of sampled reasoning paths on performance.

\paragraph{Tool-Use Ability.} 
Following ~\cite{yang2023foundation} and ~\cite{li2023apibank}, evaluating the tool-use ability of language models involves examining: 
(1) \textbf{tool-use triggering}: whether the LM can determine when to use tools, 
(2) \textbf{tool-use accuracy}: whether the LM selects the correct tools to perform a specific task, 
(3) \textbf{tool-use proficiency}: whether the LM can effectively and efficiently use the tools for reasoning and decision making~\citep{yao2022react}. 
For example, 
Toolformer~\citep{toolformer} evaluates tool-use triggering by calculating the API invoking rate, that is, how often the model requires the corresponding API to help with some text generation tasks such as machine translation, question answering, and mathematical calculation. 
ART~\citep{art-tool} compares various strategies employed by language models to select prompts from a task library for correct tool-use prompting, which can also be viewed as a measurement of tool-use accuracy. 
WebShop~\citep{yao2022webshop} examines the success rate of language agents in utilizing web tools for shopping tasks, which serves as an indicator of the tool-use proficiency of language models. 
Since tool-use ability has been shown to be capable of enhancing the reasoning capability of language models in Tool-Augmented Learning settings~\citep{qin2023tool}, certain metrics used to evaluate CoT ability (\S\ref{eval-cot}) can also be employed to assess tool-use proficiency~\citep{chen2022program, math-verifier, yao2022react}. 

\paragraph{Collaborative Behavior Analysis.} 
The analysis of collaborative behaviors of language model agents can be divided into \textbf{result-oriented analysis} and \textbf{process-oriented analysis}. 
Result-oriented analysis mainly focuses on the output or the final result of the collaboration. 
For example, 
MindCraft~\citep{bara-etal-2021-mindcraft} measures whether a task can be solved through collaboration between two language model agents. 
Socratic Models~\citep{zeng2022socratic} evaluates the success rate of tasks that are solved through closed-loop interaction involving a large language model, a vision-language model, and an audio-language model. 
BIG-bench~\citep{srivastava2022beyond} has incorporated several tasks for evaluating Theory of Mind (ToM) abilities. 
Furthermore, ~\cite{kosinski2023theory} demonstrates that ChatGPT has achieved a performance level comparable to that of nine-year-old children on certain ToM tasks. 
Process-oriented analysis is more concerned with the dynamics of the collaboration itself. 
For example, 
MindCraft~\citep{bara-etal-2021-mindcraft} also assesses the communication efficiency by measuring the number of dialogue exchanges required to accomplish a task.

\subsection{Evaluating Environment-in-the-loop Interaction} 
Since the objective of environment-in-the-loop NLP primarily focuses on utilizing language model agents for embodied tasks, which significantly differ from typical text generation tasks, the evaluation of LM-Env interaction may require a distinct paradigm. 
Generally, the research in this field primarily focuses on two main aspects: (1) constructing embodied task platforms, and (2) determining the appropriate evaluation metrics. 
We also refer the readers to \S\ref{embodied-ai} and \S\ref{games} for more information. 

\paragraph{Embodied Task Platforms.} 
To name a few, 
\textit{EvalAI}~\citep{DBLP:journals/corr/abs-1902-03570} provides an open-source platform for evaluating models and agents, which involves tasks of visual question answering, visual dialog, and embodied question answering~\citep{DBLP:conf/cvpr/DasDGLPB18}. 
\textit{SAPIEN}~\citep{DBLP:conf/cvpr/XiangQMXZLLJYWY20} introduces a simulated household environment specifically designed to simulate daily objects and their interactions within a household setting. 
Similarly, \textit{Alexa Arena}~\citep{gao2023alexa} introduces a user-centric simulation platform for Embodied AI models and presents an instruction-following benchmark based on human-agent dialogue. 
\cite{ALFRED} build a benchmark, \textit{ALFRED}, to evaluate the ability of agents to transform visual observations and textual instructions into action sequences for everyday tasks. 
Building upon this setup, \cite{ALFREDL} introduce a new test set called \textit{ALFRED-L}. This test set is created by modifying instructions from examples in the \textit{ALFRED} validation splits. The modifications involve eliminating the need for actions manipulating objects and incorporating directives to approach more objects. 
Furthermore, \cite{alfworld} extend both \textit{TextWorld}~\citep{cote2019textworld} and \textit{ALFRED} platforms to create \textit{ALFWorld}. 
This novel platform establishes a connection between textual descriptions and commands, and the physical simulation of embodied robots, offering an enhanced alignment between them.
Lastly, \cite{bara-etal-2021-mindcraft} explore collaborative situations where agents are placed in interactive scenarios. They aim to model the mental states of participants in these collaborative interactions. To facilitate this research, they introduce \textit{MindCraft}, a fine-grained dataset that includes the beliefs of partners when collaborating on tasks within the virtual world of \textit{Minecraft}, a 3D environment consisting of blocks. 
We refer the readers to \cite{yang2023foundation} for more examples. 

\paragraph{Evaluation Metrics.} 
To name a few, 
the mission success rate and the average number of robot actions are both metrics that can be utilized to assess the model's capabilities in terms of interacting with the environment~\citep{gao2023alexa}. The mission success rate measures the model's effectiveness by computing the average proportion of missions in which all the goal conditions are met, indicating successful completion of the mission. While the average action number examines the model's efficiency by recording the average number of actions performed by the robot for each task. 
\cite{ahn2022can} employ the plan success rate and the execution success rate to evaluate the performance and effectiveness of the system. The plan success rate measures the system's ability to generate appropriate plans for given instructions, while the execution success rate assesses the system's capability to successfully execute those plans and accomplish the specified tasks. 
\cite{huang2023grounded} report the text instruction generation speed in the inference stage (referred to as ``token count'' in the paper) to evaluate the efficiency of the involved large language model.

\section{Application\label{apps}}

\subsection{Controllable Text Generation\label{ctg}}

The Controllable Text Generation (CTG) technique is an NLP approach that empowers language models to generate text that is not only coherent and meaningful but also allows users to control particular aspects of the output. 
The need for CTG arises from the desire to customize generated text according to user-defined constraints, such as length constraint~\citep{li2022diffusion, zhou2023controlled}, inclusion of particular keywords~\citep{hokamp-liu-2017-lexically, carlsson-etal-2022-fine}, and adherence to a specific sentiment or style~\citep{dathathri2019plug, contrastive-prefixes}. 
Conventional CTG methods typically involve training models with explicit objectives that optimize for the control attributes~\citep{hu2017toward, li2022diffusion, keskar2019ctrl, lu2022quark, clive2021control, zhou2023controlled}, constrained decoding~\citep{hokamp-liu-2017-lexically, anderson-etal-2017-guided, post-vilar-2018-fast, lu-etal-2021-neurologic, lu-etal-2022-neurologic, qin2022cold, kumar-etal-2022-gradient}, and prompting~\citep{zou2021controllable}. 
We refer the readers to ~\cite{zhang2022survey} and ~\cite{weng2021conditional} for more information. 
In this section, we briefly discuss the potential applications of iNLP in CTG. 

\textbf{Interacting with humans} can enhance controllability in CTG by enabling users to directly provide their preferences and constraints during the generation or training process. 
AI Chains~\citep{ai-chains}, for example, allows users to chain together LLM steps and modify them in a modular way, improving transparency, controllability, and collaboration. 
~\cite{Evaluating_Human_Language_Model_Interaction} suggest the importance of controlling the intermediate generation process rather than just the final output, and highlight the need to consider more control attributes related to first-person subjective experience and user preferences. 
~\cite{christiano2017deep, lu2022quark, ouyang2022training, yaofu-notion-blog} imply the use of RL or RLHF for controlled text generation, which enables not only control over helpfulness, harmlessness, and honesty, but also allows potential optimization to meet length constraints and other criteria. 

\textbf{Interacting with knowledge bases} can potentially enhance the robustness and factuality of CTG, 
as suggested by \cite{working-memory}, who propose Knowledge-Aware Fine-Tuning (KAFT) to improve the controllability of language models while maintaining their robustness. KAFT fine-tunes LMs using a combination of the vanilla supervised dataset and augmented data, which includes instances with counterfactual contexts (i.e., contexts that contradict the model's memorized knowledge) and irrelevant contexts (i.e., contexts that are unrelated to the task). 

\textbf{Interacting with models and tools} may have the potential for more complex and fine-grained control over text generation. 
For example, 
classifier-guided CTG approaches put a classifier in the loop to provide control signals or feedback~\citep{dathathri2019plug, li2022diffusion}. 
Similar to Diffusion-LM~\citep{li2022diffusion} which iteratively denoises the text with the control feedback from a classifier at each iteration, Self-Refine~\citep{madaan2023selfrefine} lets a LLM generate an output and then provide multi-aspect feedback on it. 
This feedback is used to iteratively refine the output until it reaches a satisfactory quality or a specific criteria. 
Notably, typical classifier-guided CTG relies on external classifiers, while Self-Refine employs the LLM itself as a classifier through self-interaction. 

\textbf{Interacting with environments} inherently requires great controllability due to the essential need for affordance grounding, as discussed in ~\S\ref{Sec:Env-in-the-loop}. SayCan~\citep{ahn2022can}, as a representative example, leverages a scoring mechanism over action candidates to achieve such controllability. 

Overall, various interactive objects may offer different avenues for optimizing CTG systems. By harnessing the power of interaction, we can achieve more user-oriented, robust, fine-grained, complex, and even reality-oriented control over text generation. 

\subsection{Writing Assistant}
Intelligent and interactive writing assistants constitute a rapidly growing area of research that explores the potential of AI-powered tools to modify, enrich, and even co-create content with humans. These assistants can be broadly categorized into four types based on their level of involvement in the content generation process: (1) Content Supporting, (2) Content Checking and Polishing, (3) Content Enrichment, and (4) Content Co-creation.

\paragraph{Content Support.} 
Content supporting writing assistants do not generate content for use, but just provide functional assistance for writers such as on-the-fly summarization~\citep{dang2022beyond} and real-time visualization~\citep{singh2022hide}. 
For example, 
~\cite{arnold2021generative} propose an interaction scheme where human writers are provided with questions for inspiration instead of content snippets for use. 
~\cite{dang2022beyond} propose a writing assistant that continuously updates summaries, keywords, and central sentences of existing content for user reference, rather than generating content directly. 
~\cite{singh2022hide} design a writing assistant offering visual and aural suggestions as writing supports. 
Although content supporting writing assistants provide minimal aids for human writers, they benefit from avoiding the dominance over the writing process in some cases, which is one of the main challenges of PLM-based writing assistants~\citep{arnold2021generative, jakesch2023co}. 
Moreover, writing support may reduce the manual effort to trigger or manipulate the writing assistant such as heavy prompt engineering~\citep{dang2023choice}. 
Specifically, 
~\cite{dang2023choice} indicate that manual effort is required for non-diegetic prompting, and thereby humans tend to prefer selecting suggestions from writing assistants over controlling aotomatic content generation through non-diegetic prompts. 
~\cite{jakesch2023co} point out that human writers' content creation can even be affected by opinionated PLM-powered writing assistants. 
These challenges can be mitigated by reducing the level of involvement of writing assistants to content supporting.

\paragraph{Content Checking and Polishing.} 
The processing procedure for content checking and polishing typically involves taking manually written sentences as input and producing output that has been grammar-checked and rephrased, allowing users to interactively and iteratively improve their writing. 
Famous real-world products include QuillBot\footnote{\url{https://quillbot.com/}}, 
Effidit~\citep{shi2022effidit}\footnote{\url{https://effidit.qq.com/en}}, Pitaya\footnote{\url{https://www.facebook.com/Mypitaya/}}, Grammarly\footnote{\url{https://www.grammarly.com/}}, and Xiezuocat\footnote{\url{https://xiezuocat.com/}}. 
There is also growing interest in incorporating iterative editing operations into writing assistants~\citep{kim-etal-2022-improving,du-etal-2022-read,du-etal-2022-understanding-iterative}, which can be considered an application of editing-based iNLP (~\S\ref{chap-edits}). 
For example, 
~\cite{kim-etal-2022-improving} suggest incorporating transfer learning from other text editing tasks to improve the quality of iterative text revision by linking editing actions to content quality. 
~\cite{du-etal-2022-read} raise a novel human-in-the-loop iterative text revision system that combines model-generated revisions with human judgments and specifically fine-tuning a PEGASUS model~\citep{zhang2020pegasus} as a revision generation model with which a revised sentence is generated based on a given sentence and an edit intention.

\paragraph{Content Enrichment.} 
Content enrichment, unlike content checking and polishing, involves more creative content generation but still relies on manually provided context or configuration. 
Classic content enrichment features include text completion (\textbf{AutoCompletion})~\citep{van-etal-2020-automets,sun-etal-2021-iga,li-etal-2021-gwlan,casacuberta-etal-2022-findings}, and keywords-to-sentence (\textbf{K2S})~\citep{miao2019cgmh,sha-2020-gradient,nie2022lexical,zheng-etal-2022-knowledge}. 
Note that both AutoCompletion and K2S simply supplement manual input, rather than co-creating new content from scratch through manual collaboration or guidance. 
AutoCompletion is an interactive writing assistant feature that involves humans in the content generation process by providing suggestions to complete their prompts, thereby enhancing their overall writing experience. 
For example, 
~\cite{sun-etal-2021-iga} propose an Intent-Guided Authoring (IGA) Assistant, which follows fine-grained author specifications to process the input text for AutoCompletion. 
The scheme proposed in IGA is similar to the recent trend of instruction tuning (~\S\ref{instruction-tuning}), which suggests that more complex and controllable user preferences in writing assistants can be formatted as instructions to further activate instruction-tuned LLMs. 
AutoCompletion can also be adapted to various NLP downstream tasks, including medical text simplification~\citep{van-etal-2020-automets}, human-computer collaborative translation~\citep{li-etal-2021-gwlan}, and interactive word completion of morphologically complex low-resource language~\citep{lane2020interactive}. 
Moreover, K2S is highly in line with controllable text generation (c.f. ~\S\ref{ctg}), but places a greater emphasis on controllable interactivity. 
Practical K2S applications typically allow users to customize the fine-grained control attributes according to their specific needs and preferences in an interactive manner. 
For example, 
CueBot~\citep{h-kumar-etal-2022-cuebot} proposes a conversational assistant capable of generating responses that can be controlled by users using cues/keywords. It suggests responses for users to choose from and incorporates a keyword loss during training to generate lexically constrained outputs. 

\paragraph{Content Co-creation.}
Content co-creation refers to the collaborative process between humans and AI systems to generate new content from scratch, rather than simply improving existing content. 
Content co-creation is widely explored in interactive fiction writing~\citep{manjavacas-etal-2017-synthetic,tapscott-etal-2018-generating}, screenplays and theatre scripts writing~\citep{mirowski2022co}, academic writing~\citep{fokcan}, and poem writing~\citep{astigarraga-etal-2017-poets,oliveira2017co,hamalainen-2018-poem}. 
For example, 
~\cite{tapscott-etal-2018-generating} models story generation as simulating role-play games and tracing player interaction sequences. 
~\cite{yang2022doc} develop \textit{DOC}, which includes a detailed outline generator and a detailed controller, significantly improves the coherence of long story generation. 
~\cite{chakrabarty2022help} proposes \textit{CoPoet}, an interactive poem writing assistant powered by instruction prompts and LLM, and verify that co-created poems are usually preferred compared to those written without \textit{CoPoet} involved. 
Dramatron~\citep{mirowski2022co} adapts hierarchically controlled PLMs to allow expert writers to control style tags, logic lines, character descriptions, and environment descriptions. This enables writers to easily generate the necessary material for various use cases. 
However, while practical writing assistants for content checking, polishing, and enriching have become increasingly mature, content co-creation writing assistants still face various challenges that need to be addressed. 
For example, 
~\cite{ippolito2022creative} point out that current NLG techniques for story generation often exhibit poor performance in maintaining the author's voice and the coherence of the storyline. 
~\cite{ai-chains, wordcraft, chen-etal-2022-balanced} demonstrate that there is often a trade-off between controllability and creativity in the generated content. 
Moreover, 
the evaluation of content co-creation-based writing assistants can also be particularly challenging due to the subjective nature of creative writing. 
Despite various efforts to construct reliable benchmarks for evaluation~\citep{lee2022coauthor,zhou2022ai,shen2023parachute}, ~\cite{mirowski2022co,ippolito2022creative} suggest that professional writers have become increasingly important for evaluation compared to crowd-sourcing annotators due to the ever-improved quality of artificial intelligence generated content (AIGC). 

\subsection{Embodied AI\label{embodied-ai}}

Embodied AI enables language models to impact the real-world and virtual environments through which agents observe and update states of themselves and their surroundings. 
One method for bridging language models with the physical world is interaction with grounded language as mentioned in ~\S\ref{Sec:Env-in-the-loop}, which allows language models to see, listen, and control external objects.

\textbf{Observation and Manipulation} are fundamental to many embodied tasks, where agents acquire external states and perform actions to update those states. Thanks to text descriptions and textual controlling interfaces, language models usually observe their surrounding environment through input text and operate on objects by sending textual commands. For instance, a visual perception mapper converts visual input to text in natural language~\citep{zhao2023chat}. Additionally, human intervention can be part of agent observation, so that agents can be guided by real-time human feedback~\citep{lynch2022interactive}. Typical observation and operation tasks include object rearrangements~\citep{huang2022inner}, tool usage~\citep{art-tool}, item creation and modification~\citep{jiang2021talk,elgohary-etal-2021-nl}, and other robotic controlling tasks. 

\textbf{Navigation and Exploration} enable agents to move around and study their surrounding environment by using dynamic observation and manipulation. That is, unlike observation and manipulation tasks, navigation and exploration tasks allow agents to move within the environment to adjust their observation and manipulation. These agents not only plan routes and actions, but also combine observations collected from different locations to make decisions, answer questions and reason, allowing them to accomplish complicated tasks that require multi-location multi-object observation and long-horizon manipulation. Text commands in both natural languages and programming languages~\citep{huang2023audio} bridge the gap between language model agents and available actions and tools. During such process, these agents also combine different data sources, including cameras, microphones, other sensors~\citep{gan2020look}, and textual commands from human controllers~\citep{sharma2022correcting,gao2022dialfred}. Moreover, agents can also work as assistants to guide human operations. For instance, an interactive driving assistant can continuously observe the driving environment and guide human drivers to handle various situations~\citep{ma-etal-2022-dorothie}.

\textbf{Multi-Role Tasks} require agents to cooperate and compete with humans and other agents to reach specific goals. Unlike agents with multiple skills, agents with social capabilities usually observe others' behaviors and communicate through textual messages, including messages in natural languages and data in more structured styles. Typical social tasks include multi-player gaming~\citep{suh2021development,lai2022werewolf}, human-AI collaboration~\citep{krishnaswamy-alalyani-2021-embodied,puig2020watch}, multi-agent collaboration~\citep{patel2021interpretation}, and other communication tasks, such as interview~\citep{xiao2020tell}, negotiation~\citep{verma-etal-2022-chai}, recruitment~\citep{nawaz2019artificial}, and opinion gathering~\citep{bittner2019bot}. In text-based gaming tasks, agents learn from human behaviors and play as human players~\citep{xu-etal-2022-perceiving}. In multi-agent environments, agents coordinate with each other to accomplish complex tasks that cannot be accomplished by any single agent~\citep{bara-etal-2021-mindcraft}. Agents also act as human delegates and communicate with others to complete day-to-day tasks, such as restaurant reservations and appointment scheduling~\citep{o2019google}. MetaAI's Cicero~\citep{diplomacy} enables language model agents to play in an online Diplomacy league.

\subsection{Text Game\label{games}}

Text games, also referred to as interactive fiction games~\citep{osborne-etal-2022-survey}, are capable of understanding player commands, simulating player states, and updating the current status of game environments~\citep{osborne-etal-2022-survey}. 
Language models have shown great potential in these game-playing scenarios~\citep{meta2022human, kramar2022negotiation, fan2022minedojo, yuan2023plan4mc}, which are a specific type of Embodied AI (c.f., ~\S\ref{embodied-ai}). 
Specifically, language models can be used to play or power text games through text-based interfaces, such as state descriptions~\citep{sironi2021adaptive}, commands~\citep{tennenholtz2019natural,ammanabrolu2020graph,zhang2022danli}, situated dialogue~\citep{bara-etal-2021-mindcraft}, and multi-party dialogue~\citep{park2023generative}. 
Thus, text games are intrinsic applications of iNLP, in which the environment or other agents are involved in the game-playing loop. 
We can divide text games into two distinct categories: (1) text-only games which rely solely on text, and (2) text-aided games which use text as a supplement to other forms of media, such as graphics or audio. 
In this subsection, we will begin by discussing interactive text game platforms. We will then provide a brief overview of how language models are utilized to play text-only games and to power text-aided games.

\paragraph{Interactive Text Game Platforms.} 
Interactive Text Game Platforms provide a framework and engine for building and running text-based games, often including features such as game state tracking, parser-based natural language understanding, and scripted events. 
Some examples of such platforms are: 

(1) \textbf{Text Adventure Games} are games that allow players to interact with adventurous worlds solely through textual descriptions and actions~\citep{ammanabrolu2019toward}. 
~\cite{osborne-etal-2022-survey} summarize two major text adventure game platforms: \textit{TextWorld}~\citep{cote2019textworld} and \textit{Jericho}~\citep{hausknecht2020interactive}. 
Additionally, they define seven major challenges that need to be addressed in developing solutions for Text Adventure Games, including partial observability, large state space, and long-term credit assignment, among others.

(2) \textbf{Social Deduction Games} are games where players attempt to discover each other's hidden role or team allegiance through strategic conversations, logical deduction, and deceitful actions\footnote{\url{https://en.wikipedia.org/wiki/Social_deduction_game}}. 
For example, classic examples of social deduction games include 
\textit{Werewolf}\footnote{\url{https://en.wikipedia.org/wiki/Mafia_(party_game)}},\textit{Mush}\footnote{\url{https://en.wikipedia.org/wiki/Mush_(video_game)}}, 
\textit{SS13}\footnote{\url{https://en.wikipedia.org/wiki/Space_Station_13}}, 
and \textit{Among Us}\footnote{\url{https://en.wikipedia.org/wiki/Among_Us}}. 
Specifically, 
~\cite{lai2022werewolf} propose a multimodal dataset containing text and visual signals to model persuasion behaviors in \textit{Werewolf}. 
~\cite{lin2020automatic} is another \textit{Werewolf}-based corpus with self-revealing and role-estimation behavior annotation. 
~\cite{tuin2021automatically} construct a corpus aimed at player role detection, based on the game \textit{Among Us}, and verify that it is a challenging yet learnable task.

(3) \textbf{Strategic Games} are games that heavily rely on player decision-making skills and situational awareness to determine the outcome\footnote{\url{https://en.wikipedia.org/wiki/Strategy_game}}. 
For example, \textit{Diplomacy}\footnote{\url{https://en.wikipedia.org/wiki/Diplomacy_(game)}} is a strategic board game that involves multiple players who each assume control of the armed forces of a European power. 
The objective of the game is to move one’s units skillfully and defeat those of opponents in order to gain possession of a majority of strategically important cities and provinces referred to as “supply centers.” 
The contested nature of this gameplay often requires players to engage in extensive and complex interactions and diplomacy with each other in order to achieve their goals.
\textit{Diplomacy} is gaining increasing attention and is widely regarded as a benchmark for autonomous agents' ability to communicate and adjust strategies like humans, which is one of the essential elements for the success of human civilization~\citep{kramar2022negotiation}. 
Cicero~\citep{meta2022human} proposes an impressive autonomous agent that combines PLM with RL and achieves human-level performance in \textit{Diplomacy}. 
Moreover, 
~\cite{kramar2022negotiation} make preliminary investigations on how negotiation algorithms and the inclination to punish traitors can enable autonomous agents to communicate like humans and cooperate more effectively in \textit{Diplomacy}. 
Apart from Diplomacy, there are numerous classic strategic games that serve as potential resources for interactive text game platforms and related NLP research, such as 
\textit{Eurogame}
\footnote{\url{https://en.wikipedia.org/wiki/Eurogame}}, 
\textit{Warhammer Fantacy}
\footnote{\url{https://en.wikipedia.org/wiki/Warhammer_(game)}}, and \textit{Paths of Glory}
\footnote{\url{https://en.wikipedia.org/wiki/Paths_of_Glory_(board_game)}}. 

(4) \textbf{Tabletop role-playing games (TRPGs)}\footnote{\url{https://en.wikipedia.org/wiki/Tabletop_role-playing_game}}, such as \textit{Dungeons and Dragons (DND)}\footnote{\url{https://en.wikipedia.org/wiki/Dungeons_\%26_Dragons}} 
and \textit{Call of Cthulhu (COC)}\footnote{\url{https://en.wikipedia.org/wiki/Call_of_Cthulhu_(role-playing_game)}}, 
as well as works of fiction, such as the \textit{Harry Potter series}~\citep{chen2022would}, have provided a rich source of situated and multi-party dialogue data that can be used to build challenging text game platforms~\citep{callison2022dungeons, rameshkumar2020storytelling, zhou2022ai, peiris2022synthesis}. 
However, the raw dialogue data documenting the game process of TRPGs is usually a mixture of in-character action descriptions and out-of-character strategy explanations~\citep{callison2022dungeons}, frequently accompanied by lengthy world-building documents~\citep{zhu2023fireball}, which can differ from one game to another. 
The problem of extracting the golden-standard game states and game commands remains a challenging yet fascinating question~\citep{zhu2023fireball}. 
For example, 
~\cite{rameshkumar2020storytelling} provide 34,243 summary dialogue fragment pairs from raw dialogue data documenting the \textit{DND} game process. 
The summaries in these summary-dialogue chunk pairs contain text descriptions of game states, which can serve as a good benchmark for abstractive game-state summarization of interactive text games. 
~\cite{callison2022dungeons} frame \textit{DND} as a dialogue system challenge, comprising both deterministic elements like dice rolls and imprecise descriptions of the game-play as partial state information. 
~\cite{zhou2022ai} introduce a novel and highly interactive task, \textit{G4C} (Goal-driven Guidance Generation in Grounded Communication), based on \textit{DND}. They train an autonomous agent acting as a game host, also known as a \textit{Dungeon Master} (\textit{DM}), using the theory of mind and RL. This approach significantly enhances the players' capacity to achieve their objectives. 

(5) \textbf{Life Simulation Games} are games that enable players to control one or more virtual characters\footnote{\url{https://en.wikipedia.org/wiki/Life_simulation_game}}. 
Classic life simulation games include \textit{Virtual Pet}\footnote{\url{https://en.wikipedia.org/wiki/Virtual_pet}}, \textit{Black and White}\footnote{\url{https://en.wikipedia.org/wiki/Black_26_White_(video_game)}}, \textit{MineCraft}\footnote{\url{https://en.wikipedia.org/wiki/Minecraft}}, and \textit{GTA Series}\footnote{\url{https://en.wikipedia.org/wiki/Grand_Theft_Auto}}. 
For example, 
~\cite{bara-etal-2021-mindcraft, fan2022minedojo, yuan2023plan4mc} explore how autonomous agents based on PLMs can learn to collaborate, communicate, and generalize across a range of tasks and objectives using \textit{MineCraft} as the interactive game platform. 
Furthermore, 
\textbf{Social Simulation Games} are a sub-genre of life simulation games that simulates social interactions and relationships between multiple artificial characters or lives in a virtual world\footnote{\url{https://en.wikipedia.org/wiki/Social_simulation_game}}. 
For example, \textit{the Sims Series}\footnote{\url{https://en.wikipedia.org/wiki/The_Sims}} is one of the classic social simulation game. 
~\cite{mehta2023improving} enhance the capability of AI agents to identify when they require additional information to enable more human-AI interactions and improve social simulations. 
As noted by~\citep{camel,park2023generative,wei2023multi}, the role-playing agents and open sandbox worlds can be easily adapted as factors in social simulation games. 
~\cite{park2023generative} configure PLM-based autonomous agents with social identity settings and conduct social simulations accordingly. 
Their experiment design is highly in line with the gameplay of \textit{the Sims Series}. 

\paragraph{Playing Text-Only Games.} 
Early work on autonomous agents of text-only games mainly relies on handcrafted reward functions~\citep{yuan2018counting} or other well-formatted data structures, such as knowledge graphs, to preserve and retrieve past information and game states~\citep{ammanabrolu2020graph}.
Although some exploratory methods before the emergence of PLMs also adapt trivial neural representations of text to help detect actions and states from text-only games, these methods focus on constrained hand-crafted template-based state and action space and are unable to understand complex and highly unstructured texts in many text-only games.
\cite{yao2021reading} point out that these methods, based on constrained hand-crafted template-based state and action space, isolate autonomous agents from understanding the meanings of words or semantics by verifying that agents without understanding semantics can achieve similar performance on these text-only games by adopting similar methods but without understanding semantics.
For designing better text-only games as testbeds for autonomous agents' ability of language understanding, the motivation and strategy implicitly expressed in words should not be detected through hand-crafted templates without understanding the semantics~\citep{yao2021reading,li2022immersive}.
The following work turns to autonomous agents based on PLM-powered neural representation which rely less on manual effort. 
For example, 
~\cite{yin2020zero} adapt sentence-level semantics representation-based clustering and deep Q learning~\citep{mnih2013playing,mnih2015human} for playing text adventure games. 
~\cite{xu2020deep} propose a lightweight transformer-based representation learning framework for text-only games and outperform previous SOTA methods. 
Recently, 
the success of LLMs has enabled text-only games to explore handling any user input~\citep{toddtowards,li2022immersive}. 
Understanding user inputs that are complex and ambiguous in their meaning requires careful attention to the actions and states explicitly described in the text. 

\paragraph{Powering Text-Aided Games.} 
Traditionally, in text-aided games, autonomous agents have used either formal language or structured natural language to model state transitions and execute actions using a highly symbolic representation~\citep{branavan2009reinforcement,vinyals2017starcraft}. 
With the emergence of PLMs, these agents transformed from using symbolic representations to using contextual and neural representations, which capture more complex and high-level semantics of textually stated strategies and communication protocols. 
As a result, we will elucidate the ways in which language interfaces and PLMs enable autonomous agents in text-aided games to communicate with other agents and make better game plans.
Since the impressive release of ChatGPT, some industrial and academic researchers have also been exploring the adaptation of LLMs to enhance the text-aided game experience. \textit{Inworld}\footnote{\url{https://www.inworld.ai/}} claims that LLMs can empower characters in games with distinct personalities and contextual awareness that stay in-world
\footnote{The in-world context of a player is the context that the player has with characters controlled by other players and NPCs when she/he is doing role-playing.} 
or on-brand\footnote{The on-brand context of a player is the context that the player has with other players and the game host when she/he is not doing role-playing.}, which tremendously improves the games' immersive experience. 
\textit{NetEase} also announces that they allow Non-Player Characters (NPCs) powered by LLMs in its online game, \textit{Nishuihan}, to communicate with considerable freedom~\footnote{\url{https://www.youtube.com/watch?v=zGVR5gPgefk}}. 
In addition to communication, language interface and NLP-powered planning have been explored in designing autonomous agents in text-aided games for reward shaping~\citep{goyal2019using}, instruction following~\citep{tuli2022learning,chen2020ask}, control policy generalization~\citep{pmlr-v139-hanjie21a}, and representation learning~\citep{karamcheti2023language}. 
In addition, language interface plays a major role in training autonomous text-aided game agents. 
For example, 
\cite{havrylov2017emergence, wong2022deep} point out that an efficient language-based communication protocol is crucial to a collaboration strategy in multi-agent text-aided games.
~\cite{NEURIPS2019_0af78794} suggests that language can naturally compose different sub-skills to enrich the non-compositional abstraction of complex text-aided games' hierarchical strategy. 
Moreover, \citep{reid2022can,li2022pre} verify that language modeling induces representations, which are even useful for offline RL strategy modeling of games.
This observation implies a relationship between the high-level semantic coherence of languages and the planning strategy adopted by text-aided games.

\subsection{Other Applications}

\paragraph{Specialization.} 
It refers to the process of adapting and customizing the ability of language models to specific tasks or domains. As assumed by ~\cite{fu2023specializing}, language models with strong modeling power may be effective across a wide range of tasks, but their performance on any individual task may be less impressive due to the distribution of their capabilities, implying the needs for LM specification. 
Although some domain specific language models are proposed for fields such as law~\citep{lawgpt}, healthcare~\citep{wang2023huatuo, wang2023chatcad}, material science~\citep{xie2023large} and finance~\citep{wu2023bloomberggpt}, they are mainly fine-tuned from large corpus of specific domain. 
iNLP, on the other hand, can provide another solution for LM specialization. That is, by observing and interacting with external objects such as medical records, legal documents, financial statements, technical specifications, domain-specific knowledge graphs or even domain-specific tool sets, language models can provide professional information to users. 
For instance, in medicine, iNLP can be used to retrieve relevant information from patient records and suggest potential diagnoses or treatments. In law, iNLP can help lawyers draft legal documents and contracts by providing suggestions based on retrieval augmentation from previous cases and legal precedents. 
Therefore, equipping language models with domain-specific interactive objects can implement the specification of them with higher data efficiency and computation efficiency.

\paragraph{Personalization.} 
It refers to the process of tailoring a language model's behavior and output to the unique needs and preferences of each user. 
This can be achieved through the model's interactions with users, learning from their inputs, demographics, and adapting its behavior accordingly
\footnote{\url{https://www.exponentlabs.io/articles/chatgpt-and-personalization-how-ai-is-changing-the-way-we-interact-with-technology}}. 
For example, 
\cite{rao2023can} suggest that ChatGPT has the potential to become more personalized and customized through learning from user interactions and individual preferences. 
\cite{salemi2023lamp} introduce a personalization benchmark and suggest to personalize LLMs through retrieval augmentation using user profiles. 
\cite{wu2022personalized} show the potential of personalizing PLMs through the use of prompts. 
\cite{madaan2022memory} personalize the PLM via an external memory with human feedback. 
Personalization can greatly enhance the user experience with language models by providing more preferred responses, improving the model's ability to understand the user's needs and intentions, and ultimately building trust and rapport between the user and the model. 
However, we should also be aware of the drawbacks brought by personalization. 
For example, 
\cite{deshpande2023toxicity} demonstrate that assigning a persona to ChatGPT can potentially magnify its toxicity up to six times.

\paragraph{Model-based Evaluation.} 
Model-based Evaluation enjoys the benefits of PLMs to compute a text quality score for each generated sample and show greater correlation with the human evaluation compared with statistical-based evaluation metrics, such as BLUERT~\citep{DBLP:conf/acl/SellamDP20}, BERTScore~\citep{DBLP:conf/iclr/ZhangKWWA20}, and COMET~\citep{DBLP:conf/emnlp/ReiSFL20}. 
Such an evaluation method can be widely implemented and used in various general NLG tasks, and even works for reference-free settings~\citep{DBLP:conf/aaai/Zhou020, DBLP:journals/corr/abs-2204-13346, DBLP:journals/corr/abs-2301-09008}. 
Additionally, existing preliminary research has indicated that LLMs have emerged with the ability to evaluate the AI generated content (AIGC) with human-like judges~\citep{gilardi2023chatgpt,wang2023chatgpt,chen2023exploring}.
Some papers also propose the fine-grained analysis of LLMs' ability to evaluate AIGC in specific NLG tasks, including summarization~\citep{luo2023chatgpt,gao2023human}, question answering~\citep{he2023annollm}, news outlet generation~\citep{yang2023large}, and translation~\citep{lu2023error}.
Moreover, \cite{liu2023gpteval,he2023annollm} propose to collaborate with LLMs to provide better and cheaper human-like evaluations. 
Through interactions between models and even humans, we can evaluate LMs in a more effective (accurate) and efficient (automatic) manner. This evaluation process can be akin to a teacher model administering ``exams'' and ``grading'' to assess the performance of a student model~\citep{hoeve2021towards}.

\section{Ethics and Safety\label{ethics}}

LLMs have demonstrated remarkable capabilities to understand, interpret, and generate human-like text. A plethora of LLM-based applications has emerged and been adopted prevalently in our daily lives. 
As a result, the utilization of these models also presents profound challenges across many societal domains. 
Therefore, it is crucial to consider the ethical implications of using LLMs, especially around the impact on education,  bias and fairness, privacy, harmful content and misinformation. 

\paragraph{Impact on Education.} 
The advent of LLMs, exemplified by ChatGPT, has introduced substantial challenges to the existing education systems.
One primary concern is the misuse of ChatGPT for academic assignments such as writing essays and solving scientific problems, which has raised deep concerns among K-12 educators, who perceive it as a potential threat to the education system \citep{chatgpt-edu-2023}. 
To address this issue, plagiarism detection tools such as GPTZero \footnote{\url{https://gptzero.me}}, AI Classifier \footnote{\url{https://platform.openai.com/ai-text-classifier}}, and DetectGPT \footnote{\url{https://detectgpt.com}} have been developed for detecting AI-generated content. 
Most of these AI detection tools focus on perplexity (text randomness) and burstiness (use of non-common terms).
Nevertheless, these tools have yet to demonstrate their effectiveness in capturing AI-generated content in a real-world setting.
Last but not least, computer-assisted writing tools, including ChatGPT, have limited capacities to assist users in learning and acquiring writing skills and principles. Their primary focus is on enhancing productivity rather than facilitating skill development, which is crucial for educational purposes. 

\paragraph{Social Bias.} 
As language models are typically trained with large-scale web corpus, it becomes highly susceptible to societal biases. It is known to further amplify the discrimination \citep{Leino-feature-bias-2018}, including the potential downgrading of resumes \citep{amazon-resume} and the generation of texts that contain stereotypes, toxicity, and racism \citep{hutchinson-etal-2020-social}.
Resume ``whitening'' has always been an issue where job applicants are forced to hide their identity as a minority gender, racial, religion, or region group to land a job interview.
This problem still exists even though many companies started to use AI-supported tools to rank and filter resumes.
Social bias in word embeddings is reflected when the word \textit{man} is closer to \textit{programmer} compared to \textit{woman} and \textit{programmer} \citep{10.5555/3157382.3157584}. 
Applications that utilize pretrained word embeddings for downstream tasks such as classification and analysis will then obtain results with social bias, causing fairness issues of the output.
~\cite{hutchinson-etal-2020-social} use toxicity prediction and sentiment analysis to assess language models' bias towards people with disabilities.
Results showed that the sentence \textit{I am a person with mental illness} and \textit{I will fight for people with mental illnesses} is more toxic than \textit{I am a tall person}. 
HERB~\citep{li-etal-2022-herb}, a bias evaluation metric which utilizes bias in a sub-region to evaluate language model's bias in a region on contextualized level, brings researchers' attention to not only focus societal bias in the whole of a region but also sub-regions. 
The aforementioned findings suggest that societal biases observed in language models could serve as an indication that stereotypes should be addressed and mitigated,
rather than leaving them to harm the minorities.
Considering that everyone now can easily access LLMs, biases should be filtered out or mitigated to ensure that they are not amplified and further affect people's thoughts in making decisions such as recruiting and assessing individuals.

\paragraph{Privacy Concern.} 
Large Language Models (LLMs) also raise concerns regarding user privacy. Training these models necessitates access to large amounts of data, often entailing the personal details of individuals. This information is typically derived from licensed or publicly accessible datasets, and can be utilized for a range of purposes, such as deducing geographical locations from phone codes in the data. There are already studies showing the possibilities of distilling sensitive information from large language models through prompting~\citep{carlini2021extracting,carlini2023extracting}. In the era of interactive NLP, humans a more actively interacting with these foundation models, which could potentially lead to more frequent user information leakage. Therefore, it is pressing to establish relevant policies for collecting and storing personal data. Furthermore, the practice of data anonymization is crucial to maintain ethical standards in dealing with privacy matters. There have been some pioneering studies that investigate privacy-preserving issues~\cite{li2023privacy,shi2022just}. We believe that more research efforts should be dedicated to privacy preservation in large language models, which will play a central role in the era of interactive Natural Language Processing (iNLP). 

\section{Future Directions\label{future}}

\paragraph{Alignment.} 
Alignment for language models can be categorized into factual alignment and value alignment. Factual alignment requires the model to tell it is not capable of answering the question when it does not perpetuate the needed knowledge~\citep{kadavath2022language}. However, factual alignment is challenging in practice since 1) it is hard to verify what knowledge has been contained in the pre-trained model~\citep{lin2022teaching}, and 2) we still lack a convenient knowledge editing method to update certain knowledge while do not impair others~\citep{de2021editing,meng2022locating}. Future work can consider developing tools to detect the ``blind spot of knowledge'' by analyzing the probability confidence in the model predictions, and efficient approaches to edit the knowledge in pre-trained models at scale.
For value alignment, existing work mainly focuses on using human~\citep{ouyang2022training} or AI feedback~\citep{bai2022constitutional} to train a reward model as the proxy of human judgment. During training, this reward model will continuously interact with the generative LM to enhance desired behaviors and inhibits undesired ones~\citep{liu-etal-2022-aligning}. RLHF is the representative approach in this manner, which has been widely used in products such as OpenAI ChatGPT. However, recent works have shown that inaccurate reward modeling can be exploited by the RL optimization~\citep{wolf2023fundamental}, which is also called ``reward hacking'' problem in the RL formalization~\citep{ibarz2018reward,hadfield2017inverse}. Future work can seek more diverse and fine-grained signals to replace scalar form rewards to aid a more stable and efficient alignment training.

\paragraph{Social Embodiment.} 
NLP models should incorporate a more comprehensive view of the world, including an embodied and social context, to simulate realistic human behavior~\citep{experience-ground, social-neuro-ai}. This is because social and cultural factors heavily influence human behavior. Recently, generative agents have been introduced as a way to simulate believable human behavior by incorporating a LLM with a complete record of the agent's experiences~\cite{park2023generative}. However, there are still challenges to be addressed to improve the accuracy and complexity of such social agents. Scaling the iNLP to handle larger and more complex environments is a potential future direction. This would enable the agent to handle more ambitious simulations of human behaviors and generate more realistic responses to user interactions.

\paragraph{Plasticity.} 
A significant challenge encountered with iNLP is the constant need for updates to adapt to changes in the real world. 
The prevalent approach in the academic literature typically utilizes gradient-based fine-tuning methods. 
These methods adjust an extensive number of parameters in the pre-trained models simultaneously, which can be overkill. 
Nonetheless, if an insufficient number of parameters are adjusted, the models may not effectively adapt to changes in real-world scenarios. 
Consequently, identifying methods for effective updates to iNLP models is essential for practical applicability~\citep{mitchell2021fast, mitchell2022memory, mitchell2022enhancing}. 
In recent years, burgeoning interest among researchers in the field of continual learning has emerged, aiming to enhance a model's capacity to learn persistently over time while minimizing the loss of previously acquired information. 
Continual learning enables the iNLP model to adapt to new data and dynamic situations without necessitating the retraining of the model from scratch. 
It is worth noting that biological neural networks acquire new skills continually within their lifetime based on neuronal plasticity~\citep{hebb2005organization}. 
Future research focusing on more human-like models~\citep{zador2023catalyzing,wang2021evolving} is anticipated to expedite advancements in continual learning for iNLP.

\paragraph{Speed \& Efficiency.} 
iNLP usually requires large language models as the backbone, thus suffers from their high latency and huge computational cost~\citep{DBLP:journals/cacm/SchwartzDSE20,DBLP:journals/corr/abs-2111-05193}. 
The high latency issue is even more crucial for iNLP compared to conventional NLP due to the need for frequent iterative calls. A large number of work has been done on improving the speed and efficiency of large language models, including both \textit{static} methods such as knowledge distillation~\citep{sanh2019distilbert,zhou-etal-2022-bert}, pruning~\citep{NEURIPS2019_2c601ad9,gordon-etal-2020-compressing}, quantization~\citep{DBLP:conf/aaai/ShenDYMYGMK20,dettmers2022gptint} and module replacing~\citep{xu-etal-2020-bert}; and \textit{dynamic} methods such as adaptive computation~\citep{graves2017adaptive}, early-exiting~\citep{schwartz-etal-2020-right,DBLP:conf/nips/ZhouXGM0W20}, and model cascade~\citep{li-etal-2021-cascadebert-accelerating,varshney-baral-2022-model}. However, most of the aforementioned methods require access to the model parameters, which may not be possible in the future since most state-of-the-art generalist models such as ChatGPT and PaLM~\citep{chowdhery2022palm, google2023palm2} are closed-sourced. Therefore, developing techniques that can accelerate inference for LLMs without access to their parameters is a promising future for efficient iNLP. 
Moreover, it is important to consider not only the acceleration ratio or preserved performance of accelerated models but also their robustness, biases, and alignment~\citep{xu-etal-2021-beyond}.

\paragraph{Context Length.} 
Context length refers to the maximum numbers of input tokens permitted by a language model. 
For example, ChatGPT has a context window of 8K tokens, while GPT-4~\citep{gpt4} extends it to 32K tokens. 
iNLP can greatly benefit from a long context window. 
The reason is three-fold: 
(1) It allows for maintaining and understanding a more extensive conversational history. 
(2) The ability to process a longer context is crucial for tasks that involve large pieces of text, such as long document-based QA and a detailed observation in the environment. 
(3) It can also facilitate the generation of long-form content. 
Recent studies on memorizing 
Transformers~\citep{wumemorizing,bulatov2022recurrent,bulatov2023scaling, liang2023unleashing} have illustrated the potential to scale the context window to tens of thousands of tokens using memory augmentation techniques. 
Additionally, Anthropic has introduced a chatbot with a 100K context window\footnote{\url{https://www.anthropic.com/index/100k-context-windows}}. 
However, despite these advancements, more research is needed to investigate the challenges associated with significantly increasing the context length.

\paragraph{Long Text Generation.} 
The capability to generate long text is crucial in iNLP contexts. 
For example, 
in real-life conversations, humans frequently convey intricate ideas and participate in extremely long discussions that necessitate numerous rounds of information exchanges. 
Moreover, for long-horizon robotic manipulation tasks, LMs need to generate a long action plan for execution. 
However, as the generated text lengthens, current language models have the propensity to produce content that may lack structure, coherence, quality, and even the relevance to the input prompts. 
Consequently, more sophisticated natural language processing techniques are needed to accurately capture the subtleties of language and produce text that is both coherent and useful.

\paragraph{Accessibility.} 
In the realm of large language model deployment, accessibility emerges as a critical concern. The most prominent LLMs, such as the GPT-family models~\citep{brown2020language, ouyang2022training, gpt4} and Bard\footnote{\url{https://bard.google.com/}}, are predominantly closed-source, creating a significant barrier for those seeking to utilize them for specific purposes. Recently, researchers have shifted their focus to developing open-source LLMs, including LLaMA~\citep{touvron2023llama}, Pythia~\citep{biderman2023pythia}, and GLM~\citep{du2022glm}. The movement towards open-sourcing large language models is expected to gain momentum in the future. 
Another emerging trend that has received limited research attention so far is the accessibility of deploying LLMs on edge devices such as smartphones, laptops, and automobiles, despite the existence of several previous works on the topic~\citep{niu2020realtime}\footnote{\url{https://github.com/mlc-ai/mlc-llm}}. 
Research towards more accessible language models can expand the possibilities for iNLP. 
For instance, it can be particularly beneficial in scenarios that involve offline interaction
\footnote{In situations where network connectivity is unavailable, relying on closed-source language models accessed through the Internet becomes infeasible. Therefore, the use of a local language model becomes necessary.}.

\paragraph{Analysis.} 
Although the interactive language models have shown a powerful ability to understand and generate complex language across a wide range of topics and contexts, its ``inner workings'' are still a black box for both the users and the researchers. 
We assume that gaining a deeper understanding of LMs and their interpretability can lead to improved interaction behaviors exhibited by LM agents. 
For example, 
~\cite{bills2023language} utilizes GPT-4 to provide explanations for all the neurons in GPT-2~\citep{radford2019language} by analyzing their activations in response to input text. 
Intuitively, such explainability can facilitate knowledge updates in language models within the context of iNLP~\citep{meng2022locating, meng2022memit}. 
Additionally, the analysis of scaling laws~\citep{kaplan2020scaling,tay2022scalingvsarch}, emergent abilities~\citep{wei2022emergent}, scaling-up performance prediction~\citep{gpt4}, trade-offs between alignment and general performance~\citep{wolf2023fundamental}, interpretability for the interaction behavior of LMs~\citep{park2023generative, kosinski2023theory}, are also promising avenues for future research. 

\paragraph{Creativity.} 
Contrary to the prevailing language modeling approach, which relies on learning statistical relationships, creativity involves the generation of original ideas, concepts, or perspectives that deviate from conventional patterns. 
The pursuit of creativity has long been a significant challenge in the AI community, driven by the desire to develop human-level agents~\citep{LeCun2022path} capable of generating novel knowledge and contributing original ideas across various domains. 
To effectively generate more creative content, it is crucial to establish a detailed definition or judging criteria of creativity. 
For instance, generating novel metaphors requires establishing conceptual mappings between the source and target domains~\citep{li-etal-2022-cm,li-etal-2023-framebert}, whereas story generation involves the creation of original, coherent, and engaging narratives and plotlines~\citep{tang-etal-2022-etrica,tang-etal-2022-ngep}.
Additionally, the ability to generate new knowledge in the generated text, rather than solely extracting existing knowledge, can contribute to enhancing creativity.
Furthermore, it is essential to explore approaches for enhancing creativity in generated content to ensure practical utility. For instance, enabling language models to discover theories or laws based on observed phenomena requires dedicated efforts and potentially entails exploring new paradigms for language models to engage in conscious thinking~\citep{bengio2017consciousness}. 
Research towards more creative language models may unlock a range of complex interactive properties or behaviors of LMs, such as the development of a more creative writing assistant or even the emergence of sophisticated debates between language model agents. 

\paragraph{Evaluation.} 
As shown in \S\ref{eval}, evaluation for iNLP is still barren and lacks diversity. 
How to design a better evaluation method will be one of the most important research topics in the future, which will profoundly affect the design and optimization direction of the iNLP frameworks. 
Specifically, evaluation methods under interactive settings may develop in the following aspects: 
(1) Pay more attention to the evaluation of the interaction process rather than just the result~\citep{Evaluating_Human_Language_Model_Interaction}. 
(2) Design a more standard evaluation benchmark to support the comparison of different interactive models. 
(3) Evaluate the interactivity of large language models.

\section{Conclusion}
In this paper, we have offered a comprehensive exploration of Interactive Natural Language Processing, a burgeoning paradigm that situates language models as interactive agents within a diverse array of contexts. We have proposed a unified definition and framework for iNLP, followed by a systematic classification that deconstructs its integral components such as interactive objects, interfaces, and methods. Furthermore, we have elucidated the varied evaluation methodologies used in the field, showcased its numerous applications, discussed its ethical and safety issues, and pondered upon future research directions. By putting a spotlight on iNLP's ability to interact with humans, knowledge bases, models, tools, and environments, we have underscored the paradigm's potential for enhancing alignment, personalizing responses, enriching representations, avoiding hallucinations, decomposing complex tasks, and grounding language in reality, etc. Ultimately, this survey presents a wide-angle view of the current state and future potential of iNLP, serving as an essential reference point for researchers eager to dive into this rapidly evolving field.

\section{Acknowledgements} 
We would like to thank Pengfei Liu for his constructive comments on this work. 
We would like to thank Haoran Zhang, Yang Liu, Wenzhen Miao, and Iman Yeckehzaare for the early-stage discussion of the paper. 
We would like to thank Ziwei Zhu, Ziqiao Ma, Yichi Zhang, Renliang Sun, Xingran Chen, and Chenghao Xiao for proofreading the paper.

\bibliographystyle{tmlr}
\bibliography{references}

\newpage

\appendix

\section{Contributions}

\paragraph{Unification Management.} 
Ge Zhang and Zekun Wang co-led this project. 
Zekun Wang was responsible for designing the survey outline, providing the motivation behind the project, and managing various aspects such as paper collection, paper reading, draft writing, discussions, and more. Ge Zhang took charge of the weekly meetings, task allocation, member invitations, and other related responsibilities. Both Ge Zhang and Zekun Wang actively monitored and adjusted the project's progress, ensuring the overall quality of the writing.

\paragraph{Paper Collection and Sharing.} 
All the members collected and presented papers in the weekly meetings. Among them, Zekun Wang shared most of the papers. Ge Zhang, Guangzheng Xiong, Kexin Yang and Shaochun Hao also shared a lot of literature with project members.

\paragraph{\textit{Introduction} Writing.} 
Zekun Wang wrote this section. Chenghua Lin wrote part of this section, provided suggestions, and edited the entire section. 

\paragraph{\textit{Interactive Objects} Writing.} 
Zekun Wang initially wrote the first version of this section. Wenhu Chen wrote part of this section, provided suggestions, and edited the entire section, primarily focusing on the subsections of ``KB-in-the-loop'' and ``Model/Tool-in-the-loop''. Wenhu Chen categorized ``Knowledge Sources'' into ``Corpus Knowledge'' and ``Internet Knowledge'', and restructured ``Model/Tool-in-the-loop'' as ``Tool-use'' and ``Multi-model Collaboration''. Later, Zekun Wang further refined ``Model/Tool-in-the-loop'' and divided it into ``Thinking'', ``Acting'', and ``Collaborating''.

\paragraph{\textit{Interaction Interface} Writing.} 
Ge Zhang and Ning Shi wrote the subsection of ``Edits''. The other subsections were written by Zekun Wang, including ``Natural Language'', ``Formal Language'', ``Machine Language'', and ``Shared Memory''. Chenghua Lin provided suggestions and edited this section.

\paragraph{\textit{Interaction Methods} Writing.} 
In this section, the outline was designed by Zekun Wang. The subsection on ``Pre-trained Language Models'' was written by Ge Zhang, Yizhi Li, and Zekun Wang. Ge Zhang primarily contributed to the tables in this subsection. Zekun Wang independently wrote the subsections on ``Standard Prompting'' and ``Prompt Chaining''. The subsections on ``Elicitive Prompting'' were written by Ruibo Liu. ``Supervised Instruction Tuning'' was authored by Qingqing Zhu. ``Continual Learning'' was collaboratively written by Xiuying Chen, Mong Yuan Sim, and Zekun Wang. ``Parameter-Efficient Fine-Tuning'' was co-written by Wangchunshu Zhou and Zekun Wang. ``Semi-Supervised Fine-Tuning'' was co-authored by Shaochun Hao and Zekun Wang. The subsection on ``Active Learning'' was written by Ge Zhang. ``Reinforcement Learning'' was authored by Guangzheng Xiong. ``Imitation Learning'' was written by Ning Shi. The subsection on ``Interaction Message Fusion'' was authored by Zekun Wang. Jie Fu and Zekun Wang provided guidance, suggestions, and editing for all the subsections in this section.

\paragraph{\textit{Evaluation} Writing.} 
Kexin Yang wrote this section. Zekun Wang partly added content for ``Knowledge Acquisition'', ``Chain-of-Thought Capability'', ``Tool-Use Ability'', and ``Collaborative Behavior Analysis''. Dayiheng Liu and Zekun Wang provided suggestions and edited this section.

\paragraph{\textit{Application} Writing.} 
Xiuying Chen, Shaochun Hao and Yizhi Li wrote ``Controllable Text Generation''. Ge Zhang wrote ``Writing Assistant''. Guangzheng Xiong wrote ``Embodied AI''. Zhenzhu Yang and Ge Zhang wrote ``Text Game''. Xiuying Chen, Kexin Yang and Mong Yuan Sim wrote the other applications. Ke Xu and Zekun Wang provided suggestions and edited this section. 

\paragraph{\textit{Ethics and Safety} Writing.} 
Mong Yuan Sim authored this section. Chenghua Lin contributed to the writing of ``Impact on Education'', provided suggestions, and edited the content of the entire section. 

\paragraph{\textit{Future Directions} Writing.} 
This section was collaboratively written by Wenhu Chen, Jie Fu, Ruibo Liu, Kexin Yang, Wangchunshu Zhou, Chenghua Lin, Zekun Wang, Qi Liu, Mong Yuan Sim, Ge Zhang, and Xiuying Chen. Ke Xu provided suggestions and edited this section. 

\paragraph{Supervision.} 
Shi Wang provided part of the funding for this project and actively participated in discussions. Jie Fu, Chenghua Lin, and Yike Guo provided valuable guidance on the entire paper throughout the whole duration of the project's lifecycle.

\end{document}